\documentclass[twoside]{article}

\usepackage[accepted]{aistats2025}
\usepackage[round]{natbib}

\usepackage{algorithm}
\usepackage{algorithmic}
\usepackage{amsmath}
\usepackage{amssymb}
\usepackage{mathtools}
\usepackage{amsthm}
\usepackage{thm-restate}

\theoremstyle{plain}
\newtheorem{theorem}{Theorem}[section]

\newtheorem{lemma}[theorem]{Lemma}

\theoremstyle{definition}

\theoremstyle{remark}

\renewcommand\thmcontinues[1]{\emph{cont.}}

\usepackage[textsize=tiny]{todonotes}
\usepackage{bm}
\usepackage{thm-restate}

\usepackage{xspace}

\newcommand{\E}{\mathbb{E}}

\newcommand{\argmin}{\operatornamewithlimits{argmin}}
\newcommand{\argmax}{\operatornamewithlimits{argmax}}

\newcommand{\F}{\mathcal{F}}

\newcommand{\cM}{\mathcal{M}}
\newcommand{\cU}{\mathcal{U}}
\newcommand{\cA}{\mathcal{A}}

\newcommand{\cH}{\mathcal{H}}
\newcommand{\cE}{\mathcal{E}}
\newcommand{\cD}{\mathcal{D}}
\newcommand{\cS}{\mathcal{S}}
\newcommand{\cO}{\mathcal{O}}
\newcommand{\bP}{\mathbb{P}}
\newcommand{\bE}{\mathbb{E}}
\newcommand{\bN}{\mathbb{N}}
\newcommand{\bR}{\mathbb{R}}

\usepackage{dsfont}
\newcommand{\indi}[1]{\mathds{1}\left(#1\right)}
\newcommand{\cN}{\mathcal{N}}
\newcommand{\cC}{\mathcal{C}}
\newcommand{\eqdef}{ := }
\newcommand{\simplex}{\triangle_{K}}

\usepackage{hyperref}
\usepackage{xcolor}
\definecolor{Bleu}{RGB}{30,144,245}
\definecolor{Red}{HTML}{FF617B}
\hypersetup{colorlinks,citecolor=Bleu,linkcolor=Red}

\begin{document}


\runningauthor{Riccardo Poiani${^\star}$, Marc Jourdan${^\star}$, Emilie Kaufmann, Rémy Degenne}

\twocolumn[

\aistatstitle{Best-Arm Identification in Unimodal Bandits}

\aistatsauthor{Riccardo Poiani${^\star}{^1}$ \And Marc Jourdan${^\star}{^2}$ \And Emilie Kaufmann$^3$ \And Rémy Degenne$^3$}

\aistatsaddress{$^1$Bocconi University, Milan, Italy $\quad$ $^2$EPFL, Lausanne, Switzerland \\$^3$Univ. Lille, Inria, CNRS, Centrale Lille, UMR 9189-CRIStAL } ]

\begin{abstract}
  We study the fixed-confidence best-arm identification problem in unimodal bandits, in which the means of the arms increase with the index of the arm up to their maximum, then decrease.
  We derive two lower bounds on the stopping time of any algorithm. 
  The instance-dependent lower bound suggests that due to the unimodal structure, only three arms contribute to the leading confidence-dependent cost.
  However, a worst-case lower bound shows that a linear dependence on the number of arms is unavoidable in the confidence-independent cost. 
  We propose modifications of Track-and-Stop and a Top Two algorithm that leverage the unimodal structure.
  Both versions of Track-and-Stop are asymptotically optimal for one-parameter exponential families.
  The Top Two algorithm is asymptotically near-optimal for Gaussian distributions and we prove a non-asymptotic guarantee matching the worse-case lower bound.
  The algorithms can be implemented efficiently and we demonstrate their competitive empirical performance.
\end{abstract}

% !TeX root = ../paper.tex

\section{INTRODUCTION}
\label{sec:intro}

In a bandit pure exploration problem, an agent interacts with a set of $K$ probability distributions denoted by $\bm\nu = (\nu_i)_{i \in [K]} \in \cD^{K}$, called arms, whose means $\bm\mu = (\mu_i)_{i \in [K]}$ belong to some subset $\cS \subseteq \bR^{K}$ materializing the possible structure between arms. 
In each time step $t \ge 1$, one arm $I_t \in [K]$ is selected, and the agent observes a realization $X_t \sim \nu_{I_t}$. 
Based on as few observations as possible, her goal is to confidently answer some questions about the unknown mean vector $\bm\mu$, while leveraging the underlying structure. 

The most studied pure exploration task is Best-Arm Identification (BAI), in which the agent seeks to identify the arm having the highest mean when $\cS = \Theta^{K}$ is the (unstructured) set of all possible means vectors. 
In this paper, we consider best-arm identification when the set $\cS$ of means is constrained to have a \emph{unimodal structure}, namely the means are known to increase up to their unique maximum and then decrease with the index. 
Let $i^\star(\boldsymbol{\mu}) = \argmax_{i \in [K]} \mu_i$ be the unique maximum, which we denote by $i^\star$ or $\star$ when $\bm \mu$ is clear from the context.
The set of \emph{unimodal instances} is
\begin{align} \label{eq:structure_Unimodal}
	\cS \eqdef \{ \boldsymbol{\lambda} \in \bR^K \mid \: & |i^\star(\boldsymbol{\lambda})| = 1, \: \forall j < i^\star(\boldsymbol{\lambda}) , \lambda_{j} \le \lambda_{j+1}\: , \nonumber \\
	  &\qquad \forall j \ge i^\star(\boldsymbol{\lambda}) , \lambda_{j} \ge \lambda_{j+1} \} \:.
\end{align}
We study the \emph{fixed-confidence} formulation of the problem~\citep{even-dar_action_nodate,jamieson_best-arm_2014,garivier2016optimal}, in which the agent aims at minimizing the number of samples used to identify $i^\star(\bm \mu)$ with confidence $1-\delta \in (0,1)$.
The agent's strategy is given by three rules: the \emph{sampling rule} $\left( I_t \right)_t$, where each $I_t$ is $\F_{t-1}$-measurable,\footnote{We denote by $\mathcal{F}_t = \sigma\left( I_1, X_1, \dots, I_t, X_t \right)$ the $\sigma$-algebra generated by the observations collected up to time $t$.} a \emph{stopping rule} $\tau_{\delta}$, which is a stopping time w.r.t. $\F_t$ and a \emph{recommendation rule} $\hat \imath_{\tau}$, which is $\F_\tau$-measurable. 
The algorithm's \emph{sample complexity} corresponds to its stopping time $\tau_{\delta}$, which counts the number of rounds before termination.
An algorithm is said to be \emph{$\delta$-correct} on the problem class $\cD^{K}$ having means $\cS$ if its probability of stopping and not recommending a correct answer is upper bounded by $\delta$, i.e. $\bP_{\bm\nu}\left(\tau_{\delta} < + \infty, \: \hat \imath_{\tau_{\delta}} \neq i^\star(\bm\mu)\right) \leq \delta$ for all instances $\bm \nu \in \cD^K$ having mean $\bm\mu \in \cS$. 
The agent should design a $\delta$-correct algorithm minimizing the expected sample complexity $\bE_{\bm\nu}[\tau_{\delta}]$. 

Unimodal bandits have been studied extensively in the regret minimization literature~\citep{yu2011unimodal,combes14unimodal,Paladino17UTS,trinh20rankone,saber21}, in which observations are rewards that the agent seeks to maximize.
It has also been tackled in the \emph{fixed-budget} best-arm identification setting~\citep{cheshire2020influence,ghoshfixed}, where the agent should minimize the probability of misidentifying $i^\star(\bm \mu)$ with a fixed number of samples~\citep{karnin2013almost,audibert2010best}. 
These studies were motivated by potential applications to sequential pricing~\citep{yu2011unimodal} or bidding in sponsored search auctions \citep{edelman05}.
Their extension to graphical unimodal bandits~\citep{combes14unimodal} can be relevant for recommender systems~\citep{Paladino17UTS,trinh20rankone}. 
In this paper, we propose the first analysis of unimodal BAI in the fixed-confidence setting.
We study unimodal bandits as a theoretical curiosity, in which the structure induces a \emph{sparsity pattern} in the optimal allocation.
This sparsity induced by the structure is a central component of our algorithms.

\paragraph{Related Work} 
In fixed-confidence BAI, the seminal work of \cite{garivier2016optimal} introduced the Track-and-Stop (TaS) algorithm for unstructured bandits.
It achieves asymptotically optimal expected sample complexity rates by computing the optimal allocation of the empirical estimator of the means at each round. 
Since then, several studies have proposed adaptation of TaS in different structured pure exploration problems, e.g., linear bandits \citep{jedra2020optimal}, multiple answers problems \citep{degenne2019pure}, and many others \citep{moulos2019optimal,agrawal2020optimal,russac2021b}. 
We will show how to account for the unimodal structure in TaS while providing a computationally efficient implementation.
Within the class of structured bandits, it is worth mentioning that sparsity patterns in the instance-dependent lower bound have arisen in the minimum threshold problem \citep{kaufmann2018sequential}, good arm identification~\citep{jourdan2023anytime} and in multi-fidelity bandits \citep{poiani2024optimal}.
% Nevertheless, none of these work take the sparsity into account while designing an optimal algorithm.\todo{I don't necessarily agree with this statement. In GAI, our algo APGAI fully leverage sparsity.}

Other approaches lead to asymptotically optimal algorithms in general structured pure exploration problems: the online optimization-based approach~\citep{menard2019gradient,degenne2019non,wang21FW} and the Top Two approach~\citep{russo2016simple,jourdan22toptwo,you2023information}. 
The online optimization-based approach could be applied to solve unimodal bandits.
However, it does not explicitly exploit the sparsity pattern, hence incur sub-optimal performance in the moderate regime of $\delta$.
In addition to being computationally less demanding for large values of $K$, Top Two algorithms can be modified to include the sparsity pattern of unimodal BAI.

\paragraph{Contributions}
\textbf{(1)} We show that the unimodal structure induces sparsity in the optimal allocation (achieving the characteristic time $T^\star(\bm \mu)$) in which only the best arm and its neighbors are contributing asymptotically (Theorem~\ref{theo:sparsity}).
 However, in the moderate-confidence regime, we prove a \emph{worst-case} lower bound whose linear dependence on $K$ suggests that sparsity is only possible asymptotically (Theorem~\ref{thm:lb_K}).
\textbf{(2)} We carefully exploit the unimodal structure in the recommendation and stopping rules, which ensure $\delta$-correctness (Lemma~\ref{lem:delta_correct_threshold}). 
As sampling rules, we propose adaptations of the Track-and-Stop sampling rules using forced exploration or optimism and a Top Two sampling rule using optimism.
We detail an efficient implementation of those three algorithms and show that they achieve competitive empirical performance.
\textbf{(3)} The two Track-and-Stop algorithms are asymptotically optimal for one-parameter exponential families (Theorems~\ref{thm:TaS_asymptotic_optimality} and~\ref{thm:SAOTaS_asymptotic_optimality}).
The optimistic Top Two algorithm enjoys a non-asymptotic guarantee matching the worse-case lower bound and recovering near asymptotic optimality (Theorem~\ref{thm:non_asymptotic_upper_bound_UniTT}).

\paragraph{Setting and Notation} 
The arms distributions belong to a given one-parameter exponential family $\cD \subseteq \mathcal P(\bR)$ (e.g. Bernoulli or Gaussian with known variance) with means in an interval $\Theta$~\citep{cappe2013kullback}. 
The instance $\bm \nu$ is fully characterized by its mean vector $\bm\mu \in \cS$.
For any arm $i \in [K]$, we denote by $\mathcal{N}(i)$ its set of \emph{neighboring arms}, i.e. $\mathcal{N}(i)$ is defined as $\{ 2 \}$ if $i=1$, $\{ K-1 \}$ if $i = K$, and $\{ i-1, i+1 \}$ otherwise. 
For two means $(\mu, \theta) \in \Theta^2$, we denote by $d(\mu,\theta)$ the KL divergence between their associated distribution in $\cD$. 
The $(K-1)$-dimensional simplex is denoted by $\simplex = \{\omega \in \bR_{+}^{K} \mid \sum_{i \in [K]} \omega_{i} = 1\}$.

% !TeX root = ../paper.tex

\section{LOWER BOUNDS}\label{sec:lb} 

We present lower bounds on the expected sample complexity of any $\delta$-correct algorithm for Unimodal BAI: an instance-dependent one scaling as $T^\star(\bm \mu) \log(1/\delta)$ (Theorem~\ref{theo:sparsity}) and a worst-case one scaling as $K/\Delta^2$ where $\Delta$ is the minimum gap (Theorem~\ref{thm:lb_K}).

\subsection{Dependence in the Bandit Model $\mu$ and the Risk Parameter $\delta$}\label{subsec:dep-delta}

We recall a standard instance-dependent lower bound on the expected stopping time $\E_{\bm{\nu}}[\tau_\delta]$ of any $\delta$-correct BAI algorithm~\citep{garivier2016optimal}.
\begin{restatable}{theorem}{lb}\label{prop:lb}
For any $\delta$-correct algorithm and any unimodal instance $\bm \nu \in \cD^{K}$ with mean $\bm{\mu} \in \mathcal{S}$, we have $\E_{\bm{\nu}}[\tau_\delta] \ge T^\star(\bm{\mu}) \log \frac{1}{2.4 \delta} $ with 
\begin{align}\label{eq:char-time}
    &T^\star(\bm{\mu})^{-1}  \eqdef  \sup_{\bm{\omega} \in \Delta_{K}} \min_{i \ne \star} f_i(\bm{\omega}, \bm{\mu}) \quad \text{and} \\ 
    &f_i(\bm{\omega}, \bm{\mu}) \eqdef \inf_{\substack{\bm{\theta} \in \textup{Alt}(\bm{\mu}): \\ i^{\star}(\bm{\theta})=i}} \sum_{j \in [K]} \omega_{j} {d(\mu_{j}, \theta_{j})} \nonumber \: ,
\end{align}
where $\textup{Alt}(\bm{\mu}) = \{ \bm{\theta} \in \mathcal{S}: i^{\star}(\bm{\theta}) \ne i^{\star}(\bm{\mu}) \}$ and $\bm{\omega}^\star(\bm{\mu})$ denote the maximizer of $T^\star(\bm{\mu})^{-1} $.
\end{restatable}
Theorem~\ref{prop:lb} introduces the value of two-player zero sum game $T^\star(\bm{\mu})^{-1}$~\citep{degenne2019pure}.
The max-player first chooses a strategy over the arm space $\bm{\omega} \in \Delta_K$ and the min-player best-responds to $\bm{\omega}$ by picking an alternative unimodal bandit $\bm{\theta}$, whose best arm is not $i^\star(\bm{\mu})$, with the goal of minimizing $\bm \theta \mapsto \sum_{i \in [K]} \omega_{i} {d(\mu_{i}, \theta_{i})}$. 
The allocation $\bm{\omega}^\star(\bm{\mu})$ is the expected proportion of pulls that an asymptotically optimal algorithm plays to identify $\star$ the fastest.

\begin{restatable}{theorem}{sparsity}\label{theo:sparsity}
Let $\bm{\mu} \in \mathcal{S}$.
Then, we have
\begin{align}\label{eq:char-time-restricted}
T^\star(\bm{\mu})^{-1} = \sup_{\bm{\omega} \in \widetilde{\Delta}_K(\bm{\mu})} \min_{i \in \mathcal{N}(\star)} g_{i}(\bm{\omega},\bm{\mu}),
\end{align}
where $g_{i}(\bm{\omega},\bm{\mu}) \coloneqq \inf_{x \in (\mu_i, \mu_{\star})} \omega_{\star} d(\mu_{\star}, x) + \omega_i d(\mu_i, x)$ and $\widetilde{\Delta}_K(\bm{\mu}) = \{\omega \in \simplex \mid \forall i \notin \cN(\star) \cup\{\star\}, \: \omega_i = 0\}$.
\end{restatable}
Theorem~\ref{theo:sparsity} introduces a rewriting of $T^\star(\bm{\mu})^{-1}$ that reveals a \emph{sparsity pattern} of the oracle weights $\bm{\omega}^\star$ by reducing the supremum over $\simplex$ in~\eqref{eq:char-time} to a supremum over the restricted simplex $\widetilde{\Delta}_K(\bm{\mu})$ in~\eqref{eq:char-time-restricted}. 
% This implies that all the arms which are not in $i^{\star}(\bm{\mu}) \cup \mathcal{N}(\star)$ have value $0$ at the optimal allocation $\bm{\omega}^\star$. 
Therefore, the optimal weights $\bm{\omega}^\star(\bm{\mu})$ are only spread among the best arm $\star$ and its neighboring arms $\cN(\star)$.

Given the expression of $g_{i}(\bm{\omega}, \bm{\mu})$, Theorem~\ref{theo:sparsity} also establishes a reduction of Unimodal BAI to unstructured BAI on the three-arms set $\{\star\} \cup \cN(\star)$.
Therefore, we inherit a fast and $K$-independent algorithm to compute the oracle weights $\bm{\omega}^\star(\bm \mu)$ for any unimodal bandit $\bm{\mu} \in \cS$.
% Indeed, given the expression of $g_{i}(\bm{\omega}, \bm{\mu})$, Equation \eqref{eq:char-time-restricted} is exactly equivalent to an unstructured BAI problem in which the arm space is restricted to the optimal arm $i^{\star}(\bm{\mu})$ and its neighborhood $\mathcal{N}(\star)$. 
% Therefore, as a direct consequence, we inherit from \cite{garivier2016optimal} a fast algorithm to compute the oracle weights $\bm{\omega}^\star$ for any unimodal bandit $\bm{\mu}$.\footnote{Note that the complexity of computing $\bm{\omega}^\star$ in unstructured BAI is proportional to the number of arms $K$. In the unimodal setting, thanks to Theorem \ref{theo:sparsity}, the number of arms involved in the computation of $\bm{\omega}^\star$ is at most $3$.} 
The reduction from Unimodal BAI to unstructured BAI holds since, for the optimal allocation only, we have $f_{i}(\bm{\omega}^\star(\bm \mu), \bm{\mu}) = g_i(\bm{\omega}^\star(\bm \mu), \bm{\mu})$. 
This equality does not hold for arbitrary weights $\bm{\omega}$.
Therefore, any algorithm computing $\min_{a \ne \star} f_{i}(\bm{\omega}, \bm{\mu})$ for arbitrary $\bm{\omega}$ (e.g. gradient-based~\citep{menard2019gradient} or game-based~\citep{degenne2019non} approaches) is computationally demanding since it requires to solve $K$-dimensional convex problems $f_{i}(\bm{\omega}, \bm{\mu})$ for each $i \ne \star$.

%
%As we shall see in Section \ref{sec:algos}, these properties we highlighted, will play an important role in the computational complexity of efficient algorithms to solve the underlying pure exploration problem. 

\subsection{Dependence in the Number of Arms}\label{subsec:dep-k}

The above lower bound depends on the risk parameter $\delta$ and the instance $\bm{\mu}$ through $T^\star(\bm{\mu})^{-1}$.
Therefore, it is independent of $K$ due to the reduction to three-arm unstructured BAI (Theorem~\ref{theo:sparsity}).  
For Gaussian distributions with unit variance, it yields that $T^\star(\bm{\mu}) \approx  \sum_{i \in \mathcal{N}(\star)} (\mu_{\star} - \mu_i)^{-2}$.
To capture the dependency in $K$ for Unimodal BAI, we derive a second lower bound showing that a linear dependence in $K$ is actually unavoidable for moderate regime of $\delta$.
The proof technique is borrowed from~\citet{simchowitz17theSimulator,al2022complexity}, see Appendix~\ref{app:proof-sec-lb}.
%, who studied the problem of identifying all arms with sub-optimality gap no worse that a given $\varepsilon$.

\begin{restatable}{theorem}{lbk}\label{thm:lb_K}
Let $\Delta>0$ and, for $i \in [K]$, let $\bm{\nu}^{(i)} \eqdef \cN(\bm{\mu}^{(i)}, I_{K})$ where ${\mu}^{(i)}_{i} = \Delta$ and ${\mu}^{(i)}_{j} = 0$ if $j \ne i$. 
Then, for any $\delta \le 1/4$ and any $\delta$-correct strategy,  
\[
    \frac{1}{K}\sum_{i \in [K]} \mathbb{E}_{\bm{\nu}^{(i)}}[\tau_\delta] \ge \frac{K}{64 \Delta^2} \: .
\]
There exists an instance $\bm{\nu}^{(i)}$ s.t. $\mathbb{E}_{\bm{\nu}^{(i)}}[\tau_\delta] \ge \frac{K}{64 \Delta^2}$.
\end{restatable}
Unlike Theorem~\ref{prop:lb}, Theorem~\ref{thm:lb_K} is not valid for any instance.
Among $K$ specific unimodal instances, any algorithm should use an average number of samples exceeding $K/(64 \Delta^2)$ on at least one of them.
On those $K$ ``hard'' instances, no arm reveals information about the optimal arm except the optimal arm itself, namely the instance is flat with a spike on the best arm. 
Therefore, an algorithm has to sample all arms before finding the best one.
In this initial exploration phase, an arm $i$ will be sampled less than the others, hence the algorithm will be slower on $\bm{\nu}^{(i)}$.
% Different algorithms may perform that initial exploration differently, but there is always an arm that will be sampled less than the others, and the algorithm will be slow to find the optimal arm on the instance where that one is the optimal arm. 

% !TeX root = ../paper.tex

\section{ALGORITHMS}
\label{sec:algos}

After specifying our recommendation and stopping rules (Section~\ref{ssec:rsp_rules}), we propose three sampling rules to solve unimodal BAI. %, each algorithmic design has its own advantages. 
We introduce U-TaS, which is an asymptotically optimal variant of TaS (Section~\ref{ssec:TaS}).
To alleviate the (wasteful) forced exploration of U-TaS, we provide an efficient implementation of the asymptotically optimal Optimistic TaS (O-TaS) algorithm, and show that O-TaS exploits the sparsity pattern (Section~\ref{ssec:SAOTaS}).
Since the non-asymptotic guaranty of O-TaS do not match Theorem~\ref{thm:lb_K}, we introduce an optimistic Top Two algorithm (Section~\ref{ssec:UniTT}) having near matching guarantees compared to Theorems~\ref{prop:lb} and~\ref{thm:lb_K}, as well as the lowest computational cost.

\paragraph{Notation}
Let $N_{i}(t) \eqdef \sum_{s \in [t-1]} \indi{I_s = i}$ and $\hat \mu_{i}(t) \eqdef N_{i}(t)^{-1}\sum_{s \in [t-1]} \indi{I_s = i} X_s$ denote the empirical count and mean of arm $i$ before time $t$.
For any arm $i \in [K]$ and any of its neighbor $j \in \cN(i)$, let
\begin{equation} \label{eq:TC_empirical}
	W_{t}(i,j) \eqdef \inf_{\theta_{j} \ge \theta_i} \sum_{k \in \{i,j\}} N_{k}(t) d(\hat{\mu}_{k}(t), \theta_{k}) \: ,
\end{equation}
denote their empirical transportation cost.
For Gaussian distributions with unit variance, we have
\begin{equation} \label{eq:Gaussian_TC}
	2 W_{t}(i,j) = \indi{\hat{\mu}_{i}(t) > \hat{\mu}_{j}(t)}\frac{(\hat{\mu}_{i}(t) - \hat{\mu}_{j}(t))^2}{1/N_{i}(t) + 1/N_{j}(t)} \: .
\end{equation}

\subsection{Recommendation and Stopping Rules}
\label{ssec:rsp_rules}

As done extensively in the literature, we consider the generalized likelihood ratio (GLR) stopping rule~\citep{garivier2016optimal}.
The GLR between two sets of bandit instances $A$ and $B$ is
\begin{align*}
\mbox{GLR}_t(A,B)
&= \log \frac{\sup_{\bm \lambda \in A} \mathcal{L}_{\bm \lambda}(X_1,\ldots,X_t)}{\sup_{\bm \theta \in B} \mathcal{L}_{\bm \theta}(X_1,\ldots,X_t)}
\: ,
\end{align*}
where $\mathcal L_{\bm\lambda}$ is the likelihood function of the distribution associated to $\bm\lambda$.
If there exists $i \in [K]$ for which $\mbox{GLR}_{t,i} \coloneqq \mbox{GLR}_t(\{\bm\lambda \in \mathcal S \mid i^\star(\bm\lambda)=i\}, \{\bm\lambda \in \mathcal S \mid i^\star(\bm\lambda) \ne i\})$ is large, then we have enough information to return answer $i$. 
Thus, we stop when $\max_i \mbox{GLR}_{t,i}$ is large.

\paragraph{Full-sum Stopping Rule}
The structure of $\mathcal S$ can make the GLR difficult to compute.
Previous work \citep[e.g.,][]{degenne2019non} often used an unstructured maximum likelihood, in which $\{\bm\lambda \in \mathbb{R}^K \mid i^\star(\bm\lambda)=i\}$ is used as first set. In this case, we stop when
\begin{equation} \label{eq:full_stopping_rule}
	\inf_{\bm{\theta} \in \textup{Alt}(\bm{\hat{\mu}}(t))} \sum_{i \in [K]} N_{i}(t) d(\hat{\mu}_{i}(t), \theta_{i}) \ge c_{K}(t-1, \delta) \: ,
\end{equation}
where $c_{K}(t, \delta) \approx \log\frac{1}{\delta} + K \log t$.
Due to the linearity in $K$, the l.h.s. of~\eqref{eq:full_stopping_rule} can be very large even if the lower bounds tells us that only 3 arms matter.
The \emph{full-sum} stopping rule denotes when we stop with~\eqref{eq:full_stopping_rule}.

In BAI, the heuristic threshold $\tilde c(t, \delta) \coloneqq \log((1+ \log t)/\delta)$ provides empirically $\delta$-correct algorithms.
However, for unimodal bandits the full-sum stopping rule~\eqref{eq:full_stopping_rule} with $\tilde c$ violates the empirical $\delta$-correctness criterion since it stops very early and returns wrong answers.
To better understand that failure, let's consider the unimodal flat instance where all arms have zero mean except for one having mean $\Delta > 0$. 
Suppose that $\Delta \ll \sigma$ where $\sigma$ is the standard deviation of the observations. 
After initialization, all empirical means are randomly spread around $0 \pm \sigma$.
Hence, we have $\sum_{i \in [K]} N_{i}(K+1) d(\hat{\mu}_{i}(K+1), \hat{\theta}_{i}(K+1)) \approx K D_\sigma$ for some distribution-dependent constant $D_\sigma$ where $\bm{\hat{\theta}}(t)$ minimizes~\eqref{eq:full_stopping_rule}. 
Therefore, removing the linear dependence in $K$ in $c_{K}(t, \delta) $ yields early stopping and errors.

\paragraph{Local GLR Stopping Rule}
By leveraging the unimodal structure, we derive an improved stopping rule.
%We show that $\max_i \mbox{GLR}_{t,i} = \max_i \min_{j \in \mathcal N(i)} W_t(i,j)$ (see Appendix~\ref{app:ssec_GLR}).
Intuitively, $W_{t}(i,j)$ represents the amount of collected evidence in favor of the hypothesis that arm $i$ has a larger mean than arm $j$ and $\min_{j \in \cN(i)} W_{t}(i,j)$ is the amount of collected evidence in favor of the hypothesis that arm $i$ is the best arm locally.
Given that the true instance is unimodal, being statistically confident of local optimality is a sufficient condition to be statistically confident of global optimality.
Since multiple arms can be locally optimal at stage $t$, we recommend the arm $i$ having the largest evidence of local optimality, i.e. the candidate answer 
\begin{equation} \label{eq:IFanswer}
	\hat \imath_t \in \argmax_{i \in [K]} \min_{j \in \cN(i)} W_{t}(i,j) \: .
\end{equation}
Given a threshold $c(t,\delta)$, we use the local GLR stopping rule associated with our candidate answer, i.e.
\begin{equation} \label{eq:GLR_stopping_rule}
	\tau_{\delta} = \inf \left\{ t \mid \min_{j \in \cN(\hat \imath_t )} W_{t}(\hat \imath_t ,j) \ge c(t-1,\delta) \right\} \: .
\end{equation}

Lemma~\ref{lem:delta_correct_threshold} gives a choice of threshold ensuring $\delta$-correctness regardless of the sampling rule (see Appendix~\ref{app:ssec_correctness}).
For other classes of one-parameter exponential families, we refer the reader to~\citet{kaufmann2021mixture} for threshold ensuring $\delta$-correctness.

\begin{lemma}\label{lem:delta_correct_threshold}
	Let $\delta \in (0,1)$. Given any sampling rule, using the threshold
	\begin{equation} \label{eq:stopping_threshold}
		c(t,\delta) = 2 \cC_{G} \left( \frac{1}{2}\log\left(\frac{K-1}{\delta} \right)\right) + 4 \log \log \frac{e^4t}{2} 
	\end{equation}
	with the stopping rule \eqref{eq:GLR_stopping_rule} yields a $\delta$-correct for Gaussian distributions with known variance and mean in $\cS$. 
	$\cC_{G}$ satisfies that $\cC_{G}(x) \approx x + \ln(x)$.
\end{lemma}

%Lemma~\ref{lem:delta_correct_threshold} would hold for any recommendation rule $i$ such that $W_{t}(i) > 0$, e.g. for the empirical best arm $i^\star(t) \eqdef i^\star(\bm{\hat{\mu}}(t))$.
%However, $\hat \imath_t$ as in~\eqref{eq:IFanswer} is the arm for which we have collected the most evidence.
%Again, local information is enough to conclude about a global property due to the unimodal structure.

This stopping rule does not suffer from the same limitation with the dependence in $K$ as the full-sum rule since $\min_{j \in \cN(\hat \imath_t)}W_{t}(\hat \imath_t,j)$ depends on only 3 arms.
It is also more computationally efficient, as computing $W_t(i,j)$ is much easier that solving the optimization problem for the infimum of the full sum.
%We now discuss different sampling rules to be combined with the stopping rule in \eqref{eq:GLR_stopping_rule}.

%\rp{Maybe we want to mention that our stopping rule is also more computationally efficient, as no convex opt. problem need to be solved.}
%\rp{Small remark: in the experiments, we are still paying a term $K$ with the new stopping rule (within the log for the union bound). Reading this paragraph it seems that we are using the usual heuristic, but usual heuristic failed in experiment at satisfying the $\delta$ requirement for large value of $K$ (although it failed way less w.r.t. the old stopping rule).}

\subsection{Track-and-stop}
\label{ssec:TaS}

Track-and-stop~\citep{garivier2016optimal} (TaS) is one of the most famous meta-algorithm to design asymptotically optimal algorithms.
To leverage the unimodal structure, we propose a computationally efficient variant of TaS, and refer to it as U-TaS.

After sampling each arm once, U-TaS solves the optimization problem defining the characteristic time of an empirical unimodal instance $\bm{\tilde{\mu}}(t)$ obtained by transforming $\bm{\hat{\mu}}(t)$.
% (at the end of this section, we explain the importance of this transformation).
Let $t > K$ and $i^\star(t) \eqdef i^\star(\bm{\hat{\mu}}(t))$.
U-TaS computes $\omega^\star(\bm{\tilde{\mu}}(t))$ where
\begin{equation} \label{eq:modified_empirical_mean}
	\tilde{\mu}_{i}(t) \eqdef \begin{cases}
		\tilde{\mu}_{i-1}(t) & \text{if } i > i^\star(t) \: \text{ and } \: \hat{\mu}_{i}(t) > \tilde{\mu}_{i-1}(t) \: , \\
		\tilde{\mu}_{i+1}(t) & \text{if } i < i^\star(t) \: \text{ and } \: \hat{\mu}_{i}(t) > \tilde{\mu}_{i+1}(t) \: , \\
		\hat{\mu}_{i}(t) & \text{otherwise.}
	\end{cases} 
\end{equation} 
While TaS solves $\omega^\star(\bm{\hat{\mu}}(t))$, $\omega^\star(\bm \lambda)$ can only be computed efficiently when $\bm \lambda \in \cS$ (see below).
Then, U-TaS computes the $l_{\infty}$ projection $\bm \omega(t)$ of $\omega^\star(\bm{\tilde{\mu}}(t))$ onto $\simplex^{(4(t + K^2))^{-1/2}}$ where $\simplex^{\epsilon} = \simplex \cap [\epsilon,1]^{K} $.
Given this allocation $\bm{\omega}(t)$, U-TaS selects the next arm to be pulled by tracking the cumulative sum of weights, i.e.
\begin{equation} \label{eq:C_tracking}
	I_t \in \argmax_{i \in [K]} \sum_{s \in [t]} \omega_i(s) - N_{i}(t) \: .
\end{equation}
This guarantees that $\bm N(t) \approx \sum_{s\in [t-1]} \bm \omega(s)$.

\paragraph{Asymptotic Optimality}
Theorem~\ref{thm:TaS_asymptotic_optimality} shows that U-TaS is asymptotically optimal (see Appendix~\ref{app:TaS}).
\begin{theorem} \label{thm:TaS_asymptotic_optimality}
	Using the GLR stopping rule as in~\eqref{eq:GLR_stopping_rule}, U-TaS is $\delta$-correct and $\limsup_{\delta \to 0} \bE_{\bm \nu}[\tau_{\delta}]/ \log(1/\delta) \le T^\star(\bm \mu)$ for all distributions $\bm \nu$ within a class $\cD$ of one-parameter exponential family having mean $\bm \mu \in \cS$.
\end{theorem}
Unfortunately, Theorem~\ref{thm:TaS_asymptotic_optimality} does not provide any insight on the non-asymptotic performance of U-TaS.
Figure~\ref{fig:mu-all} reveals that forced exploration is wasteful when arms are highly sub-optimal for large $K$ is large. 

\paragraph{Efficient Implementation}
TaS is one of the most computationally expensive among pure exploration algorithms due to the per-round computation of $\bm{\omega}^\star(\bm{\hat{\mu}}(t))$.
In structured bandits, there is no general oracle that returns the optimal allocation efficiently, e.g. using $\omega^\star(\bm{\hat{\mu}}(t))$ as in Theorem~\ref{prop:lb} would be inefficient. 
Due to the equivalence with an unstructured bandit on three arms when $\bm \mu \in \cS$, there exists an efficient oracle to compute $\omega^\star(\bm \mu)$ as in Theorem~\ref{theo:sparsity}~\citep{garivier2016optimal}.
Since the estimator of the mean might not be unimodal, i.e. $\bm{\hat{\mu}}(t) \notin \cS$, we modify it into a unimodal instance $\bm{\tilde{\mu}}(t) \in \cS$ for which we can compute $\omega^\star(\bm{\tilde{\mu}}(t))$ with a cost that is independent of $K$.
We consider~\eqref{eq:modified_empirical_mean} due to its $\cO(K)$ cost, but other choices could have been made.
The computational bottleneck in $\mathcal{O}(\textup{poly}(K))$  is the projection $\bm{\omega}^\star(\bm{\tilde{\mu}}(t))$ onto $\Delta_K^{\epsilon}$ (convex problem with $K$ variables). 

\subsection{Optimistic Track-and-Stop}
\label{ssec:SAOTaS}

Optimistic Track-and-Stop~\citep{degenne2019non} (O-TaS) is a variant of TaS using optimism to foster exploration.
We provide an efficient implementation of O-TaS for Unimodal BAI, and show it fosters sparsity.

\paragraph{Structured Confidence Region}
Let $\Theta_t$ be a confidence region around the empirical model $\bm{\hat{\mu}}(t)$, namely
\begin{equation} \label{eq:unstructured_CI}
	\Theta_t \eqdef \left\{ \bm{\theta}  \mid \forall i \in [K], \: N_i(t) d(\hat{\mu}_i(t), \theta_i) \le f(t)  \right\} \: ,
\end{equation}
where $f$ is distribution-dependent (e.g. a logarithm).
We have $\Theta_t = \bigotimes_{i \in [K]} [\alpha_i(t), \beta_i(t)] $ where $[\alpha_i(t), \beta_i(t)]$ by independence of the arms.
Let $\gamma_{1}, \gamma_{2} > 1$ and $f_{u}(t) \eqdef 2\gamma_{1}(1+\gamma_{2}) \log t$.
For $\bm \nu = \cN(\bm \mu, I_{K})$, let
\[
	\cE_t \eqdef \bigcap_{s \in [t^{1/\gamma_{1}}, t]} \bigcap_{i \in [K]} \{\mu_{i} \in [\hat{\mu}_{i}(s) \pm \sqrt{f_{u}(s)/ N_{i}(s)}] \} \: .
\]
It satisfies that $\bP(\cE_t^\complement) \le K/t^{\gamma_{2}}$ for all $t \in \bN$.
The structured confidence region is defined as $\widetilde{\Theta}_t \eqdef \Theta_t \cap \cS$.
% This restriction has a central role in our analyses. 

\paragraph{Optimistic Unimodal Instance}
After sampling each arm once, O-TaS compute an optimistic unimodal bandit $\bm{\mu}^+(t)$, and its corresponding oracle weights $\bm{\omega}(t)$, as the model within $\widetilde{\Theta}_t$ that maximizes the function $\bm \lambda \mapsto T^\star(\bm \lambda)^{-1}$, i.e. 
\begin{align*}
	&\bm{\mu}^+(t) \in \argmax_{\bm{\lambda} \in \widetilde{\Theta}_t} \sup_{\bm{\omega} \in \widetilde{\Delta}_K(\bm{\lambda})} \min_{i \in \mathcal{N}(i^{\star}(\bm{\lambda}))} g_{i}(\bm{\omega}, \bm{\lambda}) \: , \\
	&\bm{\omega}(t) \in \argmax_{\bm{\omega} \in \widetilde{\Delta}_K(\bm{\mu}^+(t))} \min_{i \in \mathcal{N}(i^{+}(t))}  g_{i}(\bm{\omega}, \bm{\mu}^+(t)) \: ,
\end{align*}
where $i^{+}(t) \eqdef i^\star(\bm{\mu}^+(t))$.
Given the proportions $\bm{\omega}(t)$, O-TaS selects the arm $I_t$ with C-tracking as in~\eqref{eq:C_tracking}.

\paragraph{Theoretical Guarantees}
We study the theoretical guarantees of O-TaS under the following assumptions. 

\begin{restatable}{assumption}{bounded}[Bounded parameters]\label{ass:bound}
There exists $M$ such that for all $\bm{\mu}, \bm{\mu'} \in \Theta$, $|| \bm{\nu}_{\bm{\mu}} - \bm{\nu}_{\bm{\mu'}} ||_{\infty} \le M$, where $\bm{\nu}_{\bm{\mu}}$  is the natural parameter vector corresponding to the mean vector $\bm{\mu}$.
\end{restatable}

\begin{restatable}{assumption}{subgauss}[Sub-Gaussianity]\label{ass:sub-gauss}
There exists $\sigma^2$ such that, for all $k \in [K]$ and, all $\bm{\mu}, \bm{\mu'} \in {\Theta}$, $d(\mu_k, \mu_k') \ge \frac{(\mu_k' - \mu_k)^2}{2\sigma^2}$.
\end{restatable}

\begin{restatable}{assumption}{expfun}[KL Concentration]\label{ass:exp-fun}
There exists a known, non decreasing function $f: \mathbb{N} \rightarrow  \mathbb{R}_{+}$ such that, for some constant $\gamma > 0$, it holds that:
\begin{align*}
\mathbb{P} \left( \exists k \in [K], t \le n, N_k(t) d(\hat{\mu}_k(t), \mu_k) > f(n) \right) \le \gamma \frac{\log n}{n^3}.
\end{align*}
\end{restatable}

These assumptions are satisfied for a general class of problems. We refer the reader to \cite{degenne2019non} for a discussion on these assumptions. We now state our main result on the performance of O-TaS.
Theorem~\ref{thm:TaS_asymptotic_optimality} is a corollary of a non-asymptotic upper bound (Theorem~\ref{theo-cmplx-O-TaS}) showing that O-TaS is asymptotically optimal.
Unfortunately, the non-asymptotic bound for O-TaS does not match the $K$-dependency from Theorem~\ref{thm:lb_K}.
\begin{theorem} \label{thm:SAOTaS_asymptotic_optimality}
	Suppose that our distributions satisfy Assumptions~\ref{ass:bound},~\ref{ass:sub-gauss} and~\ref{ass:exp-fun}.
	Using the GLR stopping rule as in~\eqref{eq:GLR_stopping_rule}, O-TaS is $\delta$-correct and $\limsup_{\delta \to 0} \bE_{\bm \nu}[\tau_{\delta}]/ \log(1/\delta) \le T^\star(\bm \mu)$ for all $\bm \mu \in \cS$.
\end{theorem}
% The proof of Theorem \ref{thm:SAOTaS_asymptotic_optimality} follows from a finite time analysis of the algorithm, and we refer the reader to Theorem \ref{theo-cmplx-O-TaS} for such guarantees. Nevertheless, it is important to note that this finite time analysis does not match the worse case dependency on $K$.

% \paragraph{O-TaS Exploits Sparsity} 
As local information is sufficient for global optimality, O-TaS exploits the sparsity pattern by limiting the samples from $[K] \setminus(\cN(i^\star) \cup\{i^\star\})$. 
By using the structured $\widetilde{\Theta}_t$, we exhibit a random time after which O-TaS only pulls arms in $\cN(i^\star) \cup\{i^\star\}$ (Theorem~\ref{thm:all-arm-elim}).
In Figure~\ref{fig:mu-all}, we see that O-TaS improves on U-TaS which lacks this property due to forced exploration.

% When dealing with sparsity patterns, we expect a good algorithm to limits the number of samples gathered at arms for which $\omega^*_i = 0$ holds. O-TaS actually ensures this property. Specifically, on the good event, we can prove that there exists a sufficiently large $T$ such that for all $t \ge T$, $N_a(t) = N_a(T)$ for all $a \notin \mathcal{N}(\star)$. (see Theorem \ref{thm:all-arm-elim} in the Appendix). This fact is a consequence of using the structured confidence intervals $\widetilde{\Theta}_t$ that we defined above.

\paragraph{Efficient Implementation}
In most pure exploration problems, Optimistic TaS is computationally intractable. 
For Unimodal BAI, O-TaS has an efficient implementation since computing $\bm{\mu}^+(t)$ requires $\mathcal{O}(K)$ operations. 
By iterating over the candidate optimistic answer $i \in [K]$, we notice that the associated optimal optimistic instance $\bm{\mu}^{(i)}(t)$ is obtained as 
\begin{equation} \label{eq:effSAO}
	{\mu}^{(i)}_j(t) = \begin{cases}
	\beta_i(t)  & \text{if } j=i \: , \\ 
	\max \{ \mu^{(i)}_{j-1}, \alpha_j(t) \} & \text{if }  j < i \: , \\
	\max \{ \mu^{(i)}_{j+1}, \alpha_j(t) \} & \text{if }  j > i \: .
	\end{cases}
\end{equation}
By using dynamic programming, we can compute $\bm{\mu}^{(i)}(t)$ from $\bm{\mu}^{(i-1)}(t)$ and check $\bm{\mu}^{(i)}(t) \in \cS$ in constant time.
Therefore, the above procedure computes $\bm{\mu}^+(t) = \argmax_{ \cS \cap \{ \bm{\mu}^{(i)}(t) \}_{i \in [K]}} T^\star(\bm{\mu}^{(i)}(t))^{-1}$ with a computational cost of $\mathcal{O}(K)$.
It uses at most $K$ calls to $T^\star(\cdot)^{-1}$.
We refer the reader to Appendix~\ref{subsec-app:O-TaS-impl} for a proof and more details on the implementation.

% Since $\bm{\mu}^+(t) = \argmax_{ \{ \bm{\mu}^{(i)}(t) \}_{i=1}^K} T^\star(\bm{\mu}^{(i)}(t))^{-1}$. Applying this simple procedure, would result in $\mathcal{O}(K^2)$ procedure for computing $\bm{\mu}^+(t)$ (i.e., constructing $K$ optimistc models, where each construction requires $\mathcal{O}(K)$ operations). Nevertheless, it is possible to compute $\bm{\mu}^{(i)}(t)$ from $\bm{\mu}^{(i-1)}(t)$ in constant time. In Appendix \ref{subsec-app:O-TaS-impl}, we provide details and a correctness analysis on this algorithm.\footnote{As a technical note, we observe that in principle, $\bm{\mu}^{(i)}$ might violate the unimodal constraints, and whenever this happens, we want to exclude such arms from the argmax. In Appendix \ref{subsec-app:O-TaS-impl} we discuss how to check if $\bm{\mu}^{(i)}$ satisfy the unimodal constraint in constant time using dynamic programming.}

% Overall, the computational complexity of O-TaS is thus $\mathcal{O}( K )$. Finally, we remark that O-TaS requires $K$ times access to $T^*(\cdot)^{-1}$, while TaS only $1$. 

\subsection{Top Two Algorithm}
\label{ssec:UniTT}

Top Two algorithms have been studied for BAI~\citep{russo2016simple,qin2017improving,shang2020fixed,jourdan22toptwo}.
However, few Top Two algorithms have been analyzed in structured settings.
By adapting TTUCB~\citep{jourdan2023NonAsymptoticAnalysis}, we propose the UniTT (Unimodal Top Two) algorithm for Unimodal BAI, which is computationally cheaper than U-TaS and O-TaS.
We show that UniTT has near matching guarantees compared to Theorems~\ref{prop:lb} and~\ref{thm:lb_K}.

After pulling each arm once, a Top Two sampling rule for Unimodal BAI defines two candidate answers, a leader $B_t \in [K]$ and a challenger $C_t \in \cN(B_t)$.
It samples the next arm $I_t \in \{B_t, C_t\}$ to verify that the leader is better than the challenger.
This choice is specified by (i) a target proportion $\beta_t(B_t,C_t) \in (0,1)$, representing the fraction of time where the leader should be sampled given this challenger arm, and (ii) a mechanism to reach it, e.g. randomization or tracking.

% \begin{algorithm}[t]
% 	\caption{\protectUniTT}
%     \label{algo:UniTT}
% 	\begin{algorithmic}
% 	 	\STATE {\bfseries Input:} $\beta \in (0,1)$, e.g. $\beta = 1/2$, $\cS$ as in~\eqref{eq:structure_Unimodal}, time $t > K$, $\Theta_t$ as in~\eqref{eq:unstructured_CI}, $W_t(i,j)$ as in~\eqref{eq:TC_empirical}.
% 	 	\STATE {\bfseries Output:} Next arm to sample $I_t$.
% 	    \STATE Get UCB leader $B_{t} \in \argmax_{i \in [K]} \max_{\boldsymbol{\lambda} \in \Theta_t \cap \cS}  \lambda_i$ \:;
% 	    \STATE Get challenger $C_{t} \in \argmin_{j \in \cN(B_{t})} W_{t}(B_t,j)$ \:;
% 	    \STATE {\bfseries Return} 
% 	    $I_{t} =\begin{cases}
% 	    	 C_{t} &\text{if } N_{C_{t}}^{B_{t}}(t) \le (1 - \beta) T_{t+1}(B_{t}, C_{t})\\
% 	    	 B_{t} &\text{otherwise.}
% 	    	 \end{cases}$
% 	\end{algorithmic}
% \end{algorithm}

\paragraph{Leader-Challenger Pair}
Using optimism with the unimodal structure, the UCB leader is defined as
\begin{equation} \label{eq:UCB_leader}
	B_{t} \in \argmax_{i \in [K]} \max_{\boldsymbol{\lambda} \in \Theta_t \cap \cS}  \lambda_i \quad \text{with $\Theta_t$ as in~\eqref{eq:unstructured_CI}} \: ,
\end{equation}
where ties are broken uniformly at random, hence $B_{t} \sim \cU([K])$ when $\Theta_t \cap \cS = \emptyset$.
To account for the local structure of Unimodal BAI, we use the challenger
\begin{equation} \label{eq:localTCchallenger}
	C_{t} \in \argmin_{j \in \cN(B_{t})} W_{t}(B_{t}, j) \quad \text{with $W_t(i,j)$ as in~\eqref{eq:TC_empirical}}\: .
\end{equation}
This challenger is linked to the GLR stopping rule~\eqref{eq:GLR_stopping_rule}.
When $\hat \imath_t = B_t$, not stopping implies that $W_{t}(B_{t}, C_t) < c(t-1,\delta)$, hence we should pull $I_t \in \{B_t, C_t\}$.

\paragraph{Fixed Tracking}
Let $\beta \in (0,1)$, e.g. $\beta = 1/2$.
We consider the fixed design $\beta_t(B_t,C_t) = \beta $ and use $2(K-1)$ independent tracking procedures, i.e. one for each pair of leader $i$ and challenger $j \in \cN(i)$.
We pull
\begin{equation} \label{eq:beta_tracking}
	I_{t} =\begin{cases}
	    	 C_{t} &\text{if } N_{C_{t}}^{B_{t}}(t) \le (1 - \beta) T_{t+1}(B_{t}, C_{t}) \: ,\\
	    	 B_{t} &\text{otherwise} \: ,
	    \end{cases}
\end{equation}
where $N^{i}_{j}(t) \eqdef \sum_{s \in [t-1]} \indi{(B_s, C_s) = (i,j), \: I_{s} = j}$ and $T_{t}(i,j) \eqdef \sum_{s \in [t-1]} \indi{(B_s, C_s) = (i,j)}$. 

\paragraph{Efficient Implementation}
Top Two algorithms enjoy one of the lowest computational cost among BAI methods.
We provide an efficient implementation of the UCB leader which is the computational bottleneck of UniTT.
As for O-TaS, we use dynamic programming to find the optimal optimistic instance $\bm{\mu}^{(i)}(t)$ in~\eqref{eq:effSAO} related to $i \in [K]$, and exclude the non-unimodal ones. 
Thus, this procedure computes $B_t = \argmax_{i \mid \: {\mu}^{(i)}_i(t) \in \cS} {\mu}^{(i)}_i(t)$ in $\cO(K)$. In Appendix~\ref{subsec-app:unitt-eff-impl} we provide proof and implementation details.
Contrary to TaS and O-TaS, UniTT never calls $T^\star(\cdot)^{-1}$.

% To this end, one can exploit~\eqref{eq:effSAO} together with dynamic programming in order to construct $\{ \bm{\mu}^{(i)}(t) \}_{i=1}^K$. 
% Then, $B_t$ can be computed as $B_t = \argmax_{i \in [K]} {\mu}^{(i)}_i(t)$.
% Overall, the total complexity of this procedure is $\mathcal{O}(K)$.\footnote{As for O-TaS, there are cases in which $\bm{\mu}^{(i)}$ might violate the unimodal constraints. Whenever this happens, we need to exclude such arms from the argmax, and, as for O-TaS this can be done without altering the computational complexity of the algorithm. Full details on the efficient implementation of UniTT are deferred to Appendix \ref{subsec-app:unitt-eff-impl}.} Furthermore, contrary to TaS and O-TaS, UniTT never requires access to $T^*(\cdot)^{-1}$. 

\subsubsection{Non-asymptotic Guarantees}
\label{sssec:UniTT_non_asymptotic}

While UniTT is defined for general classes of distributions, we analyze it for Gaussian distributions with unit variance.
Theorem~\ref{thm:non_asymptotic_upper_bound_UniTT} gives an upper bound on the expected sample complexity holding for any $\delta$ and any Gaussian instances having mean $\boldsymbol{\mu} \in \cS$. 
It yields non-asymptotic guarantees and recover \emph{asymptotic $1/2$-optimality}, defined as satisfying $\limsup_{\delta \to 0 } \bE_{\boldsymbol{\nu}}[\tau_{\delta}]/\log(1/\delta) \le T^{\star}_{1/2}(\bm{\mu})$ where 
\begin{equation} \label{eq:beta_characteristic_time}	
	2T^{\star}_{\beta}(\bm{\mu})^{-1} \eqdef \underset{\bm \omega : \frac{\bm \omega}{1-\beta} \in \triangle_{|\cN(i^\star)|}}{\sup}  \min_{i \in \mathcal{N}(i^\star)} \frac{(\mu_{i^\star}- \mu_{i})^2}{1/\beta + 1/\omega_i} \: ,
\end{equation}
is achieved at the $\beta$-optimal allocation $\bm \omega^{\star}_{\beta}$.
It is a corollary of a result for any $\beta \in (0,1)$ and $\gamma_{1},\gamma_{2} > 1$.
\begin{theorem} \label{thm:non_asymptotic_upper_bound_UniTT}
Using the GLR stopping rule as in \eqref{eq:GLR_stopping_rule}, UniTT with $\beta=1/2$ and $f_{u}/2$ with $\gamma_{1} = \gamma_{2} = 1.2$ in~\eqref{eq:unstructured_CI} is $\delta$-correct and, for all Gaussian instances with unit variance and mean $\boldsymbol{\mu} \in \cS$, $\bE_{\boldsymbol{\nu}}[\tau_{\delta}] \le T_{\boldsymbol{\mu}}(\delta) + 15K$ with
\[
	\limsup_{\delta \to 0 } \frac{T_{\boldsymbol{\mu}}(\delta)}{\log(1/\delta)} \le \frac{T^{\star}_{1/2}(\bm{\mu})}{1 - \indi{|\cN(i^\star)|=2} \gamma/a_{\boldsymbol{\mu}}} \: ,
\]
for any $\gamma \in (0, a_{\boldsymbol{\mu}})$ with $a_{\boldsymbol{\mu}} \eqdef \min_{i \in \cN(i^\star)}\omega^{\star}_{1/2,i}$.
Define $H_{1}(\boldsymbol{\mu}) \eqdef \sum_{i \ne i^\star}\Delta_{i}^{-2}$ and $\Delta_{\min} \eqdef  \min_{i} \Delta_{i}$ with $\Delta_{i}  \eqdef \mu_{i^\star} - \mu_{i}$.
Then, for any $\epsilon \in (0,1]$, we have
\begin{align*}
	&T_{\boldsymbol{\mu}}(\delta) \le \max\{T_{0}(\delta), C_{\boldsymbol{\mu}, |\cN(i^\star)|}\} \text{ with } \\
	&T_{0}(\delta) \eqdef \sup \{ t \mid t  \le \frac{(1+\epsilon)T^{\star}_{1/2}(\bm{\mu})}{1 - \indi{|\cN(i^\star)|=2} \gamma/a_{\boldsymbol{\mu}}} \\
	&\qquad \qquad \qquad \qquad  ( \sqrt{c(t-1,\delta)} + \sqrt{3.2  \log t} )^2 \} \: , \\
	&C_{\boldsymbol{\mu},1} \eqdef \max\{h_{1}\left(338  H_{1}(\boldsymbol{\mu}) / \epsilon, 48K/\epsilon \right),(4/\epsilon)^{6}\} \: ,\\
	&C_{\boldsymbol{\mu},2} \eqdef (6D_{\boldsymbol{\mu}}/(a_{\boldsymbol{\mu}} - \gamma) + 2)/\epsilon \: ,  \\
	& D_{\boldsymbol{\mu}} \eqdef h_{1} \left( 13   T^{\star}_{1/2}(\bm{\mu}) / \gamma^2, (L_{\boldsymbol{\mu}}+ 4)r (\gamma) /(a_{\boldsymbol{\mu}}-\gamma) \right) \: , \\
	&L_{\boldsymbol{\mu}} \eqdef \max\{ h_{1}\left(96 \left(H_{1}(\boldsymbol{\mu}) + 34\Delta_{\min}^{-2} \right), 26K  \right) , (2/\epsilon)^{6}\} \: , 
\end{align*}
where $h_{1}$ as in Lemma~\ref{lem:inversion_upper_bound} and $r(x) = 1+2(\sqrt{1+2x}-1)^{-1}$. We have $h_{1}(z,\cdot) \underset{+ \infty}{\approx} z \log z$ and $h_{1}(\cdot,y) \underset{+ \infty}{\approx} y$.

Moreover, on instances such that $|\cN(i^\star)|=2$, UniTT satisfies that $\bE_{\boldsymbol{\nu}}[\tau_{\delta}] \le \widetilde T_{\boldsymbol{\mu}}(\delta) + 15K$ with
\[
	\limsup_{\delta \to 0 } \frac{\widetilde T_{\boldsymbol{\mu}}(\delta)}{\log \frac{1}{\delta}} \le 2 T^{\star}_{1/2}(\bm{\mu}) \text{ , } \widetilde T_{\boldsymbol{\mu}}(\delta) \le \max\{\widetilde T_{0}(\delta), C_{\boldsymbol{\mu},1}\}
\]
and $\widetilde T_{0}(\delta) = \sup \{ t \mid t  \le 2(1+\epsilon)T^{\star}_{1/2}(\bm{\mu}) $ $( \sqrt{c(t-1,\delta)} + \sqrt{3.2 \log t} )^2 \}$.
\end{theorem}

\paragraph{Asymptotic $1/2$-Optimality} 
Our asymptotic upper bound is $T^{\star}_{1/2}(\bm{\mu})$ (take $\gamma \to 0$ when $|\cN(i^\star)| = 2$), hence UniTT is asymptotically $1/2$-optimal.
When $|\cN(i^\star)| = 1$, UniTT is asymptotically optimal since $T^{\star}(\bm{\mu}) = T^{\star}_{1/2}(\bm{\mu})$ (see Appendix~\ref{app:ssec_Gaussian_characteristic_times}).
When $|\cN(i^\star)| = 2$, the ratio $T^\star_{1/2}(\bm{\mu}) / T^{\star}(\bm{\mu})$ lies numerically in $(1,r_2]$ with $r_{2} \approx 1.03$, which is achieved when neighboring arms $\cN(i^\star)$ have the same mean (see Figure~\ref{fig:simulation_ratio_times}).

\paragraph{Non-asymptotic Dependency}
The upper bound $\max\{T_{0}(\delta), C_{\boldsymbol{\mu}, |\cN(i^\star)|}\} + 15K$ has a linear dependency in $K$.
Since large $\epsilon \in (0,1]$ or $\gamma \in (0,a_{\boldsymbol{\mu}})$ trades off smaller $C_{\boldsymbol{\mu},|\cN(i^\star)|}$ for larger $T_{0}(\delta)$, we take $\epsilon=1$ and $\gamma=a_{\boldsymbol{\mu}}/2$ in the following.
Then, we obtain $C_{\boldsymbol{\mu},1} =\cO(\max\{K, H_{1}(\boldsymbol{\mu}) \log H_{1}(\boldsymbol{\mu})\})$ and $C_{\boldsymbol{\mu},2} = \cO(\max\{K, H_{1}(\boldsymbol{\mu}) \log H_{1}(\boldsymbol{\mu})\}/a_{\boldsymbol{\mu}}^3)$ since $T^{\star}_{1/2}(\bm{\mu}) = \cO(H_{1}(\boldsymbol{\mu}))$ and $r(x) \approx_{x \to 0} 2/x$.
While $a_{\boldsymbol{\mu}} = 1/4$ when neighboring arms $\cN(i^\star)$ have the same mean, $1/a_{\boldsymbol{\mu}}$ diverge to infinity when one neighbor has a negligible gap compared to the other one.
To alleviate paying the factor $1/a_{\boldsymbol{\mu}}$, the second part of Theorem~\ref{thm:non_asymptotic_upper_bound_UniTT} gives a non-asymptotic upper bound scaling as $C_{\boldsymbol{\mu},1} $ at the cost of a sub-optimal multiplicative constant $2$ in the definition of $\widetilde T_{0}(\delta)$.
On the hard instance of Theorem~\ref{thm:lb_K} which satisfies $H_{1}(\boldsymbol{\mu}) = (K-1)/\Delta^2$, we obtain that $C_{\boldsymbol{\mu},1} \approx 48K + 338K\Delta^{-2} \log(K/\Delta^2) + \mathcal{O}(K\Delta^{-2} \log (\Delta^{2} + \log (K\Delta^{-2} )))$. 
Therefore, up to a logarithmic factor, our non-asymptotic upper bound matches the lower bound of Theorem~\ref{thm:lb_K}.

% To understand the range of $a_{\boldsymbol{\mu}}$, we consider two extreme cases.
% First, when neighboring arms $\cN(i^\star)$ have the same mean, we have $a_{\boldsymbol{\mu}} = 1/4$ which is its largest possible value.
% Recall that those instances are the ones for which $T^\star_{1/2}(\bm{\mu}) / T^{\star}(\bm{\mu})$ is the largest ($\approx 1.03$).
% Let $j^\star \in \argmax_{i \in \cN(i^\star)} \mu_i$ and $j \in \cN(i^\star)\setminus \{j^\star\}$.
% Second, we consider instances such that $x = (\mu_{i^\star} - \mu_{j})/(\mu_{i^\star} - \mu_{j^\star}) \to + \infty$.
% Recall that those instances are the ones for which $T^\star_{1/2}(\bm{\mu}) / T^{\star}(\bm{\mu})$ is the smallest ($=_{x \to + \infty} 1$).
% Using the equality at equilibrium, one can show that $a_{\boldsymbol{\mu}} \approx_{x \to + \infty} 1/(4x^2)$.
% Therefore, $1/a_{\boldsymbol{\mu}}$ will be large, hence it yields a sub-optimal dependency.
% To alleviate paying the factor $1/a_{\boldsymbol{\mu}}$ when it is large, we derive another non-asymptotic upper bound when $|\cN(i^\star)|=2$.
% At the cost of a sub-optimal multiplicative constant $2$ in the definition of $\widetilde T_{0}(\delta)$, we recover the non-asymptotic upper bound of the case where $|\cN(i^\star)|=1$, i.e. $\bE_{\boldsymbol{\nu}}[\tau_{\delta}] \le \max\{\widetilde T_{0}(\delta), C_{\boldsymbol{\mu}, 1}\} + 15K$ where $ C_{\boldsymbol{\mu}, 1} = \cO(\max\{K, H_{1}(\boldsymbol{\mu}) \log H_{1}(\boldsymbol{\mu})\})$.

\paragraph{Optimal Design IDS}
	To reach asymptotic optimality, UniTT should be used with the optimal design IDS~\citep{you2023information}, namely by taking $\beta_{t}(B_{t},C_{t}) = N_{C_{t}}(t)/(N_{B_{t}}(t)+ N_{C_{t}}(t))$ and tracking its cumulative sum instead of $\beta T_{t+1}(B_t,C_t)$.
	Asymptotically, Unimodal BAI is equivalent to BAI on $\cN(i^\star) \cup \{i^\star\}$, and UniTT is equivalent to TTUCB on $\cN(i^\star) \cup \{i^\star\}$.
	Therefore, UniTT with IDS will be asymptotically optimal by adapting existing proof~\citep{jourdan2024SolvingPureExploration}.
	We leave the details to the reader for two reasons.
	First, our focus is on obtaining non-asymptotic guarantees, yet this is challenging with IDS.
	Second, asymptotic $1/2$-optimality is close to optimality for Unimodal BAI. 
	When $|\cN(i^\star)|=1$ it is the same, see above.

% \paragraph{Beyond Gaussian Distributions}
% 	Theorem~\ref{thm:non_asymptotic_upper_bound_UniTT} holds for $1$-sub-Gaussian random variable and could be generalized for known heterogenous variances.
% 	Obtaining non-asymptotic guarantees for general distributions will come at the price of more technical arguments and less explicit non-asymptotic terms.

\begin{figure*}[t]
\centering
  \includegraphics[width=0.38\textwidth]{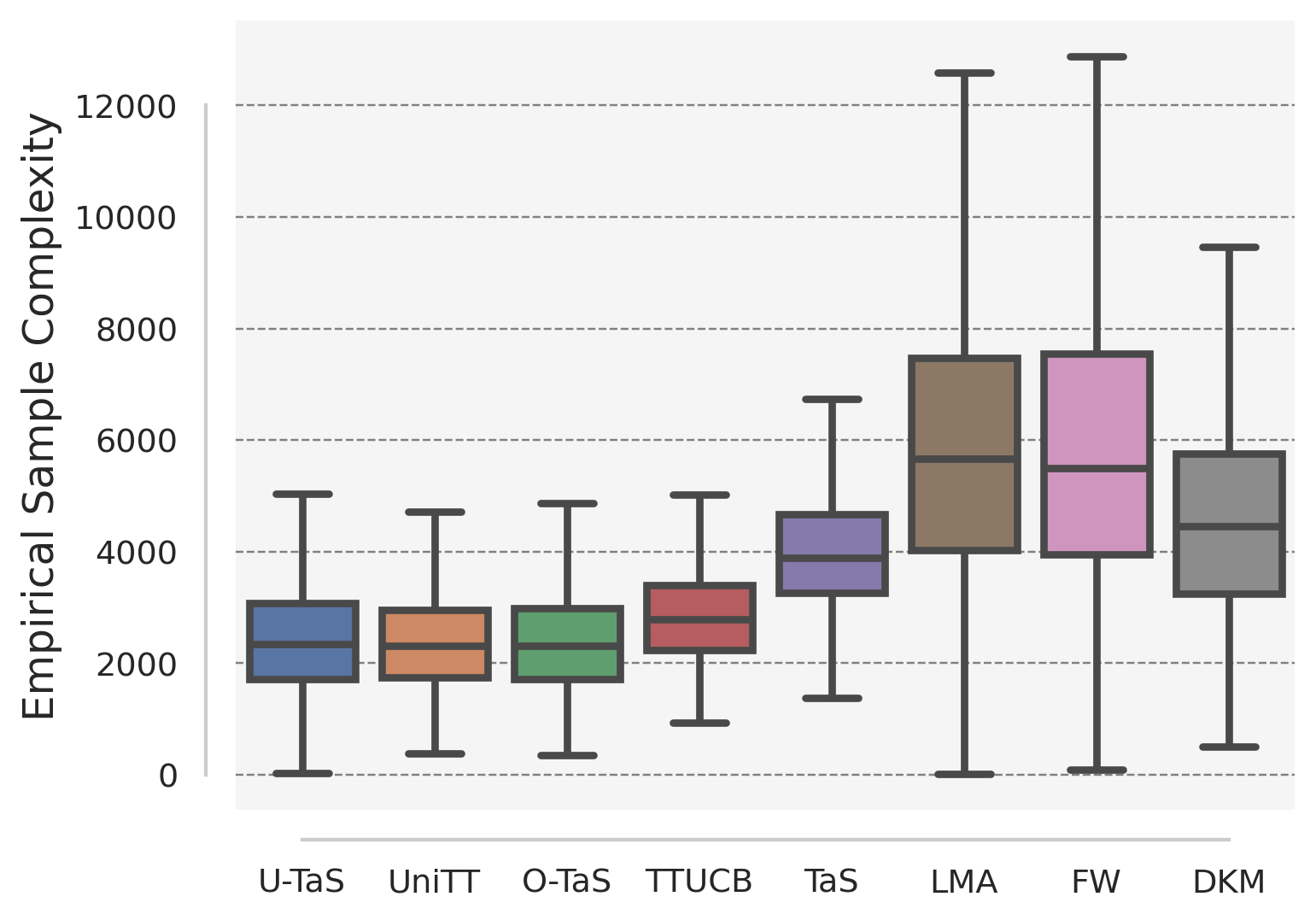}
  \hspace{0.2cm}
  \includegraphics[width=0.38\textwidth]{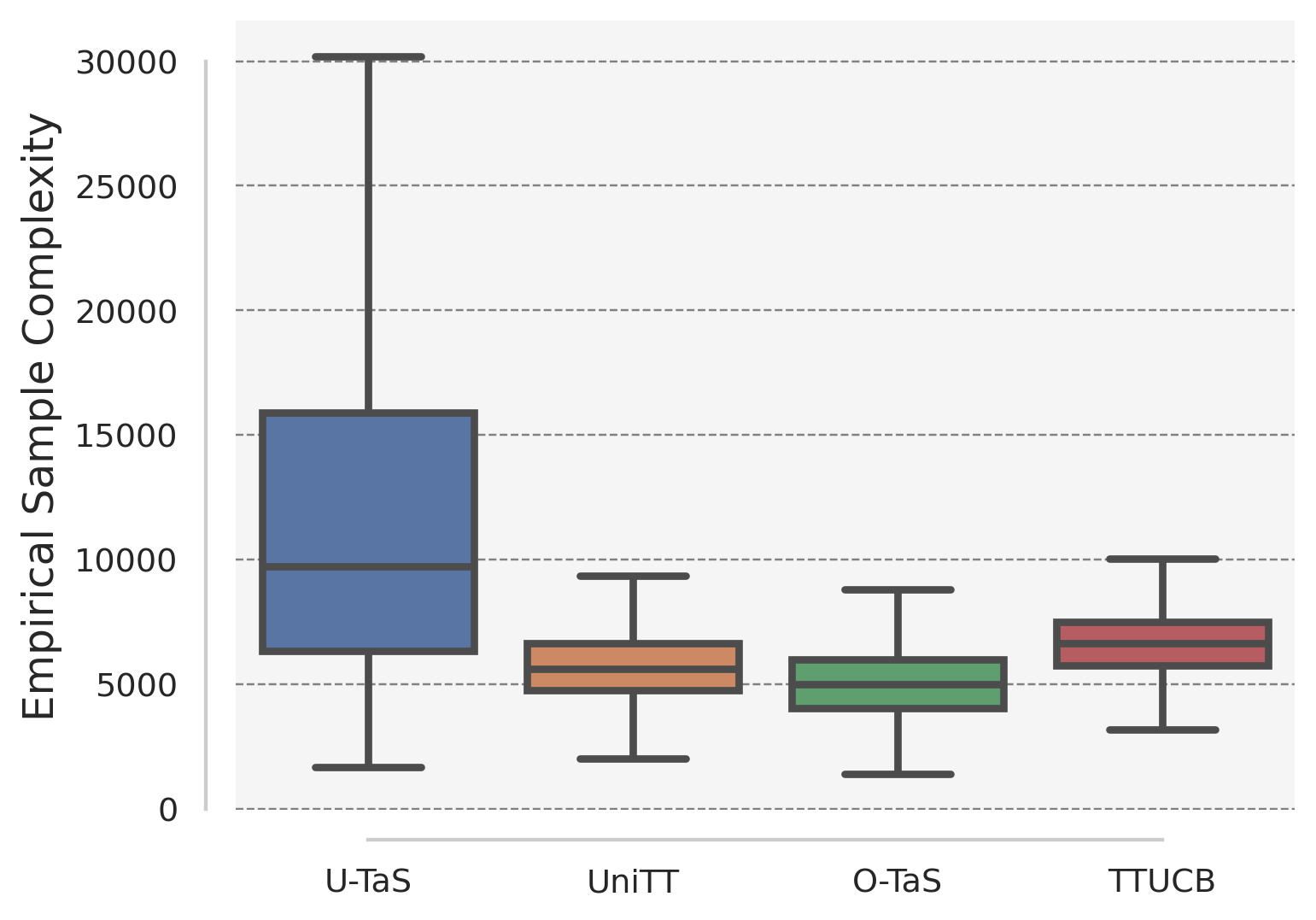}\\
  \includegraphics[width=0.38\textwidth]{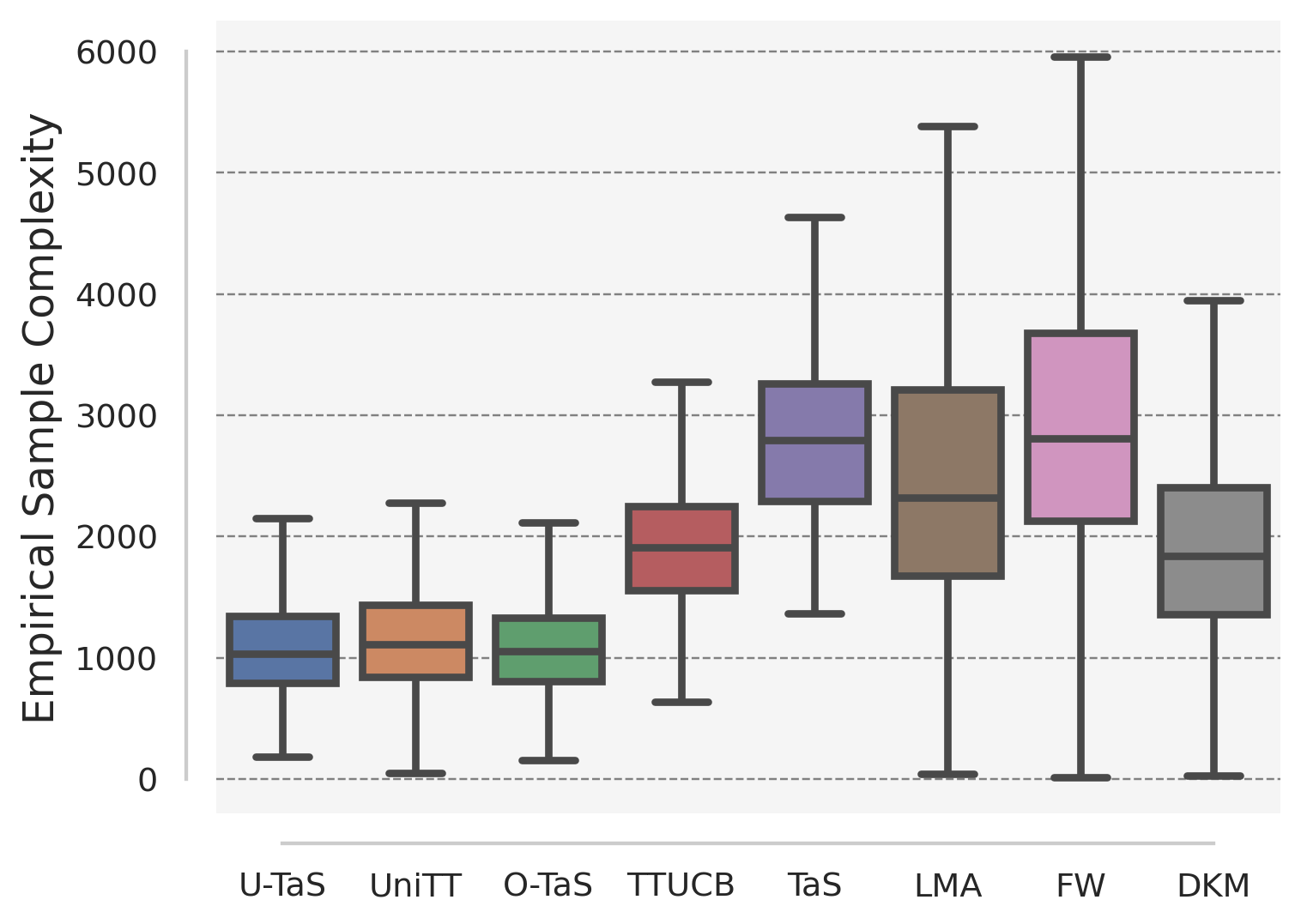}
  \hspace{0.2cm}
  \includegraphics[width=0.38\textwidth]{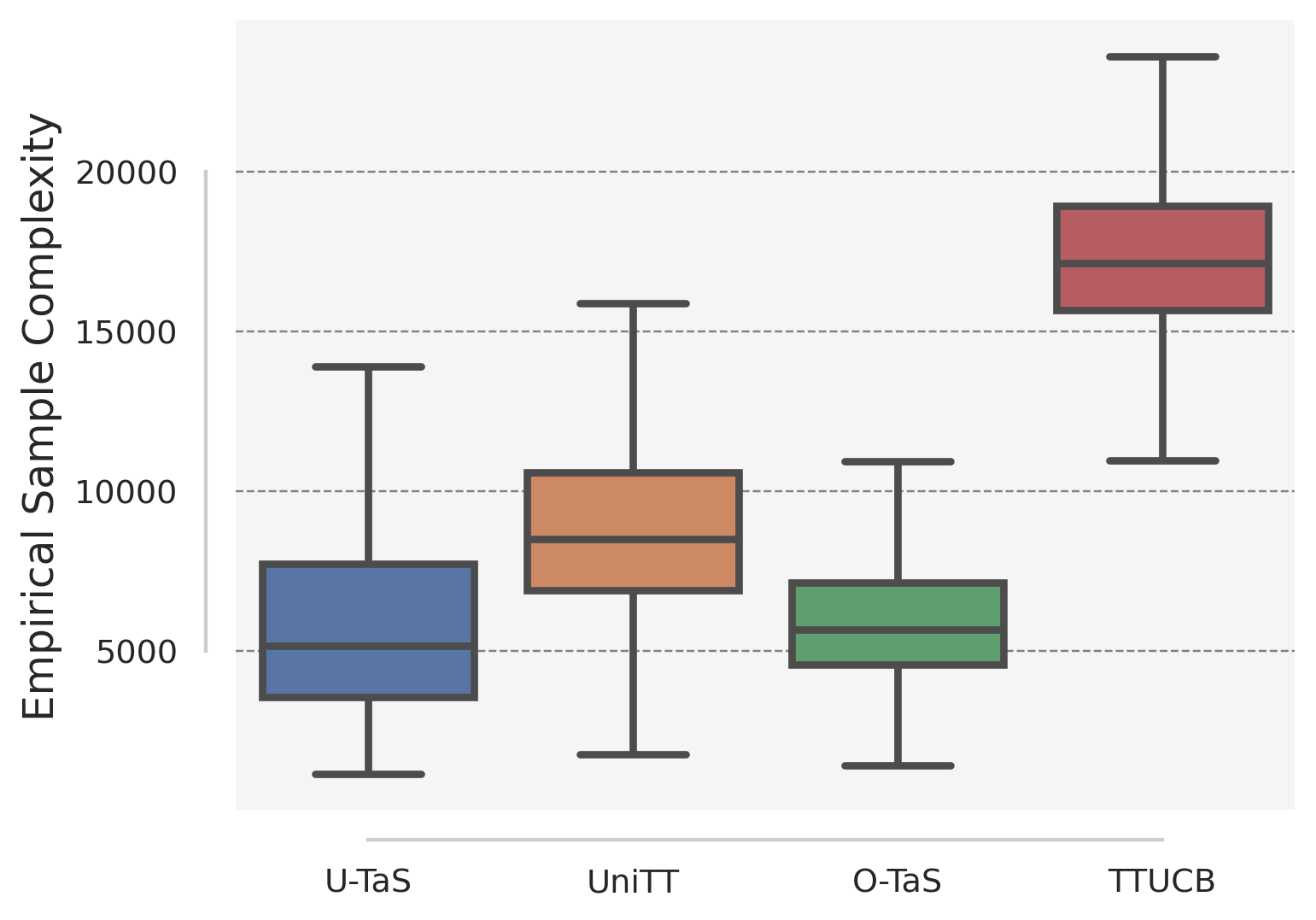}
  \caption{Empirical stopping time ($\delta = 0.01$) on (top) random instances with $K \in \{10,100\}$, denoted as $\bm{\mu}_{\textup{R}, 10}$ and $\bm{\mu}_{\textup{R}, 100}$, and (bottom) flat instances with $K \in (11,101)$, denoted as $\bm{\mu}_{\textup{F}, 11}$ and $\bm{\mu}_{\textup{F}, 101}$.}
  \label{fig:mu-all}
\end{figure*}

\subsubsection{Proof Sketch of Theorem~\ref{thm:non_asymptotic_upper_bound_UniTT}}
\label{sssec:proof_sketch}

We sketch the proof Theorem~\ref{thm:non_asymptotic_upper_bound_UniTT} below and highlight its novelties, see Appendix~\ref{app:UniTT} for details.
Under $\cE_t$, we prove that $B_t = i^\star$ except for a sublinear number of times (Lemma~\ref{lem:ucb_leader_lower_bound_counts}).
By definition of the stopping rule~\eqref{eq:GLR_stopping_rule} with $\hat \imath_t$ as in~\eqref{eq:IFanswer}, not stopping implies that
\[
	c(t-1,\delta) \ge \max_{i \in [K]} \min_{j \in \cN(i)} W_{t}(i,j) \ge W_{t}(B_t,C_t) \: ,
\]
where we use the challenger in~\eqref{eq:localTCchallenger}.
Under suitable concentration and for $B_t = i^\star$, using $W_t$ in~\eqref{eq:Gaussian_TC} yields
\begin{equation} \label{eq:ccl_step}
	c(t-1,\delta) \ge \frac{2 + 1/\omega^{\star}_{1/2,C_t}}{\frac{t}{N_{i^\star}^{i^\star}(t)}+ \frac{t}{N_{C_t}^{i^\star}(t)}} \frac{t}{T^\star_{1/2}(\bm \mu)} - \cO(\log t) \: .
\end{equation}
Using tracking (Lemma~\ref{lem:tracking_guaranties}), we have $N_{i^\star}^{i^\star}(t) \approx t/2$.
When $|\cN(i^\star)|=1$, we have $N_{C_t}^{i^\star}(t) \approx t/2$ as well.
Re-ordering terms in~\eqref{eq:ccl_step} yields a $(t,\delta,\bm \mu)$-dependent condition.
As $T_{\bm \mu}(\delta)$ is defined as the last time this condition can hold, it yields our first upper bound.
When $|\cN(i^\star)|=2$, we prove that the ratio in the r.h.s. of~\eqref{eq:ccl_step} is larger than $1$, e.g. by showing that $N_{C_t}^{i^\star}(t) \gtrsim t \omega^{\star}_{1/2,C_t}$.
We consider two different approaches.

\paragraph{Coarse Analysis}
The coarse analysis is an improvement on~\citet{jourdan2023NonAsymptoticAnalysis} that yields the last upper bound in Theorem~\ref{thm:non_asymptotic_upper_bound_UniTT}.
While the expected sample complexity of TTUCB is in $\cO(\max\{K,H(\bm \mu) \log H(\bm \mu)\}^{\gamma_{1}})$ where $\gamma_{1} > 1$, our bound is in $\cO(\max\{K, H_{1}(\boldsymbol{\mu}) \log H_{1}(\boldsymbol{\mu})\})$, hence shaving this sub-optimal dependency in $\gamma_{1}$. 

Refining their pigeonhole argument, we show that there exists an arm $i_0 \in \cN(i^\star)$ being over-sampled compared to $t \omega_{i_0}$ where $\omega_{i_0} \ge \max \{ \frac{1}{4},\omega^{\star}_{1/2,i_0}\}$. 
Let $n_t$ be the last time with $(B_{n_t},C_{n_t}) = (i^\star, i_0)$.
When $n_t \ge t^{1/\gamma_{1}}$, we show that $N_{i_0}^{i^\star}(n_t)$ is larger than $t \omega_{i_0}$ (similarly as them).
When $n_t < t^{1/\gamma_{1}}$, for all $s \in [t^{1/\gamma_{1}},t]$ where $B_s = i^\star$, the arm $j_0 \in \cN(i^\star)\setminus\{i_0\}$ is the challenger.
Let $\tilde n_t$ be the last time with $B_{\tilde n_t} = i^\star$.
Then, we show that $N_{j_0}^{i^\star}(\tilde n_t) \approx t/2$.
In both cases, we can re-order terms in~\eqref{eq:ccl_step} and define $\widetilde T_{\bm \mu}(\delta)$ as the last time this condition can hold.
This concludes the proof.

The refined pigeonhole argument and the dichotomy are the technical novelties yielding the improved non-asymptotic terms.
The coarse analysis sacrifices a multiplicative constant $2$ in the definition of $\widetilde T_0(\delta)$. 
Below, we introduce an analysis recovering asymptotic $1/2$-optimality, with $\cO(1/a_{\bm \mu}^3)$ as non-asymptotic cost.

\paragraph{Tight Analysis}
We develop a novel non-asymptotic analysis using $\gamma \in (0,a_{\boldsymbol{\mu}})$ to control the gap between $N^{i^\star}_{i}(t)/t$ and $\omega^{\star}_{1/2,i}$ for any neighboring arm $i \in \cN(i^\star)$.
As technical tool, we use that \emph{the number of times one can increment a bounded positive variable by one is also bounded}, see Lemma~\ref{lem:simple_key_observation} by~\citet{jourdan2023epsilonbestarm}.

We exhibit a time $L_{\bm \mu}$ after which the UCB leader will always be the best arm provided concentration holds (Lemma~\ref{lem:time_leader_correct}), hence $N^{i^\star}_{i^\star}(t) \approx t/2$.
This also implies that UniTT exploits the sparsity pattern and satisfy that $I_{t} \in \cN(i^\star) \cup \{i^\star\}$ for all $t > L_{\bm \mu}$ (similarly to O-TaS).
Then, $D_{\bm \mu}$ is a time after which the empirical allocations of the neighbors $\cN(i^\star)$ stay $\gamma$-close to their $1/2$-optimal allocation (Lemma~\ref{lem:time_neighbors_controlled}), hence $N^{i^\star}_{i}(t) \approx t \omega^{\star}_{1/2,i}$ for all $i \in \cN(i^\star)$.
By re-ordering terms in~\eqref{eq:ccl_step}, we define $T_{\bm \mu}(\delta)$ similarly as before to conclude.

% !TeX root = ../paper.tex

\section{EXPERIMENTS} \label{sec:expe}

% \begin{figure*}[t]
% \centering
% \begin{minipage}{.4\textwidth}
%   \centering
%   \includegraphics[width=\textwidth]{plots/rnd10/plot.png}
%   \caption{Empirical stopping time on $\bm{\mu}_{\textup{R}, 10}$.}
%   \label{fig:mu-r-10}
% \end{minipage}%
% \hspace{.55cm} % Adjust the horizontal space between figures here
% \begin{minipage}{.4\textwidth}
%   \centering
%   \includegraphics[width=\textwidth]{plots/flat11/plot.png}
%   \caption{Empirical stopping time on $\bm{\mu}_{\textup{F}, 11}$.}
%   \label{fig:mu-f-11}
% \end{minipage}
% %\hfill
% \end{figure*}

% \begin{figure*}[t]
% \centering
% \begin{minipage}{.4\textwidth}
%   \centering
%   \includegraphics[width=\textwidth]{plots/rnd100/plot.png}
%   \caption{Empirical stopping time on $\bm{\mu}_{\textup{R}, 100}$.}
%   \label{fig:mu-r-100}
% \end{minipage}%
% \hspace{.55cm} % Adjust the horizontal space between figures here
% \begin{minipage}{.4\textwidth}
%   \centering
%   \includegraphics[width=\textwidth]{plots/flat101/plot.png}
%   \caption{Empirical stopping time on $\bm{\mu}_{\textup{F}, 101}$.}
%   \label{fig:mu-f-101}
% \end{minipage}
% %\hfill
% \end{figure*}

We present experiments to (i) compare our three proposed algorithms, (ii) highlight the pitfalls of other asymptotic optimal algorithms such as DKM \citep{degenne2019non}, LMA \citep{menard2019gradient} or FW \citep{wang21FW}), and (iii) showcase the benefits of exploiting the unimodal structure by comparing ourselves to unstructured TaS and TTUCB. We also compare theoretically all these choiches in Appendix~\ref{app:algo-comparison}.
We report the boxplots of the stopping times for $1000$ independent runs using $\delta=0.01$.
We use the heuristic $c_K(t, \delta) = \log( K \log(e t) / \delta ) $ and observe empirical errors lower than $\delta$.
Appendix~\ref{app:add_expe} provides additional experiments.

% We test all these algorithms in the $4$ different domains described below. In all cases, we set $\delta=0.01$ and report the empirical sample complexity over $1000$ independent runs. Additional details, and further experiments are deferred to Appendix \ref{app:add_expe}.

\paragraph{Instances} 
As Bayesian benchmark, we randomly generate unimodal Gaussian bandits with unit variance for $K \in \{10,100\}$.
First, we draw the optimal arm $i \sim \cU([K])$ and set $\mu_{i} = 1$.
Then, for all $j \ne i$, we draw $x_{j} \sim \cU([0.2,0.3])$ and set $\mu_{j} = \mu_{j+1} - x_j$ if $j < i$, and $\mu_{j} = \mu_{j-1} - x_j$ otherwise. 
% First, we test all baselines on randomly generated unimodal Gaussian bandits with unitary variance. Precisely, we first generate the optimal arm, and then we generate the mean of its neighborhood by sampling a small positive gap to be subtracted to the mean of the optimal arm. We repeat this step until we have generated the values for all $K$ arms. Further details on this generation procedure are deferred to Appendix \ref{app:add_expe}. We will present results of two random instances with $K=10$ and $K=100$, which we will refer to as $\bm{\mu}_{\textup{R}, 10}$ and $\bm{\mu}_{\textup{R}, 100}$ respectively. 
As hard instances from Theorem \ref{thm:lb_K}, we consider the flat unimodal Gaussian bandits with unit variance for $K \in \{11,101\}$.
We use ${\star} = \lceil K/2 \rceil$, $\mu_{\star} = 1.0$ and $\mu_i = 0.6$ for all $i \ne \star$.
% Specifically, we considered Gaussian bandits with unitary variance where $\mu_{\star} = 1.0$ and $\mu_i = 0.6$ for all $i \ne \star$. We will present the results for $K=11$ and $K=501$. We will refer to these bandit models to as $\bm{\mu}_{\textup{F}, 11}$ and $\bm{\mu}_{\textup{F}, 101}$ respectively.
%  In both cases, we centered the optimal arm, i.e., ${\star} = \lfloor \frac{K}{2} \rfloor + 1$. 

\paragraph{Results} 
In Figure~\ref{fig:mu-all}, on $\bm{\mu}_{\textup{R}, 10}$ and $\bm{\mu}_{\textup{F}, 11}$, we see U-TaS, O-TaS, and UniTT have the best performance.
Moreover, DKM, FW, and LMA perform poorly, even worse than unstructured BAI algorithms like TaS and TTUCB. 
Both TaS and TTUCB have worse performance in $\bm{\mu}_{\textup{F}, 11}$, which is also the hard instance in unstructured BAI as all arms are near optimal. 

For larger $K$, we use U-TaS, O-TaS, UniTT and TTUCB due to their improved empirical performance for smaller $K$ and faster implementations.
In Figure~\ref{fig:mu-all}, on $\bm{\mu}_{\textup{R}, 100}$, UniTT and O-TaS are the most competitive methods, while U-TaS performs worse than the unstructured TTUCB algorithm.
We attribute the poor results of U-TaS to the forced exploration mechanism which is wasteful when arms are highly sub-optimal for large $K$.
In constrast, on $\bm{\mu}_{F,101}$, the forced exploration in U-TaS yields performance comparable to O-TaS for instances where sub-optimal arms have similar means and should be explored as much, see $\bm{\mu}_{F,101}$.
As before, the gap between TTUCB and UniTT or O-TaS is larger for flat instances. 
%Appendix~\ref{app:add_expe} contains an ablation study on this phenomena which amplifies with larger $K$.

\section{PERSPECTIVES}
\label{sec:conclusion}

% Summary paper
We studied the unimodal structure in fixed-confidence BAI.
The derived lower bounds suggest that sparse optimal allocations are sufficient asymptotically, yet dense exploration is unavoidable, from a minimax perspective, in the moderate confidence regime.
Algorithmically, we showcased how to exploit the sparsity induced by the unimodal structure in three different algorithms having efficiently implementation.
On top of their good empirical performance, the proposed algorithms enjoy strong theoretical guarantees on their expected sample complexity: asymptotic optimality or non-asymptotic guarantees.

% Future Works
Despite our promising results leveraging a known structural assumption, the unimodal structure is rather an exception than the norm.
In all generality, it is challenging to fully exploit the structure algorithmically.
Even for linear bandits, extending the analysis of Top Two algorithms is still an open problem.

\subsubsection*{Acknowledgements}

This work started while Riccardo Poiani was visiting the Scool team of Inria Lille and a student at Politecnico di Milano. He acknowledges the funding of a MOB-LIL-EX grant from the University of Lille. Riccardo Poiani is partially supported by the Cariplo CRYPTONOMEX grant and MUR - PRIN 2022 project 2022R45NBB funded by the Next Generation EU program. This work was conducted while Marc Jourdan was a PhD student in the Scool team of Inria Lille, partially supported by the THIA ANR program ``AI\_PhD@Lille''. 
The authors acknowledges the support of the French National Research Agency under the projects BOLD (ANR-19-CE23-0026-04) and FATE (ANR22-CE23-0016-01).

\bibliographystyle{abbrvnat}
\bibliography{biblio}

%%%%%%%%%%%%%%%%%%%%%%%%%%%%%%%%%%%%%%%%%%%%%%%%%%%%%%%%%%%%

%\iffalse
\section*{Checklist}

% %%% BEGIN INSTRUCTIONS %%%
% The checklist follows the references. For each question, choose your answer from the three possible options: Yes, No, Not Applicable.  You are encouraged to include a justification to your answer, either by referencing the appropriate section of your paper or providing a brief inline description (1-2 sentences). 
% Please do not modify the questions.  Note that the Checklist section does not count towards the page limit. Not including the checklist in the first submission won't result in desk rejection, although in such case we will ask you to upload it during the author response period and include it in camera ready (if accepted).

% \textbf{In your paper, please delete this instructions block and only keep the Checklist section heading above along with the questions/answers below.}
% %%% END INSTRUCTIONS %%%

\begin{enumerate}

 \item For all models and algorithms presented, check if you include:
 \begin{enumerate}
   \item A clear description of the mathematical setting, assumptions, algorithm, and/or model. \textbf{Yes, see Sections~\ref{sec:intro},~\ref{sec:lb} and~\ref{sec:algos}.}
   \item An analysis of the properties and complexity (time, space, sample size) of any algorithm. \textbf{Yes, see Sections~\ref{sec:algos} and~\ref{sec:expe}, as well as Appendix~\ref{app:add_expe}.}
   \item (Optional) Anonymized source code, with specification of all dependencies, including external libraries. \textbf{Yes, see Appendix~\ref{app:add_expe} and the code attached in the supplementary materials.}
 \end{enumerate}

 \item For any theoretical claim, check if you include:
 \begin{enumerate}
   \item Statements of the full set of assumptions of all theoretical results. \textbf{Yes, see Sections~\ref{sec:lb} and~\ref{sec:algos}.}
   \item Complete proofs of all theoretical results. \textbf{Yes, see Section~\ref{sssec:proof_sketch} and Appendices~\ref{app:proof-sec-lb},~\ref{app:stopping_rules},~\ref{app:TaS},~\ref{app:SAOTaS},~\ref{app:UniTT}}
   \item Clear explanations of any assumptions. \textbf{Yes, see Sections~\ref{sec:intro},~\ref{sec:lb} and~\ref{sec:algos}.}    
 \end{enumerate}

 \item For all figures and tables that present empirical results, check if you include:
 \begin{enumerate}
   \item The code, data, and instructions needed to reproduce the main experimental results (either in the supplemental material or as a URL).  \textbf{Yes, see Appendix~\ref{app:add_expe} and the code attached in the supplementary materials.}
   \item All the training details (e.g., data splits, hyperparameters, how they were chosen).  \textbf{Yes, see Appendix~\ref{app:add_expe} and the code attached in the supplementary materials.}
   \item A clear definition of the specific measure or statistics and error bars (e.g., with respect to the random seed after running experiments multiple times).  \textbf{Yes, see Section~\ref{sec:expe} and Appendix~\ref{app:add_expe}.}
   \item A description of the computing infrastructure used. (e.g., type of GPUs, internal cluster, or cloud provider).  \textbf{Yes, see Appendix~\ref{app:add_expe}.}
 \end{enumerate}

 \item If you are using existing assets (e.g., code, data, models) or curating/releasing new assets, check if you include:
 \begin{enumerate}
   \item Citations of the creator If your work uses existing assets. \textbf{Not Applicable.}
   \item The license information of the assets, if applicable. \textbf{Not Applicable.}
   \item New assets either in the supplemental material or as a URL, if applicable. \textbf{Not Applicable.}
   \item Information about consent from data providers/curators. \textbf{Not Applicable.}
   \item Discussion of sensible content if applicable, e.g., personally identifiable information or offensive content. \textbf{Not Applicable.}
 \end{enumerate}

 \item If you used crowdsourcing or conducted research with human subjects, check if you include:
 \begin{enumerate}
   \item The full text of instructions given to participants and screenshots. \textbf{Not Applicable.}
   \item Descriptions of potential participant risks, with links to Institutional Review Board (IRB) approvals if applicable. \textbf{Not Applicable.}
   \item The estimated hourly wage paid to participants and the total amount spent on participant compensation. \textbf{Not Applicable.}
 \end{enumerate}

 \end{enumerate}
%\fi

\begin{appendix}

\onecolumn

\section{OUTLINE}
\label{app:outline}

The appendices are organized as follows:
\begin{itemize}
  \item In Appendix~\ref{app:algo-comparison}, we present a table of comparison among the different algorithms that can be used to solve Unimodal BAI problems.
  \item In Appendix~\ref{app:proof-sec-lb}, we present the proofs of our lower bounds on the expected sample complexity of any $\delta$-correct algorithm for Unimodal BAI: an instance-dependent one (Theorem~\ref{theo:sparsity}) and a worst-case one (Theorem~\ref{thm:lb_K}).
  \item Additional details on the stopping rules and their GLR origins is given in Appendix~\ref{app:stopping_rules}.
  \item The asymptotic optimality of Track-and-Stop (Theorem~\ref{thm:TaS_asymptotic_optimality}) is proven in Appendix~\ref{app:TaS}.
  \item The asymptotic optimality of Optimistic Track-and-Stop (Theorem~\ref{thm:SAOTaS_asymptotic_optimality}) is proven in Appendix~\ref{app:SAOTaS}.
  \item In Appendix~\ref{app:UniTT}, we prove our non-asymptotic upper bound on the expected sample complexity of \hyperlink{UniTT}{UniTT} (Theorem~\ref{thm:non_asymptotic_upper_bound_UniTT}).
  \item Appendix~\ref{app:technicalities} gathers existing and new technical results which are used for our proofs.
  \item Implementation details and additional experiments are presented in Appendix~\ref{app:add_expe}.
\end{itemize}

\section{ALGORITHM COMPARISON}\label{app:algo-comparison}

Table~\ref{tab:comparison} compares different algorithms that can be used to solve unimodal BAI problems.
In the table, we denote by $\widetilde{T}^{\star}(\bm\mu)$ the characteristic time over unstructured BAI problems, namely:
\begin{align*}
	\widetilde{T}^{\star}(\bm\mu)^{-1} = \sup_{\bm\omega \in \Delta_K} \inf_{\substack{\bm\lambda \in \widetilde{\mathcal{S}}: \\ i^{\star}(\bm\lambda) \ne i^{\star}(\bm\mu)}} \sum_{k \in [K]} \omega_k d(\mu_k, \lambda_k) 
\end{align*}
where $\widetilde{S}$ is the unstructured space of possible means. Similarly, we denote by $\widetilde{T}^{\star}_{\beta}(\bm\mu)$ a $\beta$-optimal algorithm in the unstructured setting, i.e., 
\begin{align*}
\widetilde{T}_{\beta}^{\star}(\bm\mu)^{-1} = \sup_{\bm\omega \in \Delta_K: \omega_{\star}=\beta} \inf_{\substack{\bm\lambda \in \widetilde{\mathcal{S}}: \\ i^{\star}(\bm\lambda) \ne i^{\star}(\bm\mu)}} \sum_{k \in [K]} \omega_k d(\mu_k, \lambda_k).
\end{align*}

\begin{table}[h]
\caption{Algorithm Comparison. (Asymp.) Upper bound on $\lim_{\delta \to 0} \bE_{\bm \nu}[\tau_{\delta}]/\log(1/\delta)$}\label{tab:comparison}
\begin{center}
\begin{tabular}{llll}
\textbf{ALGO}  & \textbf{ASYMP.} & \textbf{FINITE TIME} & \textbf{COMPLEXITY} \\
\hline \\
TaS       & $\widetilde{T}^{\star}(\bm\mu)$ & N/A & $\textup{poly}(K)$, $1$ call to $\widetilde{T}^{\star}(\cdot)^{-1}$ \\
TTUCB     & $\widetilde{T}_{\beta}^{\star}(\bm\mu)$ & Thm. 2.4 in \cite{jourdan2023NonAsymptoticAnalysis} & $\mathcal{O}(K)$ \\
U-TaS     & $T^{\star}(\bm\mu)$ & N/A & $\textup{poly}(K)$, $1$ call to $T^{\star}(\cdot)^{-1}$ \\
O-TaS     & $T^{\star}(\bm\mu)$ & Theorem~\ref{theo-cmplx-O-TaS} & $\mathcal{O}(K)$, $K$ calls to $T^{\star}(\cdot)^{-1}$ \\
UniTT     & $T_{1/2}^{\star}(\bm\mu)$ & Theorem~\ref{thm:non_asymptotic_upper_bound_UniTT} & $\mathcal{O}(K)$, $0$ calls to $T^{\star}(\cdot)^{-1}$ \\
LMA		  & $T^{\star}(\bm\mu)$ & N/A & $K-1$ convex problems of $K$ variables \\
DKM		  & $T^{\star}(\bm\mu)$ & Thm. 2 in \cite{degenne2019non} & $K-1$ convex problems of $K$ variables \\
FW		  & $T^{\star}(\bm\mu)$ & N/A & $K-1$ convex problems of $K$ variables \\
\end{tabular}
\end{center}
\end{table}

We take the chance to remark that TTUCB enjoys a dependency on $K$ which is at least $\mathcal{O}\left( (K \log(K))^{\alpha} \right)$ with $\alpha > 1$, while DKM \cite{degenne2019non} shows an $\mathcal{O}(K^2)$ dependency.

% !TeX root = ../paper.tex

\section{LOWER BOUNDS}\label{app:proof-sec-lb}

\subsection{Proof of Theorem~\ref{theo:sparsity}}

\lb*
\begin{proof}
Given $x,y \in (0,1)$, we denote by $\textup{kl}(x,y) = x \log(x/y) + (1-x)\log((1-x)/(1-y))$ the binary relative entropy.
Consider $\delta \in (0,1)$, a unimodal bandit model $\bm{\mu}$ and an alternative instance $\bm{\theta} \in \textup{Alt}(\bm{\mu})$. Then, by Lemma 1 in \cite{kaufmann2016complexity}, we can connect the expected number of draws of each arm to the KL divergence of the two unimodal bandit models. More specifically, we have that:
\begin{align*}
    \sum_{j \in [K]} \E_{\bm{\nu}}[N_{j}(\tau_\delta)] d(\mu_{j}, \theta_{j}) \ge \textup{kl}(\delta, 1-\delta).
\end{align*}
Similar to Theorem 1 in \citep{garivier2016optimal}, we now use the previous inequality by considering all the alternative models $\bm{\theta} \in \mathcal{S}$ in which the optimal arm is different:
\begin{align*}
    \textup{kl}(\delta, 1-\delta) & \le \inf_{\bm{\theta} \in \textup{Alt}(\bm{\mu})}\sum_{j \in [K]} \E_{\bm{\nu}}[N_{j}(\tau_\delta)] d(\mu_{j}, \theta_{j}) \\ & = \E_{\bm{\nu}}[\tau_\delta] \inf_{\bm{\theta} \in \textup{Alt}(\bm{\mu})}\sum_{j \in [K]} \frac{\E_{\bm{\nu}}[N_{j}(\tau_\delta)]}{\E_{\bm{\nu}}[\tau_\delta]} {d(\mu_{j}, \theta_{j})}  \\ & \le \E_{\bm{\nu}}[\tau_\delta] \sup_{\bm{\omega} \in \Delta_K} \inf_{\bm{\theta} \in \textup{Alt}(\bm{\mu})}\sum_{j \in [K]} \omega_j {d(\mu_{j}, \theta_{j})} \\ & = \E_{\bm{\nu}}[\tau_\delta] \sup_{\bm{\omega} \in \Delta_K} \min_{i \ne i^{\star}(\bm{\mu})} \inf_{\substack{\bm{\theta} \in \mathcal{S} \\ i^{\star}(\bm{\theta})=i}} \sum_{j \in [K]} \omega_j d(\mu_j, \theta_j),
\end{align*}
where in the last step we have used the definition of $\textup{Alt}(\bm{\mu})$.
We conclude the proof by using that $\textup{kl}(\delta, 1-\delta) \ge \log \frac{1}{2.4 \delta}$.
\end{proof}

Given this standard result, we continue by analyzing in greater details the shape of $T^\star(\bm{\mu})^{-1}$. Before diving into the proof, we introduce the following notation. 
For all $i \ne \star$, we define $\bm{\theta}^{*, i}(\bm{\omega})$ in the following way:
\begin{align}\label{eq:theta-star-def}
\bm{\theta}^{*, i}(\bm{\omega}) \in \argmin_{\substack{\bm{\theta} \in \textup{Alt}(\bm{\mu}): \\ i^{\star}(\bm{\theta})=i}} \sum_{j \in [K]} \omega_j d(\mu_j, \theta_j).
\end{align}
Given this notation, we first show an preliminary step that orders the functions $f_j$'s.

\begin{restatable}{lemma}{domination}\label{lemma:domination}
Fix $\bm{\omega} \in \Delta_K$ and consider any unimodal bandit $\bm{\mu} \in \mathcal{S}$. Then, it holds that:
\begin{align*}
& f_j(\bm{\omega},\bm{\mu}) \ge f_{j+1}(\bm{\omega}, \bm{\mu}) \quad \forall j > i^{\star}(\bm{\mu}) \\
& f_j(\bm{\omega},\bm{\mu}) \le f_{j+1}(\bm{\omega},\bm{\mu}) \quad \forall j < i^{\star}(\bm{\mu}) -1.
\end{align*} 
\end{restatable}
\begin{proof}
We prove the result for $j < \star - 1$. The prove for the case in which $j > \star$ follows identical reasonings.
%Let us denote $\bm{\theta}^{*,j} \in \argmin_{\substack{\bm{\theta} \in \textup{Alt}(\bm{\mu}), \theta_j > \theta_{i^\star(\bm{\mu})}}} \sum_{a} \omega_a d(\mu_a, \theta_a)$. 
%We continue by splitting the analysis into two cases. 
%First, we consider the case in which $\mu_{j+1} > {\theta}^{*,j}_j$.
%Let $\mathcal{A} = \{ k \in [K] : \mu_k >  {\theta}^{*,j}_j\}$. Then, we have that:
%\begin{align*}
%f_j(\bm{\omega}, \bm{\mu}) & = \sum_{k \in \mathcal{A}} \omega_k d(\mu_k, {\theta}^{*,j}_k) + \sum_{k \notin \mathcal{A}}  \omega_k d(\mu_k, {\theta}^{*,j}_k) \\ & \ge \sum_{k \in \mathcal{A}} \omega_k d(\mu_k, {\theta}^{*,j}_k) \\ & \ge \sum_{k \in \mathcal{A}} \omega_k d(\mu_k, {\theta}^{*,j}_j) \\ & \ge \sum_{k \in \mathcal{A}: \mu_k > \mu_{j+1}} \omega_k d(\mu_k, {\theta}^{*,j}_k) \\ & \ge \sum_{k \in \mathcal{A}: \mu_k > \mu_{j+1}} \omega_k d(\mu_k, \mu_{j+1})  \\ & \ge f_{j+1}(\bm{\omega}, \bm{\mu}),
%\end{align*}
%where (i) in the second inequality we used the fact that $\mu_k > \theta^{*,j}_j \ge \theta_k^{*,j}$ by definition of $\mathcal{A}$ and by the fact that $j$ is the optimal arm in $\bm{\theta}^{*,j}$, (ii) in the forth inequality we have used $\mu_{k} > \mu_{j+1} > \theta^{*,i}_j$ by hyphotesis, and (iii) in the last step, we used the fact that $\theta_k = \mu_{j+1}$ for all $k$ such that $k \in \mathcal{A}: \mu_k \ge \mu_{j+1}$ and $\theta_a = \mu_a$ for all the remaining arms, is a unimodal bandit for which $\mu_{j+1}$ is an optimal arm.
%We now continue by considering the case in which $\mu_{j+1} \le \theta^{*,j}_j$. Here, we have that:
Specifically, we have that:
\begin{align*}
f_{j}(\bm{\omega}, \bm{\mu}) & \ge \sum_{k=j+1}^{K} \omega_k d(\mu_k, \theta^{*,j}_k(\bm{\omega})) \\ & = \sum_{k=1}^{j} \omega_k d(\mu_k, \mu_k) + \sum_{k=j+2}^K \omega_k d(\mu_k, \theta^{*,j}_{k}(\bm{\omega})) + \omega_{j+1} d(\mu_{j+1}, \theta^{*,j}_{j+1}(\bm{\omega})) \\ & \ge \sum_{k=1}^{j} \omega_k d(\mu_k, \mu_k) + \sum_{k=j+2}^K \omega_k d(\mu_k, \theta^{*,j}_{k}(\bm{\omega})) + \omega_{j+1} d(\mu_{j+1}, \max\{ \mu_{j+1}, \theta^{*,j}_{j+1}(\bm{\omega}) \}) \\ & \ge f_{j+1}(\bm{\omega}, \bm{\mu}),
\end{align*}
where in the last step we used the fact that $\theta_{j+1} = \max \{ \mu_{j+1}, \theta^{*,j}_{j+1}(\bm{\omega})) \}$, $\theta_k = \mu_k$ for all $k < j+1$ and $\theta_k = \theta^{*,j}_{j+1}(\bm{\omega}))$ is a unimodal bandit where $j+1$ is an optimal arm. Indeed, since $\bm{\theta}^{*,j}$ is a unimodal bandit where $j$ is the optimal arm, it follows that $\theta^{*,j}_{j+1}(\bm{\omega}) \ge \theta^{*,j}_k(\bm{\omega})$ for all $k > j+1$. Furthermore, $\max \{ \mu_{j+1}, \theta^{*,j}_{j+1}(\bm{\omega}) \} \ge \mu_{j}$ holds since $j < \star - 1$ and $\bm{\mu}$ is a unimodal bandit.
\end{proof}

A direct consequence of Lemma \ref{lemma:domination} is that we can rewrite $T^\star(\bm{\mu})^{-1}$ in the following way:
\begin{align}
T^\star(\bm{\mu})^{-1} & = \sup_{\bm{\omega} \in \Delta_{K}} \min_{a \in \mathcal{N}(\star)} f_a(\bm{\omega}, \bm{\mu}).
\end{align}
We now investigate in detail $f_{\star - 1}(\bm{\omega}, \bm{\mu})$ and $f_{\star + 1}(\bm{\omega}, \bm{\mu})$. 

\begin{restatable}{lemma}{alt}\label{lemma:alt}
Fix $\bm{\omega} \in \Delta_K$ and consider $i \in \mathcal{N}(\star)$. Then, if $i$ is such that $\mu_{i} \ge \mu_{j}$ for all $j \ne \star$, then it holds that:
\begin{align}
& {\theta}^{*,i}_i(\bm{\omega}) = {\theta}^{i}_{\star}(\bm{\omega}) = \argmin_{\eta \in (\mu_i, \mu_{\star})} \sum_{a \in \{\star, i\}} \omega_a d(\mu_a, \eta) =  \frac{\omega_{\star}}{\omega_{\star} + \omega_i} \mu_{\star} + \frac{\omega_{i}}{\omega_{\star} + \omega_i} \mu_{i} \label{eq:alt-1} \\
& {\theta}^{*,i}_a(\bm{\omega}) = \mu_a \quad  \forall a \notin \{ \star, i \}. \label{eq:alt-2}
\end{align}
If, instead, there exists $j \ne i^{\star}(\bm{\mu})$ such that $\mu_{j} > \mu_{i}$, then,
\begin{align}\label{eq:eta-opt}
f_{i}(\bm{\omega}) = \inf_{\eta \in \mathbb{R}} \omega_{\star} d(\star, \eta) + \omega_i d(\mu_i, \eta) + \sum_{a \notin \{ \star, i \}} \omega_a d(\mu_a, \eta) \bm{1}\{ \mu_a \ge \eta \}.
\end{align}
\end{restatable}
\begin{proof}
We begin by considering the case in which $i$ is such that $i$ is such that $\mu_{i} \ge \mu_{j}$ for all $j \ne \star$.
Suppose there exists $j \notin \{i, \star \}$ such that $\theta^{*,i}_j(\bm{\omega}) \ne \mu_j$. Define, for brevity:
\begin{align*}
\eta^* \coloneqq \frac{\omega_{\star}}{\omega_{\star} + \omega_i} \mu_{\star} + \frac{\omega_{i}}{\omega_{\star} + \omega_i} \mu_{i}.
\end{align*}
Then, we have that:
\begin{align*}
f_i(\bm{\omega}, \bm{\mu}) & = \sum_{i=1}^K \omega_k d(\mu_k, \theta^{*,i}_k(\bm{\omega})) \\ & \ge \omega_{\star} d(\mu_{\star}, \theta^{*,i}_{\star}) + \omega_i d(\mu_i, \theta_i^{*,i}(\bm{\omega})) \\ & \ge \inf_{x_1, x_2 \in [\mu_i, \mu_{\star}]^2: x_1 \le x_2} \omega_{\star} d(\mu_{\star}, x_1) + \omega_i d(\mu_i, x_2) \\ & = \inf_{\eta \in (\mu_i, \mu_{\star})} \omega_{\star} d\left(\mu_{\star}, \eta \right) + \omega_{i} d \left( \mu_i, \eta \right)  \\ & = \omega_{\star} d\left(\mu_{\star}, \eta^* \right) + \omega_{i} d \left( \mu_i, \eta^* \right),
\end{align*}
thus showing that ${\theta}^{*,i}_i(\bm{\omega}) = {\theta}^{i}_{\star}(\bm{\omega}) = \eta^*$ actually belongs to the argmin (i.e., Equation \eqref{eq:theta-star-def}).

We now conclude by proving Equation \eqref{eq:eta-opt}. The proof follows by analyzing in detail:
\begin{align}\label{eq:proof-alt-new-1}
\inf_{\substack{\bm{\theta} \in \textup{Alt}(\bm{\mu}): \\ i^{\star}(\bm{\theta})=i}} \sum_{a \in [K]} \omega_a d(\mu_a, \theta_a)
\end{align}
Suppose, without loss of generality that $i = \star - 1$. Consider $\bm{\theta}^{*,\star-1}$ and fix an arbitrary optimal values for ${\theta}^{*,i}_{\star-1}$ (i.e., $\eta_{\star-1}$) and ${\theta}^{*,i}_{\star}$ (i.e., $\eta_{\star}$). 
Then, for all $j < \star - 1$, it is easy to see that Equation \eqref{eq:proof-alt-new-1} is minimized by  (i) $\theta^{*,i}_j = \eta_{\star-1}$ if $\mu_j > \eta_{\star-1}$ and (ii) $\theta^{*,i}_j = \mu_j$ if $\mu_j \le \eta_{\star-1}$.\footnote{These facts are due to the unimodal constraint and the fact that $d(x,y)$ is increases as the distance between $x$ and $y$ increases.} For similar arguments, we have that, for all $j \ge \star + 1$, Equation \eqref{eq:proof-alt-new-1} is minimized by  (i) $\theta^{*,i}_j = \eta_{\star}$ if $\mu_j > \eta_{\star}$ and (ii) $\theta^{*,i}_j = \mu_j$ if $\mu_j \le \eta_{\star}$. Therefore, we have reduced Equation \eqref{eq:proof-alt-new-1} to the following expression:
\begin{align*}
\inf_{\eta_{\star - 1} \ge \eta_{\star}} \omega_{\star} d(\mu_{\star}, \eta_{\star}) + \omega_{\star - 1} d(\mu_{\star - 1}, \eta_{\star - 1}) + \sum_{j < \star - 1} \omega_j d(\mu_j, \eta_{\star-1}) \bm{1}\{ \mu_j > \eta_{\star-1} \} + \sum_{j \ge \star - 1} \omega_j d(\mu_j, \eta_{\star}) \bm{1}\{ \mu_j > \eta_{\star} \},
\end{align*}
which is a convex optimization problem of two variables. Nevertheless, we note that in its unconstrained version (i.e., $\eta_{\star-1}, \eta_{\star} \in \mathbb{R}^2$), the objective function is $0$ (i.e., $\eta_{\star-1} = \mu_{\star-1}$ and $\eta_{\star} = \mu_{\star}$). Therefore, due to the convexity, an optimal of the constrained problem lies at $\eta_{\star - 1} = \eta_{\star}$, thus concluding the proof.
\end{proof}

In other words, Lemma \ref{lemma:alt} states that (i) for the arm whose mean is the closest to the optimal one, we have a closed form expression for the alternative model, that is the classical one of BAI problems (see Lemma 3 in \cite{garivier2016optimal}). Instead, (ii) if $i \in \mathcal{N}(\star)$ but its mean is not the closest to the optimal one, then, the alternative instance is expressed as the solution of a 1D-optimization problem. Furthermore, we highlight that, Equation \eqref{eq:eta-opt} is due to the fact that ${\theta}^{*, i}_a(\bm{\omega}) = \eta^*(\bm{\omega})$ if $a \in \{ \star, i \}$ or if $\mu_a \ge \eta^*(\bm{\omega})$, and ${\theta}^{*, i}_a(\bm{\omega}) = \mu_a$ otherwise.

It is important to note that Equation \eqref{eq:eta-opt} also captures the case in which $i \in \argmax_{a \ne \star} \mu_i$ (i.e., Equation \eqref{eq:alt-1}). Indeed, by looking at Equation \eqref{eq:alt-1}, we can note that:
\begin{align*}
\frac{\omega_{\star}}{\omega_{\star} + \omega_i} \mu_{\star} + \frac{\omega_{i}}{\omega_{\star} + \omega_i} \mu_{i} \ge \mu_i.
\end{align*}
Plugging this into Equation \eqref{eq:eta-opt}, we have that:
\begin{align*}
f_i(\bm{\omega}, \bm{\mu}) = \inf_{\eta \in (\mu_i, \mu_{\star})} \omega_{\star} d(\star, \eta) + \omega_i d(\mu_i, \eta),
\end{align*}
whose optimal value is exactly $\eta_i^*(\bm{\omega}) \coloneqq \frac{\omega_{\star}}{\omega_{\star} + \omega_i} \mu_{\star} + \frac{\omega_{i}}{\omega_{\star} + \omega_i} \mu_{i}$.
Therefore, in the following we denote by $\eta_{i}^*(\bm{\omega})$ a solution of the optimization problem in Equation \eqref{eq:eta-opt}, keeping in mind that, whenever $i \in \argmax_{a \ne \star} \mu_a$, we have a closed-form expression that is given by Equation \eqref{eq:alt-1}.\footnote{Note that, in $\eta_i^*(\bm{\omega})$ we have dropped the dependence on $\bm{\mu}$ to simplify the notation}.
Given these considerations, the next result shows how $T^\star(\bm{\mu})^{-1}$ can be expressed as a maximization problem over a restricted portion of the simplex.

\begin{lemma}\label{lemma:only-three-arms}
Consider any unimodal bandit $\bm{\mu}$, and define $\widetilde{\Delta}_K(\bm{\mu}) \subseteq \Delta_K$ in the following way:
\begin{align}
\widetilde{\Delta}_K(\bm{\mu}) = \left\{ \bm{\omega} \in \Delta_K: \omega_i = 0 \quad \forall i \notin \mathcal{N}(\star) \right\}.
\end{align}
Then, it holds that:
\begin{align}
T^\star(\bm{\mu})^{-1} = \sup_{\bm{\omega} \in \widetilde{\Delta}_K} \min_{a \in \mathcal{N}(\star)} f_a(\bm{\omega}, \bm{\mu}).
\end{align}
\end{lemma}
\begin{proof}
First, we prove the claim whenever $\star = 1$ (but an identical reasoning can be applied to the case in which $\star=K$). In this case, by applying Lemma \ref{lemma:domination}, we obtain that:
\begin{align*}
T^\star(\bm{\mu}) = \sup_{\omega \in \Delta_K} f_{2}(\bm{\omega}, \bm{\mu}).
\end{align*}
Then, from Lemma \ref{lemma:alt}, we have that:
\begin{align*}
T^\star(\bm{\mu}) & = \sup_{\omega \in \Delta_K} \omega_1 d( \mu_{1}, \eta_2^*(\bm{\omega})) + \omega_2 d( \mu_{2}, \eta_2^*(\bm{\omega})) \\ & = \sup_{\omega \in \widetilde{\Delta}_K(\bm{\mu})} \omega_1 d( \mu_{1}, \eta_2^*(\bm{\omega})) + \omega_2 d( \mu_{2}, \eta_2^*(\bm{\omega}) ) \\ & =  \sup_{\omega \in \widetilde{\Delta}_K(\bm{\mu})} \min_{a \in \mathcal{N}(\star)} f_a(\bm{\omega}, \bm{\mu}),
\end{align*}
where in the second equality we have used that there is no dependency on $\omega_i$ for $i > 2$. 

It remains to analyze the situation in which $\star \in \{2, \dots, K-1 \}$. We suppose that $\mu_{\star - 1} \le \mu_{\star + 1}$, as the case in which $\mu_{\star - 1} \ge \mu_{\star + 1}$ follows symmetrical reasonings. 

We proceed by cases. 
\paragraph{Case 1}First of all, suppose that $\mu_{\star - 1} = \mu_{\star + 1}$. Then, by Lemma \ref{lemma:domination} and \ref{lemma:alt}, we have that $\min_{a \in \mathcal{N}(\star)} f_a(\bm{\omega})$ can be rewritten as a function of $\omega_{\star}$, $\omega_{\star + 1}$ and $\omega_{\star - 1}$, thus showing that:
\begin{align*}
T(\bm{\mu})^{-1} = \sup_{\omega \in \widetilde{\Delta}_K(\bm{\mu})} \min_{a \in \mathcal{N}(\star)} f_a(\bm{\omega}, \bm{\mu}).
\end{align*}

\paragraph{Case 2}We now continue by analyzing the case in which $\mu_{\star - 1} < \mu_{\star + 1}$. We first show that for all $j < \star - 1$, it holds that $\omega^*_j = 0$. Indeed, suppose that there $j < \star - 1$, it holds that $\omega^*_j > 0$ holds. By applying Lemma \ref{lemma:domination}, we first rewrite:
\begin{align*}
T^\star(\bm{\mu}) = \sup_{\omega \in \Delta_K} \min_{a \in \mathcal{N}(\star)} f_{a}(\bm{\omega}, \bm{\mu}).
\end{align*}
We now show that such $\bm{\omega}^\star$ cannot belong to the supremum. First of all, from Lemma \ref{lemma:alt} we have that:
\begin{align*}
f_{\star + 1}(\bm{\omega}^\star, \bm{\mu}) = \omega_{\star + 1}^* d(\mu_{\star + 1}, \eta_{\star+1}^*(\bm{\omega}^\star)) + \omega_{\star}^* d(\star, \eta_{\star+1}^*(\bm{\omega}^\star)),
\end{align*}
which is strictly increasing in $\omega_{\star + 1}$ and $\omega_{\star}$, thus showing that there exists $\bm{\tilde{\omega}}$ such that $f_{\star + 1}(\bm{\tilde{\omega}}, \bm{\mu}) > f_{\star + 1}(\bm{\omega}^\star, \bm{\mu})$, i.e., $\tilde{\omega}_{\star} = \omega_{\star}^* + \omega_j^*$ and $\tilde{\omega}_j = 0$. Similarly, consider $f_{\star - 1}(\bm{\omega}^\star, \bm{\mu})$. Consider a generic $\bm{\omega}$, and recall that from Lemma \ref{lemma:alt}, we have that:
\begin{align*}
f_{\star - 1}(\bm{\omega}, \bm{\mu}) = \inf_{\eta \in \mathbb{R}} \omega_{\star} d(\mu_{\star}, \eta) + \omega_{\star - 1} d(\mu_{\star - 1}, \eta) + \sum_{k \notin \{ \star, \star - 1 \}} \omega_k d(\mu_k, \eta) \bm{1}\{ \mu_k \ge \eta \}.
\end{align*}
Nevertheless, it is easy to see that there always exists $\eta \ge \mu_{\star - 1}$ such that $\eta$ belongs to the argmin of $f_{\star - 1}$. Indeed:
\begin{align*}
f_{\star - 1}(\bm{\omega}, \bm{\mu}) & \ge \inf_{\eta \in \mathbb{R}} \omega_{\star} d(\mu_{\star}, \eta) + \omega_{\star - 1} d(\mu_{\star-1}, \eta) + \sum_{\substack{k \notin \{ \star, \star - 1\}: \\ \mu_j \ge \mu_{\star - 1}}} \omega_k d(\mu_k, \eta) \bm{1}\{ \mu_k \ge \eta \} \\ & = \inf_{\eta \ge \mu_{\star - 1}} \omega_{\star} d(\mu_{\star}, \eta) + \omega_{\star - 1} d(\mu_{\star-1}, \eta) + \sum_{\substack{k \notin \{ \star, \star - 1\}: \\ \mu_j \ge \mu_{\star - 1}}} \omega_k d(\mu_k, \eta) \bm{1}\{ \mu_k \ge \eta \} \\ & = \inf_{\eta \ge \mu_{\star - 1}} \omega_{\star} d(\mu_{\star}, \eta) + \omega_{\star - 1} d(\mu_{\star-1}, \eta) + \sum_{k \notin \{ \star, \star - 1\}} \omega_k d(\mu_k, \eta) \bm{1}\{ \mu_k \ge \eta \} \\ & \ge f_{\star - 1}(\bm{\omega}, \bm{\mu}).
\end{align*}
Therefore, by restricting the optimization of $f_{\star-1}(\bm{\omega}, \bm{\mu})$ to $\eta \ge \mu_{\star - 1}$, we have that $f_{\star - 1}$ is a strictly continuous function of $\omega_{\star}$, thus showing that $\bm{\omega}^\star$ cannot belong to the supremum. Indeed, it holds that $\min_{a \in \mathcal{N}(\star)} f_a(\bm{\omega}^\star, \bm{\mu}) < \min_{a \in \mathcal{N}(\star)} f(\bm{\tilde{\omega}}, \bm{\mu})$.

Finally, we now show that for all $j > \star + 1$, it holds that $\omega_j^* = 0$.
By applying Lemma \ref{lemma:domination} and Lemma \ref{lemma:alt}, we can rewrite $T^\star(\bm{\mu})^{-1}$ as follows:
\begin{align*}
T^\star(\bm{\mu})^{-1} = \sup_{\bm{\omega} \in \Delta_K} \min \left\{ f_{\star - 1}(\bm{\omega},\bm{\mu}), f_{\star + 1}(\bm{\omega},\bm{\mu})\right\},
\end{align*}
where $f_{\star - 1}(\bm{\omega}, \bm{\mu}), f_{\star + 1}(\bm{\omega}, \bm{\mu})$ are given by:
\begin{align*}
& f_{\star + 1}(\bm{\omega}, \bm{\mu}) = \omega_{\star + 1} d(\mu_{\star + 1}, \eta_{\star + 1}^*(\bm{\omega})) + \omega_{\star} d(\mu_{\star}, \eta_{\star + 1}^*(\bm{\omega})) \\
& f_{\star - 1}(\bm{\omega}, \bm{\mu}) = \inf_{\eta \ge \mu_{\star - 1}} \omega_{\star} d(\mu_{\star}, \eta) + \omega_{\star + 1} d(\mu_{\star + 1}, \eta) + \sum_{a > \star + 1 } \omega_a d(\mu_a, \eta) \bm{1}\{ \mu_a \ge \eta \}
\end{align*}
We split the analysis into two further cases. First, we consider $\bm{\omega}^\star$ such that $\eta_{\star - 1}^*(\bm{\omega}^\star) \ge \mu_{\star + 1}$. Nevertheless, in this case, we have that $T^\star(\bm{\mu})^{-1}$ is only a function of $\omega_{\star}$, $\omega_{\star + 1}$ and $\omega_{\star - 1}$, thus showing that:
\begin{align*}
T(\bm{\mu})^{-1} = \sup_{\omega \in \widetilde{\Delta}_K(\bm{\mu})} \min_{a \in \mathcal{N}(\star)} f_a(\bm{\omega}, \bm{\mu}).
\end{align*}
Secondly, we consider the case in which $\eta_{\star - 1}^*(\bm{\omega}^\star)  \in [\mu_{\star - 1} ,\mu_{\star + 1})$. Suppose that $f_{\star + 1}(\bm{\omega}^\star, \bm{\mu}) \ge f_{\star - 1}(\bm{\omega}^\star, \bm{\mu})$. Nevertheless, we would have that:
\begin{align*}
f_{\star - 1}(\bm{\omega}^\star, \bm{\mu}) & > \omega^*_{\star} d(\mu_{\star}, \eta^*_{\star - 1}(\bm{\omega}^\star)) + \omega^*_{\star - 1} d(\mu_{\star - 1}, \eta^*_{\star - 1}(\bm{\omega}^\star)) + \omega^*_{\star + 1} d(\mu_{\star + 1}, \eta^*_{\star - 1}(\bm{\omega}^\star)) \\ & \ge \omega^*_{\star} d(\mu_{\star}, \eta_{\star - 1}^*(\bm{\omega}^\star)) + \omega_{\star + 1}d(\mu_{\star + 1}, \eta^*_{\star - 1}(\bm{\omega}^\star) \\ & \ge f_{\star + 1}(\bm{\omega}^\star, \bm{\mu}),
\end{align*}
thus leading to a contradiction.\footnote{In the first step, we have used the fact that, since there exist $j > \star + 1$ such that $\omega^*_{j} > 0$, then it must hold that $\omega^*_j d(\mu_j, \eta^*_{\star-1}(\bm{\omega}^\star)) > 0$. Otherwise, one could define an alternative $\tilde{\bm{\omega}}$ where $\tilde{\omega}_j = \omega^*_j - \epsilon$ and $\tilde{\omega}_{\star} =  \omega^*_{\star} + \epsilon$ for a sufficiently small $\epsilon > 0$. It is direct to verify that $\tilde{\bm{\omega}}$ contradicts the optimality of $\bm{\omega}^\star$.}
Similarly, if $f_{\star + 1}(\bm{\omega}^\star, \bm{\mu}) < f_{\star - 1}(\bm{\omega}^\star, \bm{\mu})$, one can define $\tilde{\bm{\omega}}$ as $\tilde{\omega}_{\star + 1} = \omega^*_{\star + 1} + \epsilon$ and $\tilde{\omega}_{j} = \omega^*_j - \epsilon$ for any $j > \star + 1$ such that $\omega^*_j > 0$. In this case, we note that $f_{\star + 1}(\tilde{\bm{\omega}}) > f_{\star + 1}(\bm{\omega}^\star)$ since $f_{\star + 1}$ is strictly increasing in $\omega_{\star + 1}$ and it does not depend on $\omega_j$. Furthermore, since $f_{\star - 1}$ is continuous in $\bm{\omega}$, by picking $\epsilon$ sufficiently small, we have that $\min_{a \in \mathcal{N}(\star)} f_a(\tilde{\bm{\omega}}, \bm{\mu}) > \min_{a \in \mathcal{N}(\star)} f_a({\bm{\omega}}^*, \bm{\mu})$, thus contradicing the optimality of the non-sparse $\bm{\omega}^\star$.

Therefore, we have shown that:
\begin{align*}
T(\bm{\mu})^{-1} = \sup_{\omega \in \widetilde{\Delta}_K(\bm{\mu})} \min_{a \in \mathcal{N}(\star)} f_a(\bm{\omega}, \bm{\mu}),
\end{align*}
thus concluding the proof.
\end{proof}

\sparsity*
\begin{proof}
To prove the result, we start from Lemma \ref{lemma:only-three-arms}, and we analyze:
\begin{align*}
T^\star(\bm{\mu})^{-1} = \sup_{\bm{\omega} \in \widetilde{\Delta}_K} \min_{a \in \mathcal{N}(\star)} f_a(\bm{\omega}, \bm{\mu}).
\end{align*}
More specifically, we will be interested in proving that, for an optimal oracle allocation $\bm{\omega}^\star$, the following holds for all $a \in \mathcal{N}(\star)$:
\begin{align}\label{eq:proof-sparsity-final-1}
f_a(\bm{\omega}^\star, \bm{\mu}) = g_{a}(\bm{\omega^*}, \bm{\mu}).
\end{align}
From Lemma \ref{lemma:alt}, the claim is direct for all $\bm{\mu}$ and all $a$ such that $a \in \argmax_{b \in \mathcal{N}(\star)} \mu_b$. 
It remains to show that Equation \eqref{eq:proof-sparsity-final-1} holds for arm $a$ whenever $a \in \mathcal{N}(\star)$ and there exists $b \ne a$ and $b \ne \star$ such that $\mu_b > \mu_a$. To this end, suppose, without loss of generality, that $\mu_{\star - 1} < \mu_{\star + 1}$.
Then, due to Lemma \ref{lemma:alt}, we know that, for all $\bm{\omega}$:
\begin{align*}
f_{\star - 1}(\bm{\omega}, \bm{\mu}) = \inf_{\eta \in \mathbb{R}} \omega_{\star} d(\mu_{\star}, \eta) + \omega_{\star - 1} d(\mu_{\star - 1}, \eta) + \sum_{a \ge \star + 1 } \omega_a d(\mu_a, \eta) \bm{1}\{ \mu_a \ge \eta \}.
\end{align*}
Suppose that $\eta_{\star - 1}^*(\bm{\omega}^\star) \ge \mu_{\star + 1}$. This directly implies that $f_{\star - 1}(\bm{\omega}^\star, \bm{\mu}) = g_{\star-1}(\bm{\omega^*}, \bm{\mu})$. On the other hand, suppose that $\eta^*(\bm{\omega}^\star) < \mu_{\star + 1}$. This would imply that:
\begin{align*}
f_{\star - 1}(\bm{\omega}^\star,\bm{\mu}) & > \omega^*_{\star} d(\mu_{\star}, \eta^*_{\star - 1}(\bm{\omega}^\star)) + \omega^*_{\star + 1} d(\mu_{\star+1}, \eta_{\star - 1}^*(\bm{\omega}^\star)) \\ & \ge f_{\star+ 1}(\bm{\omega}^\star,\bm{\mu}).
\end{align*}
where in the first step we have used that an optimal solution $\bm{\omega}^\star$, it holds that $\omega^*_{\star-1} d(\mu_{\star - 1}, \eta_{\star-1}^*(\bm{\omega}^\star)) > 0$ holds.\footnote{Indeed, if this was not the case, one could define an alternative $\tilde{\bm{\omega}}$ where $\tilde{\omega}_{\star-1} = \omega^*_{\star-1} - \epsilon$ and $\tilde{\omega}_{\star} =  \omega^*_{\star} + \epsilon$ for a sufficiently small $\epsilon > 0$. It is direct to verify that $\tilde{\bm{\omega}}$ contradicts the optimality of $\bm{\omega}^\star$.} 
Nevertheless, $f_{\star - 1}(\bm{\omega}^\star,\bm{\mu}) > f_{\star + 1}(\bm{\omega}^\star,\bm{\mu})$ contradicts the optimality of $\bm{\omega}^\star$. Indeed, one can define $\tilde{\bm{\omega}}$ as $\tilde{\omega}_{\star + 1} = \omega^*_{\star + 1} + \epsilon$ and $\tilde{\omega}_{\star-1} = \omega^*_{\star-1} - \epsilon$. In this case, we note that $f_{\star + 1}(\tilde{\bm{\omega}}) > f_{\star + 1}(\bm{\omega}^\star)$ since $f_{\star + 1}$ is strictly increasing in $\omega_{\star + 1}$ and it does not depend on $\omega_{\star-1}$. Furthermore, since $f_{\star - 1}$ is continuous in $\bm{\omega}$, by picking $\epsilon$ sufficiently small, we have that $\min_{a \in \mathcal{N}(\star)} f_a(\tilde{\bm{\omega}}, \bm{\mu}) > \min_{a \in \mathcal{N}(\star)} f_a({\bm{\omega}}^*, \bm{\mu})$.
\end{proof}

\subsection{Proof of Theorem~\ref{thm:lb_K}}

This proof uses techniques developped in \citep{al2022complexity} for the problem of finding all good arms in a bandit.

For $i \in [K]$, let $\mu^{(i)}$ be the problem where arm $i$ has mean $\Delta>0$ and all other arms have mean $0$. All arms are Gaussian with variance 1. These are unimodal bandit problems.

We write $\mathbb{P}^\tau_{\mu}$ for the restriction of $\mathbb{P}_\mu$ to the $\sigma$-algebra generated by $\tau$. For any $\tau$-measurable event $E$ (for example $\{N_j(\tau) > n\}$), $\mathbb{P}_\mu^\tau(E) = \mathbb{P}_\mu(E)$.

We describe a bandit model by specifying the law of each successive reward from each arm: the first rewards queried from arm $k$ will have a given distribution, then the second reward will have a (possibly different) distribution, etc. We call the sequence of distributions an array of reward laws.
In true bandit models, on which the algorithms are required to be $\delta$-correct, the distribution does not change. However for the construction of the lower bound we will use arrays where the distribution changes after some number $n$.

For $i,j \in [K]$, we write $\eta_{i \to j}^n$ for the following array of reward laws:
\begin{itemize}
    \item For $k \notin \{i,j\}$, $\eta_{i \to j,k}^n$ is constant equal to $\mathcal N(0,1)$.
    \item $\eta_{i \to j,i}^n$ is constant equal to $\mathcal N(\Delta,1)$.
    \item For the first $n$ rewards, $\eta_{i \to j,j}^n$ is $\mathcal N(0,1)$. For the next rewards, $\eta_{i \to j,j}^n$ is $\mathcal N(\Delta, 1)$.
\end{itemize}

\begin{lemma}\label{lem:TV_eta_le_prob}
For all $n \in \mathbb{N}$ and all $i,j \in [K]$,
\begin{align*}
TV(\mathbb{P}_{\mu^{(i)}}^\tau, \mathbb{P}_{\eta_{i \to j}^n}^\tau)
&\le \mathbb{P}_{\mu^{(i)}}(N_j(\tau) > n)
\: .
\end{align*}
\end{lemma}

\begin{proof}
We first show that 
\begin{align*}
TV(\mathbb{P}_{\mu^{(i)}}^\tau, \mathbb{P}_{\eta_{i \to j}^n}^\tau)
&\le \mathbb{P}_{\mu^{(i)}}(N_j(\tau) \le n) TV(\mathbb{P}_{\mu^{(i)}}^\tau(\cdot \mid N_j(\tau) \le n), \mathbb{P}_{\eta_{i \to j}^n}^\tau(\cdot \mid N_j(\tau) \le n))
    \\&\quad + \mathbb{P}_{\mu^{(i)}}(N_j(\tau) > n) TV(\mathbb{P}_{\mu^{(i)}}^\tau(\cdot \mid N_j(\tau) > n), \mathbb{P}_{\eta_{i \to j}^n}^\tau(\cdot \mid N_j(\tau) > n))
\: .
\end{align*}
This could be obtained by an application of the general property that conditioning increases $f$-divergences, but we give here a direct proof:
\begin{align*}
\mathbb{P}_{\mu^{(i)}}^\tau
&= \mathbb{P}_{\mu^{(i)}}(N_j(\tau) \le n) \mathbb{P}_{\mu^{(i)}}^\tau(\cdot \mid N_j(\tau) \le n)
    + \mathbb{P}_{\mu^{(i)}}(N_j(\tau) > n) \mathbb{P}_{\mu^{(i)}}^\tau(\cdot \mid N_j(\tau) > n)
\: , \\
\mathbb{P}_{\eta_{i \to j}^n}^\tau
&= \mathbb{P}_{\eta_{i \to j}^n}(N_j(\tau) \le n) \mathbb{P}_{\eta_{i \to j}^n}^\tau(\cdot \mid N_j(\tau) \le n)
    + \mathbb{P}_{\eta_{i \to j}^n}(N_j(\tau) > n) \mathbb{P}_{\eta_{i \to j}^n}^\tau(\cdot \mid N_j(\tau) > n)
\\
&= \mathbb{P}_{\mu^{(i)}}(N_j(\tau) \le n) \mathbb{P}_{\eta_{i \to j}^n}^\tau(\cdot \mid N_j(\tau) \le n)
    + \mathbb{P}_{\mu^{(i)}}(N_j(\tau) > n) \mathbb{P}_{\eta_{i \to j}^n}^\tau(\cdot \mid N_j(\tau) > n)
\: .
\end{align*}
The last equality is due to the two arrays of distributions being equal before $n$. By joint convexity of $TV$, we then have
\begin{align*}
TV(\mathbb{P}_{\mu^{(i)}}^\tau, \mathbb{P}_{\eta_{i \to j}^n}^\tau)
&\le \mathbb{P}_{\mu^{(i)}}(N_j(\tau) \le n) TV(\mathbb{P}_{\mu^{(i)}}^\tau(\cdot \mid N_j(\tau) \le n), \mathbb{P}_{\eta_{i \to j}^n}^\tau(\cdot \mid N_j(\tau) \le n))
    \\&\quad + \mathbb{P}_{\mu^{(i)}}(N_j(\tau) > n) TV(\mathbb{P}_{\mu^{(i)}}^\tau(\cdot \mid N_j(\tau) > n), \mathbb{P}_{\eta_{i \to j}^n}^\tau(\cdot \mid N_j(\tau) > n))
\: .
\end{align*}
Since $\mathbb{P}_{\mu^{(i)}}(\cdot \mid N_j(\tau) \le n) = \mathbb{P}_{\eta_{i \to j}^n}(\cdot \mid N_j(\tau) \le n)$, the first TV is zero. The second TV is at most 1.
\end{proof}

\begin{lemma}\label{lem:KL_eta}
For all $n \in \mathbb{N}$ and all $i,j \in [K]$,
\begin{align*}
KL(\mathbb{P}_{\eta_{i \to j}^n}^\tau, \mathbb{P}_{\eta_{j \to i}^n}^\tau)
&\le n \Delta^2
\: .
\end{align*}
\end{lemma}

\begin{proof}
Since restricting to a sigma-algebra decreases $KL$ (data-processing inequality), $KL(\mathbb{P}_{\eta_{i \to j}^n}^\tau, \mathbb{P}_{\eta_{j \to i}^n}^\tau) \le KL(\mathbb{P}_{\eta_{i \to j}^n}, \mathbb{P}_{\eta_{j \to i}^n}) = n \Delta^2$.
\end{proof}

\begin{lemma}\label{lem:TV_ub}
For all $n \in \mathbb{N}$ and all $i,j \in [K]$,
\begin{align*}
TV(\mathbb{P}_{\mu^{(i)}}^\tau, \mathbb{P}_{\mu^{(j)}}^\tau)
\le \mathbb{P}_{\mu^{(i)}}(N_j(\tau) > n) + \mathbb{P}_{\mu^{(j)}}(N_i(\tau) > n) + \sqrt{\frac{1}{2}n \Delta^2}
\: .
\end{align*}

\end{lemma}

\begin{proof}
Since $TV$ is symmetric and satisfies the triangle inequality,
\begin{align*}
TV(\mathbb{P}_{\mu^{(i)}}^\tau, \mathbb{P}_{\mu^{(j)}}^\tau)
\le TV(\mathbb{P}_{\mu^{(i)}}^\tau, \mathbb{P}_{\eta_{i \to j}^n}^\tau)
    + TV(\mathbb{P}_{\mu^{(j)}}^\tau, \mathbb{P}_{\eta_{j \to i}^n}^\tau)
    + TV(\mathbb{P}_{\eta_{i \to j}^n}^\tau, \mathbb{P}_{\eta_{j \to i}^n}^\tau)
\end{align*}
For the first two distances on the r.h.s., we use Lemma~\ref{lem:TV_eta_le_prob}. For the third one, we use Pinsker's inequality to get
\begin{align*}
TV(\mathbb{P}_{\mu^{(i)}}^\tau, \mathbb{P}_{\mu^{(j)}}^\tau)
&\le \mathbb{P}_{\mu^{(i)}}(N_j(\tau) > n) + \mathbb{P}_{\mu^{(j)}}(N_i(\tau) > n) + \sqrt{\frac{1}{2}KL(\mathbb{P}_{\eta_{i \to j}^n}^\tau, \mathbb{P}_{\eta_{j \to i}^n}^\tau)}
\\
&= \mathbb{P}_{\mu^{(i)}}(N_j(\tau) > n) + \mathbb{P}_{\mu^{(j)}}(N_i(\tau) > n) + \sqrt{\frac{1}{2}n \Delta^2}
\: .
\end{align*}
\end{proof}

\begin{lemma}\label{lem:TV_lb_delta}
For all $n \in \mathbb{N}$ and all $i,j \in [K]$, for any $\delta$-correct algorithm,
\begin{align*}
1 - 2 \delta
\le TV(\mathbb{P}_{\mu^{(i)}}^\tau, \mathbb{P}_{\mu^{(j)}}^\tau)
\: .
\end{align*}
\end{lemma}

\begin{proof}
By the $\delta$-correct assumption, and since $\{\hat{a}_\tau = i\}$ is $\tau$-measurable,
\begin{align*}
1 - 2 \delta \le \mathbb{P}_{\mu^{(i)}}(\hat{a}_\tau = i) - \mathbb{P}_{\mu^{(j)}}(\hat{a}_\tau = i)
\le TV(\mathbb{P}_{\mu^{(i)}}^\tau, \mathbb{P}_{\mu^{(j)}}^\tau)
\: .
\end{align*}
\end{proof}

\begin{lemma}\label{lem:lb_K_n}
For all $n \in \mathbb{N}$, for any $\delta$-correct algorithm,
\begin{align*}
\frac{n K}{2}\left(1 - 2 \delta - \sqrt{\frac{1}{2}n \Delta^2}\right)
\le \frac{1}{K}\sum_{i}\mathbb{E}_{\mu^{(i)}}[\tau]
\: .
\end{align*}
\end{lemma}

\begin{proof}
Combining Lemma~\ref{lem:TV_lb_delta} and Lemma~\ref{lem:TV_ub}, we have for all $i,j$,
\begin{align*}
1 - 2 \delta
\le \mathbb{P}_{\mu^{(i)}}(N_j(\tau) > n) + \mathbb{P}_{\mu^{(j)}}(N_i(\tau) > n) + \sqrt{\frac{1}{2}n \Delta^2}
\: .
\end{align*}
We sum over all pairs $i,j$ and divide by the number of pairs, $K^2$, to get
\begin{align*}
1 - 2 \delta
\le \frac{2}{K^2} \sum_{i,j} \mathbb{P}_{\mu^{(i)}}(N_j(\tau) > n) + \sqrt{\frac{1}{2}n \Delta^2}
\: .
\end{align*}
Then use Markov's inequality: $\mathbb{P}_{\mu^{(i)}}(N_j(\tau) > n) \le \mathbb{E}_{\mu^{(i)}}[N_j(\tau)]/n$. We finally use $\sum_j \mathbb{E}_{\mu^{(i)}}[N_j(\tau)] = \mathbb{E}_{\mu^{(i)}}[\tau]$.
\begin{align*}
1 - 2 \delta
\le \frac{2}{n K^2} \sum_{i} \mathbb{E}_{\mu^{(i)}}[\tau] + \sqrt{\frac{1}{2}n \Delta^2}
\: .
\end{align*}
The result of the lemma is obtained by reordering that inequality.

\end{proof}

\lbk*
\begin{proof}
We choose $n = \frac{2}{\Delta^2}\left(\frac{1 - 2 \delta}{2}\right)^2$ in Lemma~\ref{lem:lb_K_n} to obtain
\begin{align*}
\frac{K}{\Delta^2}\left(\frac{1 - 2 \delta}{2}\right)^3 \le \frac{1}{K}\sum_{i}\mathbb{E}_{\mu^{(i)}}[\tau]
\: .
\end{align*}
The bound of the theorem finally uses $\delta \le 1/4$.

\end{proof}

% !TeX root = ../paper.tex

\section{STOPPING RULES} 
\label{app:stopping_rules}

\subsection{Generalized Likelihood Ratio (GLR)}
\label{app:ssec_GLR}

The generalized log-likelihood ratio between the model space $\mathcal{M}$ and a subset $\Lambda \subseteq \mathcal{M}$ is 
\[
	\mbox{GLR}_t^\mathcal{M}(\Lambda) \eqdef \log \frac{\sup_{\bm \lambda \in \mathcal{M}} \mathcal{L}_{\bm \lambda}(X_1,\ldots,X_t)}{\sup_{\bm \theta \in \Lambda} \mathcal{L}_{\bm \theta}(X_1,\ldots,X_t)} \: .
\]
For a canonical one-parameter exponential family, the likelihood ratio for two models with means $\bm\lambda, \bm\theta \in \mathcal{M}$ is
\begin{align*}
    \log \frac{\mathcal{L}_{\bm\lambda}(X_1,\ldots,X_t)}{\mathcal{L}_{\bm\theta}(X_1,\ldots,X_t)}
    = \sum_{i \in [K]} N_{i}(t) \left( d (\hat \mu_{i}(t) , \theta_{i}) - d(\hat \mu_{i}(t) , \lambda_{i}) \right) \: .
\end{align*}
The maximum likelihood estimator (MLE) is defined as
\begin{equation} \label{eq:mle_estimator}
	\bm{\tilde{\mu}} (t) \eqdef \argmin_{\bm \lambda \in\mathcal{M}} \sum_{i \in [K]} N_{i}(t) d(\hat \mu_{i}(t) ,\lambda_{i}) \: .
\end{equation}
When $\bm{\hat{\mu}}(t)  \in \mathcal{M}$, the maximum likelihood estimator coincide with the empirical mean, i.e. $\bm{\hat{\mu}}(t)  = \bm{\tilde{\mu}}(t)$.

\paragraph{Model Space}
In Unimodal bandits, our model space is the set $\cS$ of unimodal instances defined in~\eqref{eq:structure_Unimodal}.
Empirically, the empirical mean often fails to reflect the unimodal structure, i.e. $\bm{\hat{\mu}}(t)  \notin \cS$.
To the best of our knowledge, there is no efficient implementation to compute the MLE defined in~\eqref{eq:mle_estimator} when considering $\cM = \cS$.
To circumvent this computational bottleneck, we consider the unstructured model space $\mathcal{M} = \bR^K$.
Therefore, the GLR for set $\Lambda$ is
\begin{equation} \label{eq:GLR_unstruct_space}
	\mbox{GLR}_t(\Lambda) =  \min_{\bm \theta \in \Lambda} \sum_{i \in [K]} N_{i}(t) d(\hat \mu_{i}(t) , \theta_{i}) \: .
\end{equation}
At this point, it is important to note that $\mbox{GLR}_t(\cS)$ can be large when the empirical evidence is conflicting with the unimodal structure.
Due to the cumulative sum over all the arms, even small violation at the level of each arm will yield a large value of $\mbox{GLR}_t(\cS)$.
Therefore, we should be cautious when choosing the subset $\Lambda$ in our parallel testing proecdures (detailed in Appendix~\ref{app:ssec_parallel_testing}).
On the one hand, $\Lambda$ should leverage the structure since smaller $\Lambda$ yields larger $\mbox{GLR}_t(\Lambda)$.
On the other hand, choosing $\Lambda$ too small by hinging too much on the structure will result in a statistic $\mbox{GLR}_t(\Lambda)$ which is larger simply due to the cumulated violations of the structural constraint.

\subsection{Parallel Testing}
\label{app:ssec_parallel_testing}

We use the generalized likelihood ratio (GLR) stopping rule~\citep{garivier2016optimal}.
The idea is to run $K$ sequential tests in parallel.
Given an answer $i \in [K]$, we should define a set $\neg i$ of \emph{alternative to $i$}.
Then, we consider the GLR statistic $\mbox{GLR}_t(\neg i)$ for the following two non-overlapping hypotheses $\cH_{0,i} : (\mu \in \neg i)$ against $\cH_{1,i} : (\mu \notin \neg i )$.
Importantly, a high value of $\text{GLR}_n(\neg i)$ indicates that we have collected a large amount of evidence to reject $\cH_{0,i}$.
Let $i_{F}(\bm{\hat{\mu}}(t), \bm N_t)$ be the set of \emph{instantaneous furthest} (or \emph{instantaneous easiest-to-verify}) answers which maximizes the GLR, i.e. $i_{F}(\bm{\hat{\mu}}(t), \bm N_t) = \argmax_{i \in [K]} \text{GLR}_n(\neg i)$ . 
At time $t$, it is natural to recommend $\hat \imath_t \in i_{F}(\bm{\hat{\mu}}(t), \bm N_t)$ as we have collected the most evidence that it is correct.
The GLR stopping rule stops as soon as one of these tests can reject the null hypothesis with sufficient confidence.
In other words, it stops as soon as the GLR statistic exceeds a given stopping threshold $c : \bN \times (0,1) \to \bR_{+}$ , 
\begin{equation} \label{eq:general_glr_stopping_rule}
	\tau_{\delta} = \inf \left\{ t \mid \max_{i \in [K]} \text{GLR}_t(\neg i) > c(t-1, \delta)\right\} \: ,
\end{equation}
and recommends $\hat \imath_{\tau_{\delta}} \in \argmax_{i \in [K]} \text{GLR}_{\tau_{\delta}}(\neg i)$ since $\bm \mu$ is believed to admit $\hat \imath_{\tau_{\delta}}$ as a correct answer.
When there are multiple correct answers, other tests could have rejected the null hypothesis if provided with more samples, i.e. several answers could satisfy that $\text{GLR}_t(\neg i) > 0$ at time $t$.
Regardless of the sampling rule, the stopping threshold $c(t,\delta)$ is chosen to ensure $\delta$-correct by using time-uniform concentration results.
Namely, it is such that the probability that there exists a time $t$ such that $\text{GLR}_t(\neg \hat \imath_t) \ge c(t-1, \delta)$ and $\hat \imath_t$ is not correct is upper bounded by $\delta$.
We refer the reader to Appendix~\ref{app:ssec_correctness} for more details on the proof of $\delta$-correctness.

The exact definition of the GLR stopping rule crucially depends on the choice of the set $\neg i$ of alternative to $i$.
We explore several choices of $\neg i$ using the unimodal structure at different levels.

\paragraph{Global Alternatives}
Given an answer $i$, the set of \emph{global} alternative to $i$ is defined as the set of unimodal instances which disagrees with the fact that arm $i$ is the best arm, i.e.
\begin{equation} \label{eq:global_alt}
	\neg^{\mathrm{G}} i = \mathcal{S} \cap \left( \bigcup_{j \ne i} \{ \bm{\lambda} \in \bR^K \mid  \lambda_{j} \ge \lambda_{i} \} \right) \: .
\end{equation}
Therefore, $\text{GLR}_{t}(\neg^{\mathrm{G}} i)$ represents the amount of collected evidence to reject that $\bm \theta$ is a unimodal instance for which arm $i$ is dominated by another arm.
When there exists an arm $i$ such that $\text{GLR}_t(\neg^{\mathrm{G}} i) > c(t-1, \delta)$, we can reject that $\bm \theta \in \neg^{\mathrm{G}} i$ at confidence level $1-\delta$.
Since we know that $\bm \theta \in \mathcal{S}$, it means that $i = i^\star$ at confidence level $1-\delta$.

Let $i^\star(t) \eqdef i^\star(\bm{\hat{\mu}}(t))$ be the empirical best arm at time $t$.
When $\bm{\hat{\mu}}(t) \in \cS$, it is direct to show that $i^\star(t) = i_{F}(\bm{\hat{\mu}}(t), \bm N_t)$ since $\text{GLR}_{t}(\neg^{\mathrm{G}} i) = 0$ for all $i \ne i^\star(t)$. 
Recall that we have $\textup{Alt}(\bm{\hat{\mu}}(t)) = \neg^{\mathrm{G}}  i^\star(t)$.
When $\bm{\hat{\mu}}(t) \notin \cS$, multiple candidate answer $i$ will satisfy that $\text{GLR}_{t}(\neg^{\mathrm{G}} i) > 0$.
Since we don't necessarily have that $i^\star(t) \in i_{F}(\bm{\hat{\mu}}(t), \bm N_t)$, using $i^\star(t)$ as a recommendation rule is a principle choice only when $\bm{\hat{\mu}}(t) \in \cS$.
At early time $t$, more than being evidence to reject that arm $i$ is dominated by another arm, the statistics $(\text{GLR}_{t}(\neg^{\mathrm{G}} i))_{i \in [K]}$ could be driven by the evidence in favor of rejecting that $\bm \theta$ is unimodal.
Echoing the discussion in Appendix~\ref{app:ssec_GLR}, this suggests that the global alternatives defined in~\eqref{eq:global_alt} include too much structure. 

\paragraph{Local Alternatives}
Instead of leveraging structure globally, we propose to use the unimodal structure locally, namely that global optimality is equivalent to local optimality in unimodal instances.
Given an answer $i$, the set of \emph{local} alternative to $i$ is defined as the set of instances which disagrees with the fact that arm $i$ is better than its neighbors, i.e.
\begin{equation} \label{eq:local_alt}
	\neg^{\mathrm{L}} i =  \bigcup_{j \in \cN(i)} \{ \bm{\lambda} \in \bR^K \mid  \lambda_{j} \ge \lambda_{i} \}  \: .
\end{equation}
Therefore, $\text{GLR}_{t}(\neg^{\mathrm{L}} i)$ represents the amount of collected evidence to reject that $\bm \theta$ is an instance for which arm $i$ is dominated by one of its neighboring arms.
When there exists an arm $i$ such that $\text{GLR}_t(\neg^{\mathrm{L}} i) > c(t-1, \delta)$, we can reject that $\max_{j \in \cN(i)} \theta_{j} \ge \theta_{i}$ at confidence level $1-\delta$.
Since it means that arm $\max_{j \in \cN(i)} \theta_{j} < \theta_{i}$ at confidence level $1-\delta$, knowing that $\bm \theta \in \cS$ implies that $i = i^\star$ at confidence level $1-\delta$.

While $i^\star(t) \in \{i\in [K] \mid \text{GLR}_{t}(\neg^{\mathrm{L}} i) > 0\}$, we don't necessarily have $i^\star(t) \in i_{F}(\bm{\hat{\mu}}(t), \bm N_t)$.

\subsection{Correctness}
\label{app:ssec_correctness}

We now prove Lemma \ref{lem:delta_correct_threshold}, i.e., the $\delta$-correctness of the stopping rule proposed in Equation \eqref{eq:GLR_stopping_rule}. 
We now specify some notation. Let $\cC_{G}(x) = \min_{]1/2, 1]} \frac{g_G(\lambda) + x}{\lambda}$, where $g_G(\lambda)$ is given by:
\begin{align*}
g_G(\lambda) = 2\lambda - 2\lambda \log(4\lambda) + 2 \log \zeta (2\lambda) - \frac{1}{2} \log(1-\lambda),
\end{align*}
where $\zeta$ is the Riemann $\zeta$ function. We are now ready to prove Lemma \ref{lem:delta_correct_threshold}.

\begin{proof}[Proof of Lemma \ref{lem:delta_correct_threshold}]
Define, for all $i \in [K]$, $\tilde{\imath}(i) \coloneqq \argmax_{j \in \mathcal{N}(i)} \mu_i$.
Then, with probabilistic arguments, we have that:
\begin{align*}
\mathbb{P}_{\bm{\mu}} \left( \tau_{\delta} < + \infty, \: \hat \imath_{\tau_{\delta}}  \ne i^{\star} \right) & \le \mathbb{P}_{\bm{\mu}} \left( \exists t \in  \mathbb{N}: \exists i \ne i^{\star}: \min_{j \in \mathcal{N}(i)}W_{t}(i ,j) \ge c(t-1,\delta) \right) \\ & \le \sum_{i \ne i^{\star}} \mathbb{P}_{\bm{\mu}} \left( \exists t \in \mathbb{N}: \min_{j \in \mathcal{N}(j)} W_t(i,j) \ge c(t-1, \delta) \right) \\ & \le \sum_{i \ne i^{\star}} \mathbb{P}_{\bm{\mu}} \left( \exists t \in \mathbb{N}: W_t(i,\tilde{\imath}(i)) \ge c(t-1, \delta) \right) \\ & \le \sum_{i \ne i^{\star}} \mathbb{P}_{\bm{\mu}} \left( \exists t \in \mathbb{N}: \sum_{k \in \{i, \tilde{\imath}(i) \}} N_k(t) d(\hat{\mu}_k(t), \mu_k) \ge c(t-1, \delta) \right) \\ & \le \delta.
\end{align*}
where (i) in the third inequality we used the definition of $\tilde{\imath}(i)$, and (ii) in the last step, we have used Theorem 9 in \cite{kaufmann2021mixture}.
\end{proof}

%s\todom{Write the proof of $\delta$-correctness here. Maybe do it for the different set of alternatives that we considered.}

% !TeX root = ../paper.tex
\section{TRACK-AND-STOP} 
\label{app:TaS}

The proof of Theorem \ref{thm:TaS_asymptotic_optimality}, which shows that TaS is asymptotically optimal for unimodal bandits, follows the one of the original TaS algorithm \citep{garivier2016optimal} for unstructured best-arm identification problems. This is expected; indeed, Theorem \ref{theo:sparsity} has reduced unimodal BAI problems to unstructured problems in which the only arms to be considered are $\star$ and $\mathcal{N}(\star)$. 

Before provind the Theorem, we first recall a result on the C-Tracking forced exploration. 
\begin{lemma}[Lemma 7 in \cite{garivier2016optimal}]\label{lemma:forced-exp-u-tas}
Let $t \ge 1$ and $i \in [K]$, then, the C-Tracking rule ensures that $N_i(t) \ge \sqrt{t + K^2} - 2K$ and:
\begin{align*}
\Big| N_i(t) - \sum_{s=0}^{t-1} \omega^*_i(s) \Big| \le K(1+\sqrt{t}).
\end{align*}
\end{lemma}

As one can see from Lemma \ref{lemma:forced-exp-u-tas}, TaS pulls all the arms (even the ones for which $\omega^*_i = 0$ holds) at least $\sqrt{t}$ times. 

We are now ready to prove Theorem \ref{thm:TaS_asymptotic_optimality}.

\begin{proof}[Proof of Theorem \ref{thm:TaS_asymptotic_optimality}]
Let $\epsilon > 0$. Then, there exists $\xi = \xi(\epsilon) \le \frac{\min_{i \ne \star} \mu_{\star} - \mu_i}{4}$ such that:
\begin{align*}
\mathcal{I}_{\epsilon} \coloneqq \prod_{i \in \star \cup \mathcal{N}(\star)} [\mu_i - \xi, \mu_i + \xi]
\end{align*}
is such that, for all $\bm{\theta} \in \mathcal{S}$ in which the arms in $\star \cup \mathcal{N}(\star)$ belongs to $\mathcal{I}_{\epsilon}$, it holds that:
\begin{align}\label{eq:proof-tas-eq-1}
|| \bm{\omega}^\star(\bm{\mu}) - \bm{\omega}^\star(\bm{\theta}) || \le \epsilon.
\end{align}
In the following, we will refer to this subset of bandit models as $\mathcal{S}_{\epsilon}$. More precisely:
\begin{align*}
\mathcal{S}_{\epsilon} \coloneqq \left\{ \bm{\theta} \in \mathcal{S}: \forall i \in \star \cup \mathcal{N}(\star):  \theta_i \in [\mu_i - \xi, \mu_i + \xi] \right\}.
\end{align*}
We note that the existence of $\xi$ such that Equation \eqref{eq:proof-tas-eq-1} is satisfied for all $\bm{\mu} \in \mathcal{S}_{\epsilon}$ is guaranteed by (i) the continuity of $\bm{\omega}^\star(\bm{\mu})$ in $\bm{\mu}$, (ii) the sparsity of the oracle weights (i.e., Theorem \ref{theo:sparsity}) and (iii) the fact that $\bm{\omega}^\star(\bm{\mu})$ is not affected by the values of the arms whih are not in $\star \cup \mathcal{N}(\star)$.

At this point, let $\mathcal{E}_{\epsilon}(T)$ be the following event:
\begin{align*}
\mathcal{E}_{\epsilon}(T) \coloneqq \bigcap_{t=h(T)}^T \{ \bm{\tilde{\mu}}(t) \in \mathcal{S}_{\epsilon} \}
\end{align*}
where $h(T)=T^{1/4}$. Due to the definition of $\bm{\tilde{\mu}}(t)$ and $\mathcal{S}_{\epsilon}$, and by applying Lemma 19 in \cite{garivier2016optimal} together with Lemma \ref{lemma:forced-exp-u-tas} (i.e., forced exploration), there exists constant $B$ and $C$ such that:
\begin{align*}
\mathbb{P}_{\bm{\mu}}(\mathcal{E}^c_{\epsilon}(T)) \le BT \exp\left( -CT^{-\frac{1}{8}} \right).
\end{align*}
The rest of the proof follows exactly as in the original TaS paper \citep{garivier2016optimal}.
\end{proof}

% !TeX root = ../paper.tex

\section{OPTIMISTIC TRACK-AND-STOP} 
\label{app:SAOTaS}

\iffalse
\subsection{Assumptions}

The proof of the theoretical guarantees of O-TaS is conducted under the following assumptions. These assumptions are satisfied for a general class of problems. For further discussion, we refer the reader to \cite{degenne2019non}.

\begin{restatable}{assumption}{bounded}[Bounded parameters]\label{ass:bound}
There exists $M$ such that for all $\bm{\mu}, \bm{\mu'} \in \Theta$, $|| \bm{\nu}_{\bm{\mu}} - \bm{\nu}_{\bm{\mu'}} ||_{\infty} \le M$, where $\bm{\nu}_{\bm{\mu}}$  is the natural parameter vector corresponding to the mean vector $\bm{\mu}$.
\end{restatable}

\begin{restatable}{assumption}{subgauss}[Sub-gaussianity]\label{ass:sub-gauss}
There exists $\sigma^2$ such that, for all $k \in [K]$ and, all $\bm{\mu}, \bm{\mu'} \in {\Theta}$, $d(\mu_k, \mu_k') \ge \frac{(\mu_k' - \mu_k)^2}{2\sigma^2}$.
\end{restatable}

\begin{restatable}{assumption}{expfun}[Exploration function]\label{ass:exp-fun}
There exists a known, non decreasing function $f: \mathbb{N} \rightarrow  \mathbb{R}_{+}$ such that, for some constant $\gamma > 0$, it holds that:
\begin{align*}
\mathbb{P} \left( \exists k \in [K], t \le n, N_k(t) d(\hat{\mu}_k(t), \mu_k) > f(n) \right) \le \gamma \frac{\log(n)}{n^3}.
\end{align*}
\end{restatable}
\fi

\subsection{Concentrations}
We proceed by definying the event $\mathcal{E}_t$ in the following way:
\begin{align}
\mathcal{E}_n = \{ \forall k \in [K], \forall t \le n, N_k(t) d(\hat{\mu}_k(t), \mu_k) \le f(n) \}.
\end{align}

We first show that, under Assumption~\ref{ass:exp-fun}, the event $\mathcal{E}_n$ enjoy nice statistical properties.
\begin{lemma}\label{lemma:prob-bad-event}
Under Assumption~\ref{ass:exp-fun},
\begin{align*}
\sum_{t=2}^{+\infty} t \mathbb{P}\left( \mathcal E_t^c \right) \le \gamma \: .
\end{align*}
\end{lemma}
\begin{proof}
\begin{align*}
\sum_{t=2}^{+\infty} t \mathbb{P}\left( \mathcal E_t^c \right)
\le \sum_{t=2}^{+\infty} \gamma\frac{\log t}{t^2}
\le \gamma \int_{x=1}^{+\infty} \frac{\log x}{x^2} dx
= \gamma
\: .
\end{align*}
\end{proof}

As a direct consequence of the definition of $\mathcal{E}_t$, and of the exploration function $f(\cdot)$, we have the following result.
\begin{lemma}\label{lemma:concentration-after-sqrt-t}
Under event $\mathcal E_t$, for all $s \in \{\lceil \sqrt{t} \rceil, \ldots, t\}$ and all $k \in [K]$,
$N_{k}(s) d(\hat{\mu}_{k}(s), \mu_k) \le f(s^2)$ .
\end{lemma}
\begin{proof}
Under the event, $N_{k}(s) d(\hat{\mu}_{k}(s), \mu_k) \le f(n)$ and since $s \ge \sqrt{n}$ and $f$ is non-decreasing, $f(n) \le f(s^2)$.
\end{proof}

In the proof of the main theorem, we need to replace $\bm{\mu}$ by $\bm{\hat{\mu}}(t)$ or inversely in expressions involving divergences. We hence need to be able to control the error we introduce in those steps. To this end, we define $\Gamma(\mu_k, \mu'_k)$ as follows:
\begin{align*}
\Gamma(\mu_k, \mu'_k) \coloneqq \sup_{\bm{\theta} \in \mathcal S} (d(\mu_k, \theta_k) - d(\mu'_k, \theta_k)).
\end{align*}

In order to control $\Gamma(\cdot, \cdot)$, we begin by recalling some properties on the KL divergence for distributions in an exponential family. 

\begin{lemma}\label{lemma:divergence}
For distributions in the exponential family with means $a, b, c$,
\begin{align*}
d(a, b) &= d(a, c) + d(c, b) + (\tilde{b} - \tilde{c})(c - a) \: ,
\end{align*}
where we write $\tilde{a}$ for the natural parameter of the distribution with mean $a$.
\end{lemma}
\begin{proof}
For any $a,b$, $d(a, b) = \phi(\tilde{b}) - \phi(\tilde{a}) - (\tilde{b}-\tilde{a})a$.
\begin{align*}
d(a, b) - d(a, c) - d(c, b)
&= - (\tilde{b}-\tilde{a})a + (\tilde{c} - \tilde{a})a  + (\tilde{b}-\tilde{c})c
\\
&= (\tilde{b} - \tilde{c})(c - a)
\: .
\end{align*}
\end{proof}

An immediate consequence of Lemma \ref{lemma:divergence} is $d(c, b) - d(a, b) \le (\tilde{c} - \tilde{b})(c - a)$.

At this point, we derive the following results.

\begin{lemma}\label{lemma:gamma-bound}
Let $\bm{\mu} \in \Theta$. Then, for all $\bm{\mu'} \in \Theta$ it holds that:
\begin{align*}
\Gamma(\mu_k, \mu'_k) \le M \vert \mu_k - \mu'_k \vert \le M \sqrt{2 \sigma^2 d(\mu_k, \mu_k')}
\end{align*}
\end{lemma}
\begin{proof}
\begin{align*}
\Gamma(\mu_k, \mu'_k)
&= \sup_{\bm{\theta} \in \mathcal{S}} (d(\mu_k, \theta_k) - d(\mu'_k, \theta_k))
\\ & \le \sup_{\bm{\theta} \in \mathcal{S}} (\nu_{\mu_k} - \nu_{\theta_k})(\mu_k - \mu'_{k})
\\ & \le M \vert \mu_k - \mu'_k\vert
\\ & \le M \sqrt{2 \sigma^2 d(\mu_k, \mu_k')}
\: .
\end{align*}
where in the first step we have used Lemma \ref{lemma:divergence}, in the second one Assumption \ref{ass:bound} and the fact that $\bm{\mu} \in \mathcal{S}$, and in the last step Assumption \ref{ass:sub-gauss}.
\end{proof}

\subsection{If the algorithm has not stopped at $t$}

To inspect the behavior of our algorithm, let us introduce some additional notation. For each step $s$, we recall that $i^+(s) \in [K]$ denotes the arm that is optimal under the bandit model $\bm{\mu}^+(s)$, namely $i^+(s) \coloneqq i^{\star}(\bm{\mu}^+(s))$. Furthermore, for any $i \in [K]$, we denote by $q_i(t) = \sum_{s=\lceil \sqrt{t} \rceil}^t \bm{1} \{ i^+(s) = i \}$ as the number of times between $\lceil \sqrt{t} \rceil$ and $t$ such that the $i^+(s) = i$. Given this definition, we first provide a lower bound on $q_{\star}$ that holds on the good event $\mathcal{E}_t$. 

\begin{lemma}\label{lemma:fi-lb}
Under the good event $\mathcal{E}_t$ it holds that:
\begin{align*}
q_{\star}(t) \ge t - \sqrt{t} - \frac{2K f(t^2) \ln(t)}{T^\star(\bm{\mu})^{-1}}.
\end{align*}
\end{lemma}
\begin{proof}
Fix $i \in [K]$ and denote, for brevity, by $\lnot i$ the set of unimodal bandits where $i$ is not the optimal arm. Then, let us define the following quantity:
\begin{align*}
\epsilon_t^i \coloneqq \inf_{\bm{\theta} \in \lnot i} \sum_{s \ge \lceil \sqrt{t} \rceil, i^+(s) = i}^t \sum_{a \in [K]} \omega_i(s) d(\mu^+_a(s), \theta_a).
\end{align*}
Then, it holds that:
\begin{align*}
\epsilon_t^i & \ge \sum_{s \ge \lceil \sqrt{t} \rceil, i^+(s) = i}^t \inf_{\bm{\theta} \in \textup{Alt}(\bm{\mu^+}(s))} \sum_{a \in [K]} \omega_a(s) d(\mu^+_a(s), \theta_a) \\ & \ge \sum_{s \ge \lceil \sqrt{t} \rceil, i^+(s) = i}^t T^\star(\bm{\mu})^{-1},
\end{align*}
where, in the first step we have used the definition of infimum together with the fact that $i^+(s)=i$, and in the second one, we have used the definition of $\mathcal{E}_t$, and the fact that $\bm{\mu} \in \widetilde{\Theta}_s$ for $s \ge \lceil \sqrt{t} \rceil$. Therefore, have that:
\begin{align}\label{eq:fi-ub-1}
\sum_{i \ne \star} \epsilon_t^i \ge (t-\sqrt{t}-q_{\star}(t)) T^\star(\bm{\mu})^{-1}.
\end{align}

Then, we proceed by upper-bounding $\sum_{i \ne \star} \epsilon_i^t$. Specifically, we have that:
\begin{align*}
\sum_{i \ne \star} \epsilon_t^i &  = \sum_{i \ne \star} \inf_{\bm{\theta} \in \lnot i} \sum_{s \ge \lceil \sqrt{t} \rceil, i^+(s)=i}^t \sum_{a \in [K]} \omega_a(s) d(\mu_a^+(s), \theta_a) \\ & \le \sum_{i \ne \star} \sum_{s \ge \lceil \sqrt{t} \rceil, i^+(s)=i}^t \sum_{a \in [K]} \omega_a(s) d(\mu^+_a(s), \mu_a) \\ & \le f(t^2) \sum_{i \ne \star} \sum_{s \ge \lceil \sqrt{t} \rceil, i^+(s)=i}^t  \sum_{a \in [K]} \frac{\omega_a(s)}{N_a(s)} \\ & \le f(t^2)  \sum_{s \ge \lceil \sqrt{t} \rceil}^t  \sum_{a \in [K]} \frac{\omega_a(s)}{N_a(s)} \\ & \le 2K f(t^2) \ln(t),
\end{align*}
where in the first inequality we used the definition of infimum together with the fact that $\bm{\mu} \in \lnot i$, in the second one we used Lemma \ref{lemma:concentration-after-sqrt-t}, and in the last one Lemma \ref{lemma:tracking}. Combining this result with Equation \eqref{eq:fi-ub-1} yields the desired result.
\end{proof}

\begin{lemma}[If the algorithm has not stopped at $t$]\label{lemma:algo-not-stopped}
If the algorithm has not stopped at $t$, then, on the event $\mathcal{E}_t$, it holds that:
\begin{align*}
(t - \sqrt{t}) T^\star(\bm{\mu})^{-1} \le c(t-1,\delta) + h_1(t) + h_2(t) + h_3(t) + h_4(t).
\end{align*}
where:
\begin{align*}
& h_1(t) \coloneqq \min_{j \in \mathcal{N}(\star)} \left( \inf_{x \in (\mu_j, \mu_{\star})} \sum_{k \in \{\star, j \}} N_k(t) d(\mu_k, x)  - W_t(\star, j) \right) \\
& h_2(t) \coloneqq \min_{j \in \mathcal{N}(\star)} \inf_{x \in (\mu_j, \mu_{\star})} \sum_{j \in \{\star, j \}} \left( \sum_{s=1}^t \omega_{k}(s) - N_k(t) \right) d(\mu_k, x) \\
& h_3(t) \coloneqq \sum_{s \in [t]} \min_{j \in \mathcal{N}(i^+(s))} \inf_{x \in (\mu_j, \mu_{i^{+}(s)})} \sum_{k \in \{ i^+(s), j \}} \omega_k(s) d(\mu_k, x) - \sum_{s \in [t]} \min_{j \in \mathcal{N}(\star)} \inf_{x \in (\mu_j, \mu_{\star})} \sum_{k \in \{ \star, j \}} \omega_k(s) d(\mu_k, x)\\
& h_4(t) \coloneqq  \sum_{s \ge \sqrt{t}} \min_{j \in \mathcal{N}(i^+(s))} \inf_{x \in (\mu_j^+(s), \mu^+_{i^{+}(s)}(s))} \sum_{k \in \{ i^+(s), j \}} \omega_k(s) d(\mu_k^+(s), x) - \min_{j \in \mathcal{N}(i^+(s))} \inf_{x \in (\mu_j, \mu_{i^+(s)})} \sum_{k \in \{ j, i^+(s) \}} \omega_k(s) d(\mu_k, x) 
\end{align*}
%& h_1(t) \coloneqq \sup_{\bm{\theta} \in \mathcal{S}} \sum_{a \in [K]} N_a(t) \left( d(\mu_a, \theta_a) - d(\hat{\mu}_a(t), \theta_a) \right) \\
\end{lemma}
\begin{proof}
If the algorithm has not stopped at $t$, then it holds that:
\begin{align*}
c(t-1,\delta) & \ge \max_{i \in [K]} \min_{j \in \mathcal{N}(i)} W_t(i,j) \ge \min_{j \in \mathcal{N}(\star)} W_t(\star, j).
\end{align*}

At this point, we futher proceed by lower-bounding $\min_{j \in \mathcal{N}(\star)} W_t(\star, j)$. 
\begin{align*}
\min_{j \in \mathcal{N}(\star)} W_t(\star, j) & \ge \min_{j \in \mathcal{N}(\star)} \inf_{x \in (\mu_j, \mu_{\star})} \sum_{k \in \{\star, j \}} N_k(t) d(\mu_k, x) - h_1(t) \\ & \ge \sum_{s \in [t]} \min_{j \in \mathcal{N}(\star)} \inf_{x \in (\mu_j, \mu_{\star})} \sum_{k \in \{ \star, j \}} \omega_k(s) d(\mu_k, x) - h_1(t) - h_2(t) \\ & = \sum_{s \in [t]} \min_{j \in \mathcal{N}(i^+(s))} \inf_{x \in (\mu_j, \mu_{i^{+}(s)})} \sum_{k \in \{ i^+(s), j \}} \omega_k(s) d(\mu_k, x) - h_1(t) - h_2(t) - h_3(t) \\ & \ge \sum_{s \ge \sqrt{t}} \min_{j \in \mathcal{N}(i^+(s))} \inf_{x \in (\mu_j, \mu_{i^{+}(s)})} \sum_{k \in \{ i^+(s), j \}} \omega_k(s) d(\mu_k, x) - h_1(t) - h_2(t) - h_3(t) \\ & \ge \sum_{s \in \sqrt{t}} \min_{j \in \mathcal{N}(i^+(s))} \inf_{x \in (\mu_j^+(s), \mu^+_{i^{+}(s)}(s))} \sum_{k \in \{ i^+(s), j \}} \omega_k(s) d(\mu_k^+(s), x) - h_1(t) - h_2(t) - h_3(t) - h_4(t) \\ & = \sum_{s \ge \sqrt{t}} \min_{j \in \mathcal{N}(i^+(s))} g_j(\bm{\omega}(s), \bm{\mu}^+(s)) - h_1(t) - h_2(t) - h_3(t) - h_4(t) \\ & \ge \sum_{s \ge \sqrt{t}} \min_{j \in \mathcal{N}(\star)} g_j(\bm{\omega}^\star, \bm{\mu}) - h_1(t) - h_2(t) - h_3(t) - h_4(t) \\ & = (t - \sqrt{t})T^\star(\bm{\mu})^{-1} - h_1(t) - h_2(t) - h_3(t) - h_4(t)
\end{align*}
where the first steps follows from the definition of $h_1(t)$, $h_2(t)$, $h_3(t)$, and $h_4(t)$, while the last inequality follows from the definition of $\bm{\mu}^+(s)$ together with the fact that, on $\mathcal{E}_t$, $\bm{\mu} \in \widetilde{\Theta}_t$.

%while the last one from the fact the definition of $g$. Finally, it remains to analyze $\sum_{s \in [t]} \min_{j \in \mathcal{N}(i^+(s))} g_j(\bm{\omega}(s), \bm{\mu}^+(s))$. Specifically, we have that:
%\begin{align*}
%\sum_{s \in [t]} \min_{j \in \mathcal{N}(i^+(s))} g_j(\bm{\omega}(s), \bm{\mu}^+(s)) & = \sum_{s \le \sqrt{t}} \min_{j \in \mathcal{N}(i^+(s))} g_j(\bm{\omega}(s), \bm{\mu}^+(s)) + \sum_{s > \sqrt{t}}^t \min_{j \in \mathcal{N}(i^+(s))} g_j(\bm{\omega}(s), \bm{\mu}^+(s)) \\ & \ge \sum_{s \le \sqrt{t}} \min_{j \in \mathcal{N}(i^+(s))} g_j(\bm{\omega}(s), \bm{\mu}^+(s)) + \sum_{s > \sqrt{t}}^t T^\star(\bm{\mu})^{-1} \\ & \ge t T^\star(\bm{\mu})^{-1} - h_5(t)
%\end{align*}
%where, (i) in the first inequality we used the fact the definition of $\bm{\mu}^{+}(s)$ together with the fact that, due to Lemma \ref{lemma:concentration-after-sqrt-t}, $\bm{\mu} \in \widetilde{\Theta}_t$ on the good event $\mathcal{E}_t$, and (ii) in the second one we have used the definition of $h_5(t)$.
\end{proof}

\begin{lemma}[Upper bounds on $h(t)$]\label{lemma:bound-h}
Under the event $\mathcal{E}_t$ the following inequalities hold:
\begin{align*}
& h_1(t) \le 3 M \sqrt{2 \sigma^2 f(t)} \\ 
& h_3(t) \le C_2 \sqrt{t} + \frac{2C_3Kf(t^2)\ln(t)}{T^\star(\bm{\mu})^{-1}} \\ 
& h_4(t) \le M \sum_{s \ge \sqrt{t}} \max_{j \in \mathcal{N}(i^+(s))}  \sum_{k \in \{ i^+(s), j \}} \omega_k(s) \Big| \mu_k^+(s) - \mu_k \Big|.
\end{align*}
for some problem dependent constant $C_1, C_2 \ge 0$.
Furthermore, it holds that:
\begin{align*}
h_2(t) \le C_1 \ln(K),
\end{align*}
for some problem dependent constant $C_1 \ge 0$.
\end{lemma}
\begin{proof}
We begin by upper-bounding $h_1(t)$. Let $\bar{\eta} \in \argmin_{\eta \in (\hat{\mu}_j(t), \hat{\mu}_{\star}(t))}$. Then, on $\mathcal{E}_t$, we have that:
\begin{align*}
h_1(t) & \coloneqq \min_{j \in \mathcal{N}(\star)} \left( \inf_{x \in (\mu_j, \mu_{\star})} \sum_{k \in \{\star, j \}} N_k(t) d(\mu_k, x)  - W_t(\star, j) \right) \\ & \le \min_{j \in \mathcal{N}(\star)} \sum_{k \in (\mu_j, \mu_{\star}) } N_k(t) \left( d(\mu_k, \bar{\eta}) - d(\hat{\mu}_k(t), \bar{\eta}) \right) \\ & \le \min_{j \in \mathcal{N}(\star)} \sum_{k \in (\mu_j, \mu_{\star}) } N_k(t) \Gamma(\mu_k, \hat{\mu}_k(t)) \\ & \le M \min_{j \in \mathcal{N}(\star)} \sum_{k \in \{ \star, j \}} N_k(t) \sqrt{2 \sigma^2 d(\hat{\mu}_k(t), \mu_k)} \\ & \le 3 M \sqrt{2 \sigma^2 f(t)},
\end{align*}
where (i) the first and second inequalities follow from the definition of $\bar{\eta}$ and $\Gamma$ respectively, (ii) the third one from Lemma \ref{lemma:gamma-bound}, and the (iii) by event $\mathcal{E}_t$.
%\rp{Note: On step 2 there is the critical point mentioned above.}

Concerning $h_2(t)$, instead, we have that:
\begin{align*}
h_2(t) & \coloneqq \min_{j \in \mathcal{N}(\star)} \inf_{x \in (\mu_j, \mu_{\star})} \sum_{j \in \{\star, j \}} \left( \sum_{s=1}^t \omega_{k}(s) - N_k(t) \right) d(\mu_k, x) \\ & \le \ln(K) \min_{j \in \mathcal{N}(\star)} \inf_{x \in (\mu_j, \mu_{\star})} \sum_{j \in \{\star, j \}} d(\mu_k, x) \\ &  \coloneqq C_1 \ln(K)
\end{align*}

Concerning $h_3(t)$, instead, we recall that:
\begin{align}\label{eq:h3-def}
h_3(t) & \coloneqq \sum_{s \in [t]} \min_{j \in \mathcal{N}(i^+(s))} \inf_{x \in (\mu_j, \mu_{i^{+}(s)})} \sum_{k \in \{ i^+(s), j \}} \omega_k(s) d(\mu_k, x) - \sum_{s \in [t]} \min_{j \in \mathcal{N}(\star)} \inf_{x \in (\mu_j, \mu_{\star})} \sum_{k \in \{ \star, j \}} \omega_k(s) d(\mu_k, x).
\end{align}
Let us first consider $\sum_{s \le \sqrt{t}} \min_{j \in \mathcal{N}(i^+(s))} \inf_{x \in (\mu_j, \mu_{i^{+}(s)})} \sum_{k \in \{ i^+(s), j \}} \omega_k(s) d(\mu_k, x)$. Specifically, for some problem dependent constant $C_2$, we have that:
\begin{align*}
\sum_{s \le \sqrt{t}} \min_{j \in \mathcal{N}(i^+(s))} \inf_{x \in (\mu_j, \mu_{i^{+}(s)})} \sum_{k \in \{ i^+(s), j \}} \omega_k(s) d(\mu_k, x) \le C_2 \sqrt{t},
\end{align*}
We now analyze $\sum_{s > \sqrt{t}}^t \min_{j \in \mathcal{N}(i^+(s))} \inf_{x \in (\mu_j, \mu_{i^{+}(s)})} \sum_{k \in \{ i^+(s), j \}} \omega_k(s) d(\mu_k, x)$. Specifically, this quantity can be rewritten as:
\begin{align*}
\sum_{\substack{s > \sqrt{t} \\ i^+(s) \ne \star}}^t \min_{j \in \mathcal{N}(i^+(s))} \inf_{x \in (\mu_j, \mu_{i^{+}(s)})} \sum_{k \in \{ i^+(s), j \}} \omega_k(s) d(\mu_k, x) - \min_{j \in \mathcal{N}(\star)} \inf_{x \in (\mu_j, \mu_{\star})} \sum_{k \in \{ \star, j \}} \omega_k(s) d(\mu_k, x) \le \sum_{\substack{s > \sqrt{t} \\ i^+(s) \ne \star}}^t C_3,
\end{align*}
for some problem dependent constant $C_3$. Furthermore, on the good event $\mathcal{E}_t$ it holds that:
\begin{align*}
\sum_{\substack{s > \sqrt{t} \\ i^+(s) \ne \star}}^t C_3 = C_3 (t-\sqrt{t}-q_{\star}(t)) \le \frac{2C_3Kf(t^2)\ln(t)}{T^\star(\bm{\mu})^{-1}},
\end{align*}
where in the equality we have used the definition of $q_{\star}(t)$, and in the inequality step Lemma \ref{lemma:fi-lb}.

Concerning $h_4(t)$, instead, let $\bar{x}(s) \in \argmin_{x \in \mathbb{R}} \sum_{k \in \{ i^+(s), j \}} \omega_k(s) d(\mu_k, x)$. Then, we have that:
\begin{align*}
h_4(t) & \coloneqq  \sum_{s \ge \sqrt{t}} \min_{j \in \mathcal{N}(i^+(s))} \inf_{x \in (\mu_j^+(s), \mu^+_{i^{+}(s)}(s))} \sum_{k \in \{ i^+(s), j \}} \omega_k(s) d(\mu_k^+(s), x) - \min_{j \in \mathcal{N}(i^+(s))} \inf_{x \in (\mu_j, \mu_{i^+(s)})} \sum_{k \in \{ i^+(s),j \}} \omega_k(s) d(\mu_k, x) \\ & \le \sum_{s \ge \sqrt{t}} \max_{j \in \mathcal{N}(i^+(s))} \left( \inf_{x \in (\mu_j^+(s), \mu^+_{i^{+}(s)}(s))} \sum_{k \in \{ i^+(s), j \}} \omega_k(s) d(\mu_k^+(s), x) - \inf_{x \in (\mu_j, \mu_{i^+(s)})} \sum_{k \in \{ i^+(s),j \}} \omega_k(s) d(\mu_k, x) \right) \\ & = \sum_{s \ge \sqrt{t}} \max_{j \in \mathcal{N}(i^+(s))} \left( \inf_{x \in \mathbb{R}} \sum_{k \in \{ i^+(s), j \}} \omega_k(s) d(\mu_k^+(s), x) - \inf_{x \in \mathbb{R}} \sum_{k \in \{ i^+(s), j \}} \omega_k(s) d(\mu_k, x) \right) \\ & \le \sum_{s \ge \sqrt{t}} \max_{j \in \mathcal{N}(i^+(s))}  \sum_{k \in \{ i^+(s), j \}} \omega_k(s) \left( d(\mu_k^+(s), \bar{x}(s)) - d(\mu_k, \bar{x}(s)) \right) \\ & \le \sum_{s \ge \sqrt{t}} \max_{j \in \mathcal{N}(i^+(s))}  \sum_{k \in \{ i^+(s), j \}} \omega_k(s) \Gamma(\mu_k^+(s), \mu_k) \\ & \le M \sum_{s \ge \sqrt{t}} \max_{j \in \mathcal{N}(i^+(s))}  \sum_{k \in \{ i^+(s), j \}} \omega_k(s) \Big| \mu_k^+(s) - \mu_k \Big|,
\end{align*}
where in the last step, we have used Assumption \ref{ass:bound}.

%\rp{Note: Also in this last step, there is a slight abuse of the assumption for how it is formalized.}
\end{proof}

\begin{restatable}{lemma}{stop}\label{lemma:algo-not-stop-with-ub}
If the algorithm has not stop at $t$, then it holds that:
\begin{align*}
(t-\sqrt{t}) T^\star(\bm{\mu})^{-1} \le c(t-1,\delta) & + 3 M \sqrt{2 \sigma^2 f(t)} + C_1 \ln(K) + \\ & + C_2 \sqrt{t} + \frac{2C_3Kf(t^2)\ln(t)}{T^\star(\bm{\mu})^{-1}} + M \sum_{s \ge \sqrt{t}} \max_{j \in \mathcal{N}(i^+(s))}  \sum_{k \in \{ i^+(s), j \}} \omega_k(s) \Big| \mu_k^+(s) - \mu_k \Big|.
\end{align*}
\end{restatable}
\begin{proof}
Combine Lemma \ref{lemma:algo-not-stopped} and Lemma \ref{lemma:bound-h}.
\end{proof}

\subsection{Elimination of sub-optimal arms}\label{subsec-app:O-TaS-elim}
In this section, we provide a more in depth the behavior of O-TaS. Specifically, we show that under a good event, the number of pulls to arms whose oracle weights is limited.
To this end, we begin by providing properties of the $\widetilde{\Theta}_t$.

\begin{restatable}{proposition}{ciub}\label{prop:ci-upper-bound}
Fix $j \in [K]$. Suppose there exists $j' > j$ such that $\exists \tilde{j} \in \{j, \dots, j'-1 \}$ such that $\beta_{\tilde{j}}(t) < \alpha_{j'}(t)$. Then, for all $\bm{\mu} \in \widetilde{\Theta}_t$ it holds that:
\begin{align}\label{eq:ciub-eq-1}
\mu_j \le \min_{i \in \{j, \dots, j' \}} \beta_{i}(t) \quad \quad \quad \mu_j \ge \max_{i \le j} \alpha_i(t).
\end{align}
Furthermore, suppose there exists $j' < j$ such that $\exists \tilde{j} \in \{j'+1, \dots, j\}$ such that $\beta_{\tilde{j}}(t) < \alpha_{j'}(t)$. Then, for all $\bm{\mu} \in \widetilde{\Theta}_t$ it holds that:
\begin{align}\label{eq:ciub-eq-2}
\mu_j \le \min_{i \in \{j', \dots, j \}} \beta_{i}(t) \quad \quad \quad \mu_j \ge \max_{i \ge j} \alpha_i(t).
\end{align}
Finally, suppose there exists $j_1, j_2 \in [K]$ such that $j_1 < j < j_2$, and $\alpha_{j_1}(t) \ge \alpha_j(t) $ and $\alpha_{j_2}(t) \ge \alpha_j(t)$. Then, it holds that:
\begin{align}\label{eq:ciub-eq-3}
\mu_j \ge \min \{\alpha_{j_1}(t), \alpha_{j_2}(t) \}
\end{align}
\end{restatable}
\begin{proof}
We begin by proving Equation \eqref{eq:ciub-eq-1}. Thanks to the definition of $j'$ and $\tilde{j}$, it is easy to verify that $i^{\star}(\bm{\mu}) \ge j'$ holds for $\bm{\mu} \in \tilde{\Theta}_t$. Suppose there exists $\bm{\mu} \in \widetilde{\Theta}_t$ such that $\mu_j > \min_{i \in j, \dots, j'} \beta_i(t)$. This implies that $\mu_{j} > \mu_{i}$, but since $i^{\star}(\bm{\mu}) \ge j'$, this means that $\mu_{i} \le \mu_{i+1}$ holds for all $i \le i^{\star}(\bm{\mu})$, and, therefore, $\bm{\mu} \notin \widetilde{\Theta}_t$.
At this point, it remains to prove that $\mu_{j} \ge \max_{i \le j} \alpha_i(t)$ holds. Let us proceed by contradiciton and suppose that there exists $\bm{\mu} \in \widetilde{\Theta}_t$ such that $\mu_j < \alpha_i(t)$ for some $i < j$. This implies that $\mu_j < \alpha_i(t) \le \mu_i$. However, since $i^{\star}(\bm{\mu}) \ge j'$, this means that $\mu_{i} \le \mu_{i+1}$ holds for all $i \le i^{\star}(\bm{\mu})$, and, therefore, $\bm{\mu} \notin \widetilde{\Theta}_t$.

The proof of Equation \eqref{eq:ciub-eq-2} follows by simmetrical reasoning.

Finally, we prove Equation \eqref{eq:ciub-eq-3}. We proceed by contradiction, suppose that there exists $\bm{\mu} \in \widetilde{\Theta}_t$ such that $\mu_j < \min \{ \alpha_{j_1}(t), \alpha_{j_2}(t) \}$, that is $\mu_j < \alpha_{j_1}(t)$ and $\mu_j < \alpha_{j_2}(t)$. By definition of $\widetilde{\Theta}_t$, we also have that $\mu_{j_1} \ge \alpha_{j_1}(t)$ and $\mu_{j_2} \ge \alpha_{j_2}(t)$. Therefore, this would imply $\mu_j < \mu_{j_1}$ and $\mu_j < \mu_{j_2}$. However, since $j_1 < j < j_2$, this is impossible since the unimodal constraints are not respected, and therefore $\bm{\mu} \notin \widetilde{\Theta}_t$, thus concluding the proof.
\end{proof}

We now provide a lower bound to the number of pulls to that any arm $a \in \mathcal{N}(\star)$. 

\begin{lemma}\label{lemma:na-lb}
Let $a \in \mathcal{N}(\star)$. Then, there exists a problem dependent constant $C_{\bm{\mu}}$ such that, on $\mathcal{E}_t$, the following holds:
\begin{align}\label{eq:lb-n-a-1}
N_a(t) \ge C_{\bm{\mu}} \left( t - \sqrt{t} - \frac{2Kf(t^2)\ln(t)}{T^\star(\bm{\mu})^{-1}} \right) - \ln(K).
\end{align}
\end{lemma}
\begin{proof}
Due to Lemma \ref{lemma:tracking}, we have that:
\begin{align}\label{eq:lb-n-a-2}
N_a(t) \ge \sum_{s \in [t]} \omega_a(s) - \ln(K) \ge \sum_{s \ge \lceil  \sqrt{t} \rceil, i^+(s)=\star} \omega_a(s) - \ln(K).
\end{align} 
Furthermore, whenever $s \ge \lceil \sqrt{t} \rceil$ and $i^+(s) = \star$, the following inequalities hold thanks to the definition of $\bm{\mu}^+(s)$ and the fact that, on $\mathcal{E}_t$, $\bm{\mu} \in \widetilde{\Theta}_s$ (i.e., Lemma \ref{lemma:concentration-after-sqrt-t}):
\begin{align*}
& \mu^+_{\star}(s) \ge \mu_{\star} \ge \mu_{\star - 1} \ge \mu^+_{\star - 1}(s) \\
& \mu^+_{\star}(s) \ge \mu_{\star} \ge \mu_{\star + 1} \ge \mu^+_{\star + 1}(s)
\end{align*}
As a direct consequence of Theorem \ref{theo:sparsity}, this also implies that $\omega_a(s) > 0$ for all $a \in \mathcal{N}(\star)$. Plugging this result within Equation \eqref{eq:lb-n-a-2} and applying Lemma \ref{lemma:fi-lb} yields Equation \eqref{eq:lb-n-a-1}.
\end{proof}

Given this result, we provide the following Lemma that shows a peculiar aspect of the behavior of O-TaS. As we shall later see, this will play in optimizing second order terms in the analysis of the expected sample complexity of O-TaS.

\begin{lemma}\label{lemma:elimination}
Consider $a \in \mathcal{N}(\star)$ such that $a \ne \star$, and define $t^*_a \in \mathbb{N}$ as the smallest integer $n$ such that the following condition is satisfied:
\begin{align*}
\mu_{\star} - \mu_a > 4 \sqrt{\frac{2 \sigma^2 f(n^2)}{C_{\bm{\mu}} \left(n - \sqrt{n} - \frac{2K f(n^2) \ln(n)}{T^\star(\bm{\mu})^{-1}}\right) - \ln(K)}}.
\end{align*} 
Consider $t \ge t^*_a$. On the good event $\mathcal{E}_t$, there does not exists any unimodal bandit $\bm{\mu}' \in \widetilde{\Theta}_{t}$ such that $i^{\star}(\bm{\mu}') = a$. 
\end{lemma}
\begin{proof}
To prove the result, we notice that a necessary condition to have a unimodal bandit $\bm{\mu}' \in \widetilde{\Theta}_{{t}}$ such that $i^{\star}(\bm{\mu}') = a$ is that $\alpha_{i^{\star}(\bm{\mu})}({t}) \le \beta_{a}({t})$ holds at a certain step ${t}$. Nevertheless, we notice that, on the good event $\mathcal{E}_t$, the following inequalities holds for all $s \ge \lceil \sqrt{t} \rceil$:
\begin{align}
& \alpha_{i^{\star}(\bm{\mu})}(s) \ge \hat{\mu}_{i^{\star}(\bm{\mu})}(s) - \sqrt{\frac{2\sigma^2 f(s^2)}{N_{i^{\star}(\bm{\mu})}(s)}} \ge {\mu}_{i^{\star}(\bm{\mu})}(s) - 2 \sqrt{\frac{2\sigma^2 f(s^2)}{N_{i^{\star}(\bm{\mu})}(s)}} \label{eq:elim-1}\\
& \beta_a(s) \le \hat{\mu}_a(s) + \sqrt{\frac{2\sigma^2 f(s^2)}{N_{a}(s)}} \ge {\mu}_{a}(s) + 2 \sqrt{\frac{2\sigma^2 f(s^2)}{N_{a}(s)}}, \label{eq:elim-2}
\end{align}
where we have applied, in all the inequalities, Lemma \ref{lemma:concentration-after-sqrt-t}, together with Assumption \ref{ass:sub-gauss}.
From Equations \eqref{eq:elim-1}-\eqref{eq:elim-2}, and applying Lemma \ref{lemma:na-lb}, we obtain that $\alpha_{i^{\star}(\bm{\mu})}(s) > \beta_{a}(s)$ as soon as the following condition is verified:
\begin{align*}
\mu_{i^{\star}(\bm{\mu})} - \mu_a > 4 \sqrt{\frac{2 \sigma^2 f(s^2)}{C_{\bm{\mu}} \left(s - \sqrt{s} - \frac{2K f(s^2)\ln(s)}{T^\star(\bm{\mu})^{-1}}\right) - \ln(K)}},
\end{align*} 
which concludes the proof.
\end{proof}

As a direct consequence of Lemma \ref{lemma:elimination}, denote $t^* = \max_{i \in \mathcal{N}(\star)} t^*_i$. Then, on the good event $\mathcal{E}_t$, there is no bandit model $\bm{\mu}'$ in $\widetilde{\Theta}_t$ such that $i^{\star}(\bm{\mu}')=i$ for all $i \ne \star$.

\begin{theorem}\label{thm:all-arm-elim}
There exists $T \in \mathbb{N}$ such that for all $t \ge T$, on $\mathcal{E}_t$, $N_a(t) = N_a(T)$ for all $a \notin \mathcal{N}(\star)$.
\end{theorem}
\begin{proof}
Let $T$ be such that $T \ge t^*$. Then, for all $t \ge T$, on $\mathcal{E}_t$, $i^+(t) = \star$ (Lemma \ref{lemma:elimination}). Furthermore, due to Theorem \ref{theo:sparsity}, $\bm{\omega}(t)$ is sparse on $\mathcal{N}(\star)$. The results than follows due to the C-tracking sampling rule.
\end{proof}

\subsection{Sample Complexity Results}

Given the results of Section \ref{subsec-app:O-TaS-elim}, we now upper bound the last remaining term in Lemma \ref{lemma:algo-not-stop-with-ub}, that is $\sum_{s \ge \sqrt{t}} \max_{j \in \mathcal{N}(i^+(s))}\sum_{k \in \{i^+(s), j \}} \omega_k(s) \vert \mu^+_k(s) - \mu_k \vert$. Finally, we will provide a finite time analysis of O-TaS (which, in turn, will imply an asymptotic optimality result).

\begin{lemma}\label{lemma:h3-upper-bound}
Let us define $t^* = \max_{a \in \mathcal{N}(\star)} t^*_a$. 
On the good event $\mathcal{E}_t$ it holds that:
\begin{align*}
\sum_{s \ge \sqrt{t}} \max_{j \in \mathcal{N}(i^+(s))}\sum_{k \in \{i^+(s), j \}} \omega_k(s) \vert \mu^+_k(s) - \mu_k \vert \le \sqrt{2 \sigma^2 f(t^2)} \min \left\{ K\ln(K) + 2\sqrt{2Kt}, 2t^* + 3\ln(3) + 2\sqrt{6t} \right\}.
\end{align*} 
\end{lemma}
\begin{proof}
First of all, we have that:
\begin{align*}
\sum_{s \ge \sqrt{t}} \max_{j \in \mathcal{N}(i^+(s))}\sum_{k \in \{i^+(s), j \}} \omega_k(s) \vert \mu^+_k(s) - \mu_k \vert & \le  \sum_{s \ge \sqrt{t}} \max_{j \in \mathcal{N}(i^+(s))}\sum_{k \in \{i^+(s), j \}} \omega_k(s) \sqrt{2\sigma^2 d(\mu^+_k(s), \mu_k)}  \\ & \le \sqrt{2 \sigma^2 f(t^2)} \sum_{s \ge \sqrt{t}} \max_{j \in \mathcal{N}(i^+(s)} \sum_{k \in \{i^+(s), j \}} \frac{\omega_k(s)}{\sqrt{N_k(s)}} \\ & \le \sqrt{2 \sigma^2 f(t^2)} \left( K \ln(K) + 2 \sqrt{2Kt} \right),
\end{align*}
where, in the first step we have used Assumption \ref{ass:sub-gauss}, in the second one we have used Lemma \ref{lemma:concentration-after-sqrt-t}, and in the last one Lemma \ref{lemma:tracking}.
%\rp{Note: Also in this first step there is the issue.}

Furthermore, $\sum_{s \ge \sqrt{t}} \max_{j \in \mathcal{N}(i^+(s))}\sum_{k \in \{i^+(s), j \}} \omega_k(s) \vert \mu^+_k(s) - \mu_k \vert$ can be decomposed into the following terms:
\begin{align}
& \sum_{s = \lceil \sqrt{t} \rceil}^{t^*} \max_{j \in \mathcal{N}(i^+(s))}\sum_{k \in \{i^+(s), j \}} \omega_k(s) \vert \mu^+_k(s) - \mu_k \vert \label{eq:O-TaS-last-1} \\ 
& \sum_{s=t^* + 1}^t \max_{j \in \mathcal{N}(i^+(s))}\sum_{k \in \{i^+(s), j \}} \omega_k(s) \vert \mu^+_k(s) - \mu_k \vert.\label{eq:O-TaS-last-2}
\end{align}
We now first upper bound Equation \eqref{eq:O-TaS-last-1}. 
\begin{align*}
\sum_{s = \lceil \sqrt{t} \rceil}^{t^*} \max_{j \in \mathcal{N}(i^+(s))}\sum_{k \in \{i^+(s), j \}} \omega_k(s) \vert \mu^+_k(s) - \mu_k \vert & \le \sum_{s = \lceil \sqrt{t} \rceil}^{t^*} \max_{j \in \mathcal{N}(i^+(s))}\sum_{k \in \{i^+(s), j \}} \omega_k(s) \sqrt{2\sigma^2 d(\mu^+_k(s), \mu_k)} \\ & \le \sum_{s = \lceil \sqrt{t} \rceil}^{t^*} \max_{j \in \mathcal{N}(i^+(s))}\sum_{k \in \{i^+(s), j \}} \frac{\omega_k(s)}{\sqrt{N_k(s)}} \sqrt{2\sigma^2 f((t^*)^2)}  \\ & \le 2 t^* \sqrt{2\sigma^2 f((t^*)^2)} 
\end{align*}
where, in the first step we have used Assumption \ref{ass:sub-gauss} and in the second one Lemma \ref{lemma:concentration-after-sqrt-t}.
Finally, it remains to bound Equation \eqref{eq:O-TaS-last-2}. Specifically, we have that:
\begin{align*}
\sum_{s=t^* + 1}^t \max_{j \in \mathcal{N}(i^+(s))}\sum_{k \in \{i^+(s), j \}} \omega_k(s) \vert \mu^+_k(s) - \mu_k \vert & = \sum_{s=t^* + 1}^t \max_{j \in \mathcal{N}(\star)}\sum_{k \in \{\star, j \}} \omega_k(s) \vert \mu^+_k(s) - \mu_k \vert \\ & \le \sum_{s=t^* + 1}^t \max_{j \in \mathcal{N}(\star)}\sum_{k \in \{\star, j \}} {\omega_k(s)} \sqrt{2\sigma^2 d(\mu^+_k(s), \mu_k)} \\ & \le \sum_{s=t^* + 1}^t \max_{j \in \mathcal{N}(\star)}\sum_{k \in \{\star, j \}} \frac{\omega_k(s)}{\sqrt{N_k(s)}} \sqrt{2\sigma^2 f(t^2)} \\ &  \le \sum_{s=0}^t \sum_{j \in \mathcal{N}(\star)} \sum_{k \in \{\star, j \}} \frac{\omega_k(s)}{\sqrt{N_k(s)}} \sqrt{2\sigma^2 f(t^2)} \\ & \le \sqrt{2\sigma^2 f(t^2)} \left( 3 \ln(3) + 2 \sqrt{6t} \right),
\end{align*}
where in the first step we have used Lemma \ref{lemma:elimination}, in the second one Assumption \ref{ass:sub-gauss}, in the third one Lemma \ref{lemma:concentration-after-sqrt-t}, and in the last one Lemma \ref{lemma:tracking}.

%\rp{Note: Also in this second step there is the issue.}

Combining these results yields the result.
\end{proof}

\begin{theorem}[O-TaS Sample Complexity]\label{theo-cmplx-O-TaS}
Let $T_0(\delta)$ be defined as the first integer $t$ such that:
\begin{align*}
(t-\sqrt{t}) T^\star(\bm{\mu})^{-1} > c(t-1,\delta) & + 3 M \sqrt{2 \sigma^2 f(t)} + C_1 \ln(K) + \\ & + C_2 \sqrt{t} + \frac{2C_3Kf(t^2)\ln(t)}{T^\star(\bm{\mu})^{-1}} + \\ & + M \sqrt{2 \sigma^2 f(t^2)} \min \left\{ K\ln(K) + 2\sqrt{2Kt}, 2t^* + 3\ln(3) + 2\sqrt{6t} \right\}.
\end{align*}
Then, it holds that:
\begin{align*}
\E_{\bm{\nu}}[\tau_\delta] \le T_0(\delta) + \gamma.
\end{align*}
\end{theorem}
\begin{proof}
The proof of $\delta$-correctness is from Lemma \ref{lem:delta_correct_threshold}, which holds for any sampling rule.
At this point, combining Lemma \ref{lemma:algo-not-stop-with-ub} with Lemma \ref{lemma:h3-upper-bound}, we have that, if the algorithm has not stopped at $t$, then on the good event $\mathcal{E}_t$, $t \le T_0(\delta)$ holds. Therefore, for every $t > T_0(\delta)$, we have that $\mathcal{E}_t \subseteq \{\tau_\delta \le t \}$. Applying Lemma \ref{lem:lemma_1_Degenne19BAI}, together with Lemma \ref{lemma:prob-bad-event} concludes the proof.
\end{proof}

\begin{proof}[Proof of Theorem \ref{thm:SAOTaS_asymptotic_optimality}]
The proof is a direct consequence of Theorem \ref{theo-cmplx-O-TaS}. Specifically, due to Theorem \ref{theo-cmplx-O-TaS}, we have that:
\begin{align*}
\limsup_{\delta \rightarrow 0} \frac{\E_{\bm{\nu}}[\tau_\delta]}{\log(1/\delta)} \le T^\star(\bm{\mu}).
\end{align*}
\end{proof}

\subsection{Efficient Implementation}\label{subsec-app:O-TaS-impl}
We now discuss in detail how to efficiently implement the O-TaS routine. In this sense, we recall that the most crucial step is the computation of the following quantities:
\begin{align*}
	&\bm{\mu}^+(t) \in \argmax_{\bm{\lambda} \in \widetilde{\Theta}_t} \sup_{\bm{\omega} \in \widetilde{\Delta}_K(\bm{\lambda})} \min_{i \in \mathcal{N}(i^{\star}(\bm{\lambda}))} g_{i}(\bm{\omega}, \bm{\lambda}) \\
	&\bm{\omega}(t) \in \argmax_{\bm{\omega} \in \widetilde{\Delta}_K(\bm{\mu}^+(t))} \min_{i \in \mathcal{N}(i^{+}(t))}  g_{i}(\bm{\omega}, \bm{\mu}^+(t)) 
\end{align*}
We first show, for the sake of presentation, how to compute these quantities with a total complexity of $\mathcal{O}(K^2)$ (i.e., Algorithm \ref{alg:O-TaS-k-squared}), and we then discuss how to leverage dynamic programming in order to obtain a fast $\mathcal{O}(K)$ procedure. 

\begin{algorithm}[t]
	\caption{A $\mathcal{O}(K^2)$ procedure for O-TaS.}
    \label{alg:O-TaS-k-squared}
	\begin{algorithmic}
		\REQUIRE{$\Theta_t = \bigotimes_{i \in [K]} [\alpha_i(t), \beta_i(t)]$}
		\STATE{ {\color{blue} // Step 1: Compute $\bm{\mu}^{(i)}(t)$ for all $i \in [K]$ }}		
		\FOR{$i \in [K]$}
	 	\STATE{Initialize $\bm{\mu}^{(i)}(t) = \{ 0 \}_{i=1}^K$ }
		\STATE{Set ${\mu}^{(i)}_i(t) = \beta_i(t)$, $\mu^{(i)}_1(t) = \alpha_1(t)$, and $\mu^{(i)}_K(t) = \alpha_K(t)$ }	 	
	 	\FOR{$j \in \{1, \dots, i-1 \}$}
	 	\STATE{Set $\mu^{(i)}_j(t) = \max \{ \mu^{(i)}_{j-1}(t), \alpha_j(t) \}$}
	 	\ENDFOR
	 	\FOR{$j \in \{K-1, \dots, i + 1 \}$}
	 	\STATE{Set $\mu^{(i)}_j(t) = \max \{ \mu^{(i)}_{j+1}(t), \alpha_j(t) \}$}
	 	\ENDFOR
	 	\ENDFOR
	 	\STATE{ {\color{blue} // Step 2: Compute the set $\mathcal{Z}$ of valid unimodal bandits }}
	 	\STATE{ $\mathcal{Z} \coloneqq \{ i \in [K]: {\mu}^{(i)}_j(t) \ge {\mu}^{(i)}_{j-1}(t) ~\forall j \le i \textup{ and } {\mu}^{(i)}_j(t) \ge {\mu}^{(i)}_{j+1}(t) ~ \forall j \ge i \} $}
	 	\STATE{ {\color{blue} // Step 3: Return the optimistic model within $\mathcal{Z}$}}
	 	\RETURN{ $i \in \argmax_{j \in \mathcal{Z}} T^*(\bm{\mu}^{(j)}(t))^{-1}$}
	\end{algorithmic}
\end{algorithm}

Algorithm \ref{alg:O-TaS-k-squared} first iterates over the different arms in order to compute an optimistic bandit model in which arm $i$ is the optimal arm (Step 1). Then, it computes the set $\mathcal{Z} \subseteq [K]$ of indices of valid unimodal bandits. Indeed, the procedure at Step 1 might produce a bandit model $\bm{\mu}^{(i)}$ that violates the unimodal constraints.\footnote{This is related to the fact that, according to the value of $\Theta_t$, there might not be any valid unimodal bandit in which $i$ is the optimal arm.} Finally, it computes $i^+(t)$ by taking the argmax, over $j \in \mathcal{Z}$ of $T^*(\bm{\mu}^{(j)}(t))^{-1}$. Overall, the total complexity is controlled by Step 1 and Step 2, which both requires $\mathcal{O}(K^2)$ computations.\footnote{On the other hand, the computational complexity of Step 3 is $\mathcal{O}(K)$ since each $\bm{\mu}^{(i)}$ is a unimodal bandit.} In the following, we will show that this simple procedure is correct, i.e., Algorithm \ref{alg:O-TaS-k-squared} outputs $i^+(t)$.
 
First of all, we prove a basic property that connects $\mathcal{Z}$ with $\widetilde{\Theta}_t$. Specifically, we will show that $\mathcal{Z}$ is an empty set if and only if $\widetilde{\Theta}_{t}$ is empty.\footnote{Notice that, in this case, O-TaS can return any possible weight and any possible answer $i$. Indeed, as we have seen in the proof of Theorem \ref{theo-cmplx-O-TaS} this events happen with low probability.} Furthermore, we also show that $i \in \mathcal{Z}$ if and only if there is at least one valid unimodal bandit within $\widetilde{\Theta}_{t}$ such that $i$ is the optimal arm.
\begin{lemma}[When $\mathcal{Z}$ is empty]\label{lemma:z-empty-set}
It holds that:
\begin{align*}
\widetilde{\Theta}_t = \emptyset \iff \mathcal{Z} = \emptyset.
\end{align*}
Furthermore, denote by $\widetilde{\Theta}_{t,i}$ the subset of $\widetilde{\Theta}_t$ where $i$ is the optimal arm. Formally:  
\begin{align*}
\widetilde{\Theta}_{t,i} \coloneqq \left\{ \bm{\mu} \in \widetilde{\Theta}_t: i^{\star}(\bm{\mu}) = i \right\}.
\end{align*} 
Then, for all $i \in [K]$, it holds that:
\begin{align*}
i \in \mathcal{Z} \iff \widetilde{\Theta}_{t,i} \ne \emptyset.
\end{align*}
\end{lemma}
\begin{proof}
We first prove that $\widetilde{\Theta}_t = \emptyset \Longrightarrow \mathcal{Z} = \emptyset$. We proceed by contradiction. Suppose that $\mathcal{Z}$ is non-empty, then $\widetilde{\Theta}_t$ cannot be empty. Indeed, fix $i \in \mathcal{Z}$, then $\bm{\mu}^{(i)}(t) \in \widetilde{\Theta}_t$ by construction.

We now show that $\mathcal{Z} = \emptyset \Longrightarrow \widetilde{\Theta}_t = \emptyset$. We proceed by contradiction. Suppose that $\widetilde{\Theta}_t$ is non-empty. Then $\mathcal{Z}$ cannot be empty as well. Indeed, fix any $\bm{\mu} \in \widetilde{\Theta}_t$ and consider $i \in \argmax_{i \in [K]} \mu_i$. Then, $\bm{\mu}^{(i)}(t) \in \widetilde{\Theta}_t$ as well (and consequently, $i \in \mathcal{Z}$). Indeed, we recall that:
\begin{equation*}
\mu^{(i)}_j(t) =
    \begin{cases}
        \beta_j(t) & \text{if } j=i\\
        \max\{ \mu^{(i)}_{j-1}(t) , \alpha_{j}(t) \} & \text{if } j < i \\
        \max\{ \mu^{(i)}_{j+1}(t) , \alpha_{j}(t) \} & \text{if } j > i \\
    \end{cases}
\end{equation*}
To prove that $\bm{\mu}^{(i)}(t)$ belongs to $\widetilde{\Theta}_t$, one can start from $\bm{\mu}$ and modify $\mu_i$ such that $\mu'_i = \beta_i(t)$. The resulting vector is within $\widetilde{\Theta}_t$. Then, we can further modify this new vector by setting $\mu'_{K} = \alpha_K(t)$ and $\mu'_1 = \alpha_1(t)$. The resulting vector is within $\widetilde{\Theta}_t$. The reasoning can then be iterated to modify all the remaining arms so that, in the end, the resulting vector is equal to $\bm{\mu}^{(i)}$.

Finally, we prove $i \in \mathcal{Z} \iff \widetilde{\Theta}_{t,i} \ne \emptyset$. We first show that $i \in \mathcal{Z} \Longrightarrow \widetilde{\Theta}_{t,i} \ne \emptyset$. If $i \in \mathcal{Z}$, it holds that $\bm{\mu}^{(i)}(t) \in \widetilde{\Theta}_t$, and consequently in $\widetilde{\Theta}_{t,i}$ as its optimal arm is $i$.
Now, we show that $\widetilde{\Theta}_{t,i} \ne \emptyset \Longrightarrow i \in \mathcal{Z}$. Fix $\bm{\mu} \in \widetilde{\Theta}_{t,i}$. Modifying $\bm{\mu}$ as in the proof of $\mathcal{Z} = \emptyset \Longrightarrow \widetilde{\Theta}_t = \emptyset$, we obtain that $\bm{\mu}^{(i)} \in \widetilde{\Theta}_t$, and, consequently, $i \in \mathcal{Z}$, thus concluding the proof. 
\end{proof}

Given this first property, we now show the correctness of Algorithm \ref{alg:O-TaS-k-squared}.
\begin{lemma}\label{lemma:correct-eff}
It holds that:
\begin{align*}
& \max_{i \in \mathcal{Z}} T^*(\bm{\mu}^{(j)}(t))^{-1} = \bm{\mu}^+(t) \\
& \argmax_{i \in \mathcal{Z}} T^*(\bm{\mu}^{(j)}(t))^{-1} = i^+(t).
\end{align*}
\end{lemma}
\begin{proof}
If $\widetilde{\Theta}_t = \emptyset$, then, the claim is direct from Lemma \ref{lemma:z-empty-set}. We now consider the case in which $\widetilde{\Theta}_t \ne \emptyset$. Let Let us denote by $\widetilde{\Theta}_{t,i}$ the subset of $\widetilde{\Theta}_t$ where $i$ is the optimal arm. Formally:
\begin{align*}
\widetilde{\Theta}_{t,i} \coloneqq \left\{ \bm{\mu} \in \widetilde{\Theta}_t: i^{\star}(\bm{\mu}) = i \right\}.
\end{align*} 
Then, the result follows by noticing that, for all $i \in \mathcal{Z}$, and all $\bm{\mu} \in \widetilde{\Theta}_{t,i}$ it holds that:
\begin{align*}
T^*(\bm{\mu}^{(i)})^{-1} \ge T^\star(\bm{\mu})^{-1}.
\end{align*}
Indeed, from Theorem \ref{theo:sparsity}, we know that, for any $\bm{\mu}$, $T^\star(\bm{\mu})^{-1}$ is given by:
\begin{align*}
T^\star(\bm{\mu})^{-1} = \sup_{\bm{\omega} \in \widetilde{\Delta}_K(\bm{\mu})} \min_{a \in \mathcal{N}(\star)} g_{a}(\bm{\omega},\bm{\mu}),
\end{align*}
where $g_{a}(\bm{\omega},\bm{\mu}) \coloneqq \inf_{x \in (\mu_a, \mu_{\star})} \omega_{\star} d(\mu_{\star}, x) + \omega_a d(\mu_a, x)$. Now, fix $\bm{\omega} \in \widetilde{\Delta}_K(\bm{\mu})$ and $a \in \mathcal{N}(\star)$. Then, $g_{a}(\bm{\omega},\bm{\mu})$ is maximized for $\mu_i$ equal to $\mu_{i}^{(i)}$ and $\mu_a$ equal to $\mu_a^{(i)}$. This is consequence of (i) the convexity of the KL divergence, and (ii) the fact that $\mu_i \le \mu_{i}^{(i)}$ for all $\bm{\mu} \in \widetilde{\Theta}_{t,i}$ and $\mu_a \ge \mu_a^{(i)}$ for all $\bm{\mu} \in \widetilde{\Theta}_{t,i}$.

Finally, to conclude the proof, it remains to show that $i^+(s)$ is such that $i^+(s) \notin \mathcal{Z}$. However, this is a direct consequence of Lemma \ref{lemma:z-empty-set}.
\end{proof}

Given the correctness of Algorithm \ref{alg:O-TaS-k-squared}, we now present a more efficient algorithm that, by exploiting dynamic programming, it obtains a computational complexity of $\mathcal{O}(K)$. The underlying idea is that it is possible to compute, in $\mathcal{O}(1)$, $\bm{\mu}^{(i)}$ starting from $\bm{\mu}^{(i - 1)}$. Specifically, we have the following result. 
\begin{lemma}\label{lemma:dynamic-programming}
Let $i > 1$. Then, it holds that:
\begin{align*}
& \mu^{(i)}_i(t) = \beta_i(t) \\
& \mu^{(i)}_{i-1}(t) = \max \{ \alpha_{i-1}(t), \mu^{(i-1)}_{i-2}(t) \} \\
& \mu^{(i)}_j(t) = \mu^{(i-1)}_j(t) \quad \forall j \notin \{ i-1, i \}.
\end{align*} 
\end{lemma}
\begin{proof}
The proof is direct by looking at the definition of $\mu^{(i)}_j(t)$ that we have given in Algorithm \ref{alg:O-TaS-k-squared}. Specifically, we recall that $\bm{\mu}^{(i)}$ is given by:
\begin{equation*}
\mu^{(i)}_j(t) =
    \begin{cases}
        \beta_j(t) & \text{if } j=i\\
        \max\{ \mu^{(i)}_{j-1}(t) , \alpha_{j}(t) \} & \text{if } j < i \\
        \max\{ \mu^{(i)}_{j+1}(t) , \alpha_{j}(t) \} & \text{if } j > i.
    \end{cases}
\end{equation*}
\end{proof}
Lemma \ref{lemma:dynamic-programming} shows how to compute $\bm{\mu}^{(i)}(t)$ starting from $\bm{\mu}^{(i-1)}(t)$. Furthermore, we note that these two bandit models differ only at arms $i-1$ and $i$, which leads to a $\mathcal{O}(1)$ construction of $\bm{\mu}^{(i)}(t)$. 

Given these considerations, the detail that remains to be discussed is how to compute $\mathcal{Z}$ efficiently (i.e., Step 2 in Algorithm \ref{alg:O-TaS-k-squared}). In other words, we need to determine which is the subset of indexes for which $\bm{\mu}^{(i)}(t)$ are unimodal bandits. We note that, a basic implementation of this step would require $\mathcal{O}(K^2)$ iterations, as it takes $\mathcal{O}(K)$ iterations for each $i \in [K]$. Nevertheless, as for the computation of $\bm{\mu}^{(i)}(t)$, one can resort to dynamic programming to determine if $i \in \mathcal{Z}$ in constant time. Specifically, from Lemma \ref{lemma:dynamic-programming}, if $i-1 \in \mathcal{Z}$, then, $i \in \mathcal{Z}$ if an only if $\mu^{(i)}_i(t) > \mu^{(i)}_{i-1}(t)$. On the other hand, if $i-1 \notin \mathcal{Z}$, the situation is more complex. Let $\mathcal{V}_i \subseteq [K]$ be the subset of arms such that $\mu^{(i)}_j(t) \notin [\alpha_j(t), \beta_j(t)]$ (note that $\mathcal{V}_i$ can be easily computed from $\mathcal{V}_{i-1}$ in $\mathcal{O}(1)$ due to Lemma \ref{lemma:dynamic-programming}). Then, if $i-1 \notin \mathcal{Z}$, then $i \in \mathcal{Z}$ if and only if the following conditions are satisfied: (i) $\mathcal{V}_i = \emptyset$, and (ii) $\mu^{(i)}_i(t) > \mu^{(i)}_{i+1}(t)$ and $\mu^{(i)}_i(t) > \mu^{(i)}_{i-1}(t)$. We give a formal statement of these claims in the following lemma.

\begin{lemma}\label{lemma:dyn-prog-2}
Consider $i > 1$ and suppose that $i-1 \in \mathcal{Z}$. Then, $i \in \mathcal{Z} \iff \mu^{(i)}_i(t) > \mu^{(i)}_{i-1}(t)$. Furthermore, for all $i \in [K]$ let $\mathcal{V}_i$ be the subset of arms such that $\mu^{(i)}_j(t) \notin [\alpha_j(t), \beta_j(t)]$. Formally:
\begin{align*}
\mathcal{V}_i \coloneqq \{ j \in [K]: \mu^{(i)}_j(t) \notin [\alpha_j(t), \beta_j(t)] \}.
\end{align*}
Then, if $i-1 \notin \mathcal{Z}$, then $i \in \mathcal{Z}$ if and only if the following the following conditions are satisfied:
\begin{align}
& \mathcal{V}_i = \emptyset \label{eq:dyn-prog-2-eq-1} \\ 
& \mu^{(i)}_i(t) > \mu^{(i)}_{i+1}(t) \textup{ and } \mu^{(i)}_i(t) > \mu^{(i)}_{i-1}(t) \label{eq:dyn-prog-2-eq-2}.
\end{align} 
\end{lemma}
\begin{proof}
We start by considering $i-1 \in \mathcal{Z}$, and we prove that $i \in \mathcal{Z} \iff \mu^{(i)}_i(t) > \mu^{(i)}_{i-1}(t)$. To this end, we start by $i \in \mathcal{Z} \Longrightarrow \mu^{(i)}_i(t) > \mu^{(i)}_{i-1}(t)$. This is clearly true since, if $i \in \mathcal{Z}$, then $\bm{\mu}^{(i)}(t)$ respects the unimodal constraints, and, therefore, $\mu^{(i)}_i(t) > \mu^{(i)}_{i-1}(t)$. We now continue by showing that $\mu^{(i)}_i(t) > \mu^{(i)}_{i-1}(t) \Longrightarrow i \in \mathcal{Z}$. Fix $j \ge i$. Then, it holds that $\mu_j^{(i)}(t) \ge \mu_{j+1}^{(i)}(t)$ since (i) $\mu_i^{(i)}(t) = \beta_{i}(t) \ge \mu_i^{(i-1)}(t)$, $\mu_k^{(i)}(t) = \mu_k^{(i-1)}(t) ~\forall k > i$ (i.e., Lemma \ref{lemma:dynamic-programming}), and (ii) $i-1 \in \mathcal{Z}$. Now, fix $j < i-1$; then, since $i-1 \in \mathcal{Z}$ and due to Lemma \ref{lemma:dynamic-programming}, it holds that $\mu_j^{(i)}(t) \le \mu_{j+1}^{(i)}(t)$. Therefore, to prove $i \in \mathcal{Z}$ it remains to show that $\mu^{(i)}_i(t) > \mu^{(i)}_{i-1}(t)$, which, however, holds by assumption, thus concluding the proof.

We now continue by showing that, if $i-1 \notin \mathcal{Z}$, then $i \in \mathcal{Z}$ if and only if Equations \eqref{eq:dyn-prog-2-eq-1}-\eqref{eq:dyn-prog-2-eq-2} hold. We first show that $i \in \mathcal{Z}$ implies that Equations \eqref{eq:dyn-prog-2-eq-1}-\eqref{eq:dyn-prog-2-eq-2} hold. Nevertheless, this is direct by the definition of unimodal bandit and the fact that $\bm{\mu}^{(i)}(t)$ belongs to $\widetilde{\Theta}_t$. We now show that Equations \eqref{eq:dyn-prog-2-eq-1}-\eqref{eq:dyn-prog-2-eq-2} implies $i \in \mathcal{Z}$. First of all, we note that, due to the definition of $\bm{\mu}^{(i)}(t)$, if $\mathcal{V}_i = \emptyset$, then it means that, for all $j > i$, $\mu^{(i)}_j(t) \ge \mu^{(i)}_{j+1}(t)$, and for all $j < i-1$, then $\mu^{(i)}_j(t) \le \mu^{(i)}_{j+1}(t)$. Furthermore, since  $\mu^{(i)}_i(t) > \mu^{(i)}_{i+1}(t) \textup{ and } \mu^{(i)}_i(t) > \mu^{(i)}_{i-1}(t)$ holds, we have that $i \in \mathcal{Z}$, thus concluding the proof.
\end{proof}

% !TeX root = ../paper.tex

%\section{NON-ASYMPTOTIC ANALYSIS} 
\section{UNIMODAL TOP TWO}
\label{app:UniTT}

In Appendix~\ref{app:UniTT}, we prove our upper bounds on the expected sample complexity of UniTT holding for any $\delta$ and any Gaussian instances having mean $\boldsymbol{\mu} \in \cS$ (Theorem~\ref{thm:non_asymptotic_upper_bound_UniTT}).
Since we consider Gaussian distributions, we first recall some important properties on the $\beta$-characteristic times in Appendix~\ref{app:ssec_Gaussian_characteristic_times}.
In Appendix~\ref{app:ssec_key_properties}, we define the concentration events under which UniTT will be anaylyzed and show that the UCB leader is the best arm except for a sublinear number of times.
In Appendix~\ref{app:ssec_leader_stays_correct}, we provide an upper bound on the expected number of samples before the UCB leader matches the best arm forever.
In Appendix~\ref{app:ssec_overshooting_challenger}, we give an upper bound on the time after which the challenger is not overshooting its $\beta$-optimal allocation when $|\cN(i^\star)|=2$.
In Appendix~\ref{app:ss_proof_finite_time_upper_bound}, we conclude the proof of the three upper bounds contained in Theorem~\ref{thm:non_asymptotic_upper_bound_UniTT}. 

\subsection{Gaussian Characteristic Times}
\label{app:ssec_Gaussian_characteristic_times}

For Gaussian distributions with unit variance, the $\beta$-characteristic times $T^\star_{\beta}(\bm\mu)$ is defined in~\eqref{eq:beta_characteristic_time}.
Let $\bm \omega^{\star}_{\beta}(\bm\mu)$ denotes its associated with their $\beta$-optimal allocation.
The results recalled below are well-known in BAI (see e.g. ~\citet{jourdan2023NonAsymptoticAnalysis}).
The $\beta$-optimal allocation is unique.
Let $\bm \omega^{\star}_{\beta}$ denote its unique element.
We have that $\min_{i \in \cN(i^\star) \cup \{i^\star\}} \omega^{\star}_{\beta,i}>0$ and $\omega^{\star}_{\beta,i} = 0$ for all $i \notin \cN(i^\star) \cup \{i^\star\}$.
Moreover, it satisfies the equality at equilibrium
\begin{equation} \label{eq:equality_equilibrium}
	\forall i \in \cN(i^\star), \quad \frac{(\mu_{i^\star}- \mu_{i})^2}{1/\beta + 1/\omega^{\star}_{\beta,i}}  = 2T^{\star}_{\beta}(\bm{\mu})^{-1} \: .
\end{equation}
When $|\cN(i^\star)| = 1$, we have $T^{\star}(\bm{\mu}) = T^{\star}_{1/2}(\bm{\mu})$ and $\bm \omega^{\star}(\bm{\mu}) = \bm \omega^{\star}_{1/2}(\bm{\mu})$.

\begin{figure}[H]
    \centering
    \includegraphics[width=0.5\linewidth]{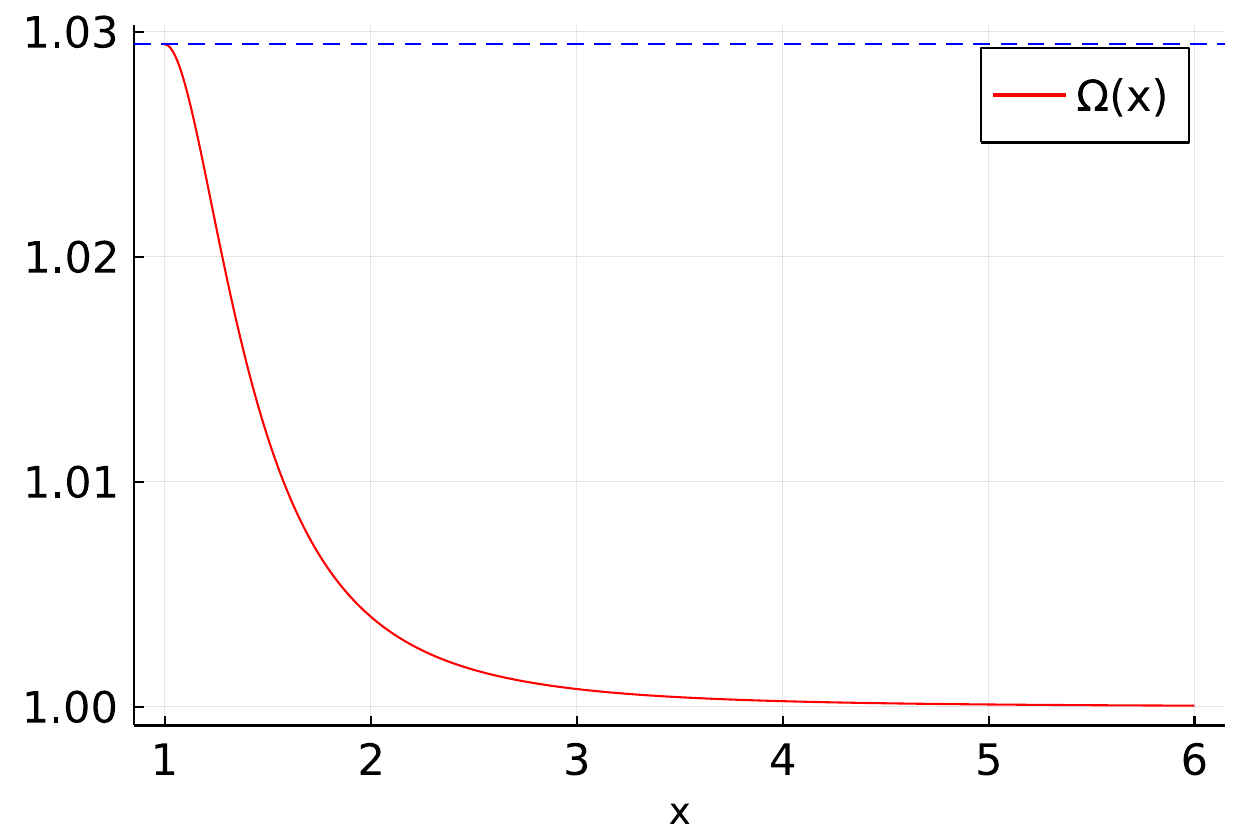}
    \caption{Ratio of characteristic times $T^\star_{1/2}(\bm{\mu}) / T^{\star}(\bm{\mu})$ when $|\cN(i^\star)|=2$ as a function of $x = (\mu_{i^\star} - \mu_{j})/(\mu_{i^\star} - \mu_{j^\star})$ where $j^\star \in \argmax_{i \in \cN(i^\star)} \mu_i$ and $j \in \cN(i^\star)\setminus \{j^\star\}$. The dashed blue line is $r_{2} = 6/(1+\sqrt{2})^2$.}
    \label{fig:simulation_ratio_times}
\end{figure}

When $|\cN(i^\star)| = 2$, let $j^\star \in \argmax_{j \in \cN(i^\star)} \mu_{j}$, $j \in \cN(i^\star) \setminus \{j^\star\}$ and $x \eqdef  \frac{ \mu_{i^\star} - \mu_{j}}{\mu_{i^\star} - \mu_{j^\star}}$.
The ratio $ \frac{T^\star_{1/2}(\bm\mu)}{T^\star(\bm\mu)}$ is independent of $\mu_{i^\star} - \mu_{j^\star}$, hence it is a function of $x$ only.
Let $\Omega : [1,+\infty) \to [1,+\infty)$ be defined as $\Omega(x) \eqdef  \frac{T^\star_{1/2}(\bm\mu)}{T^\star(\bm\mu)}$, and we plot it in Figure~\ref{fig:simulation_ratio_times} (Figure 2(a) in~\citet{jourdan2023NonAsymptoticAnalysis}).
Numerically, we see that $\Omega$ is a decreasing function achieving its maximum $r_3 = 6/(1+\sqrt{2})^2 \approx 1.029$ at $x = 1$, i.e. for neighbors having the same mean.
Lemma C.6 in~\citet{jourdan2023NonAsymptoticAnalysis} shows that $\Omega(1)  = r_{3}$ and $\Omega'(1) = 0$.
While $1$ is an extremal point of $\Omega$, more properties should be proven to show that it is the global maximum as suggested by Figure~\ref{fig:simulation_ratio_times}.

\subsection{Concentration Events and UCB Leader}
\label{app:ssec_key_properties}

\paragraph{Concentration Events}
Let $\gamma_{1} > 1$ and $\gamma_{2} > 1$.
Let $(\cE_{t})_{t > K}$ be the sequence of concentration events defined as $\cE_{t} \eqdef \cE_{1,t} \cap \cE_{2,t}$ for all $t > K$ where
\begin{equation} \label{eq:event_concentration_per_arm}
	\cE_{1,t} \eqdef \left\{ \forall k \in [K], \forall s \in [t^{1/\gamma_{1}}, t], \: |\hat{\mu}_{k}(t) - \mu_k| < \sqrt{\frac{f_{u}(t)}{N_{k}(t)}} \right\} \quad \text{with} \quad f_{u}(t) = 2\gamma_{1}(1+\gamma_{2}) \log t \quad \text{and} \quad
\end{equation}
\begin{equation}\label{eq:event_concentration_per_pair}
	 \cE_{2,t} \eqdef \left\{ \forall (i,k) \in (\cN(i^\star) \cup \{i^\star\})^2 \: \text{s.t.} \: i \ne k, \forall t \le n, \: \frac{|(\hat{\mu}_{i}(t) - \hat{\mu}_{k}(t)) - (\mu_{i} - \mu_{k})|}{\sqrt{1/N_{i}(t) + 1/N_{k}(t)}} < \sqrt{2 (2+\gamma_{2}) \log t} \right\} \: .
\end{equation}
We only concentrate pairwise comparison between arms in $\cN(i^\star) \cup \{i^\star\}$, which consists of three terms.
Lemma E.2 in~\citet{jourdan2023NonAsymptoticAnalysis} shows that $\bP (\cE_{1,t}^{\complement}) \le K t^{-\gamma_{2}}$ for all $t > K$.
Using the proof of Lemma 41 in~\citet{jourdan2023epsilonbestarm}, we have $\bP (\cE_{1,t}^{\complement}) \le 3 t^{-\gamma_{2}}$ for all $t > K$.
Therefore, we have $\sum_{t > K} \bP (\cE_{t}^{\complement}) \le (K+3)\zeta(s)$.

For all $t > K$, recall that $\Theta_t = \bigotimes_{i \in [K]}  [\hat{\mu}_{i}(t) \pm \sqrt{f_{u}(t)/ N_{i}(t)}]$.
Under $\cE_{1,t}$, we have $\boldsymbol{\mu} \in \Theta_s$ for all $s \in [t^{1/\gamma_{1}}, t]$.
Since $\boldsymbol{\mu} \in \cS$, we have $\boldsymbol{\mu} \in \widetilde \Theta_s \eqdef \Theta_s \cap \cS$.
Therefore, $\widetilde \Theta_s \ne \emptyset$ for all $s \in [t^{1/\gamma_{1}}, t]$.

\paragraph{Unimodal UCB Leader}
Lemma~\ref{lem:ucb_leader_lower_bound_counts} shows that the unimodal UCB leader is different from $i^\star$ for only a sublinear number of times under a certain concentration event.
\begin{lemma} \label{lem:ucb_leader_lower_bound_counts}
	Let $(\cE_{1,t})_{t}$ and $f_{u}$ as~\eqref{eq:event_concentration_per_arm}.
	Let $H_{1}(\boldsymbol{\mu}) \eqdef \sum_{i \ne i^\star(\boldsymbol{\mu})} (\mu_{i^\star} - \mu_{i})^{-2}$.
	Under the event $\cE_{1,t}$,
\begin{equation} \label{eq:ucb_leader_property}
	\sum_{s \in [t^{1/\gamma_{1}}, t]} \indi{B_s \ne i^\star} \le \frac{4 f_{u}(t)}{\beta} H_{1}(\boldsymbol{\mu})+ \frac{2(K-1)}{\beta}  \: .
\end{equation}
\end{lemma}
\begin{proof}
	Let $s \in [t^{1/\gamma_{1}}, t]$ and $k \ne i^\star$ such that $B_s = k$.	
	Under the event $\cE_{1,t}$, we have $\boldsymbol{\mu} \in \widetilde \Theta_s$ and the definition of $B_s$ in~\eqref{eq:UCB_leader} yields
\begin{align*}
\mu_{i^\star} = \max_{i \in [K]} \mu_{i} \le \max_{\boldsymbol{\lambda} \in \widetilde \Theta_{s}} \max_{i \in [K]} \lambda_i = \max_{\boldsymbol{\lambda} \in \widetilde \Theta_{s}}  \lambda_k
\le \hat{\mu}_{k}(s) + \sqrt{\frac{f_{u}(s)}{N_{k}(s)}}
\le \mu_k + \sqrt{\frac{4 f_{u}(s)}{N_{k}(s)}}
\le \mu_k + \sqrt{\frac{4 f_{u}(t)}{N_{k}(s)}}
\: .
\end{align*}
Lemma~\ref{lem:tracking_guaranties} yields that $N_{k}(s) \ge N_{k}^{k}(s) \ge \beta\sum_{j \in \cN(k)}T_{s}(k,j)  - 2$. 
By re-ordering the inequalities, we obtain
\[
	\sum_{j \in \cN(k)}T_{s}(k,j) \le  \frac{4 f_{u}(t)}{\beta (\mu_{i^\star} - \mu_{k})^2} + \frac{2}{\beta} \quad \text{and} \quad \sum_{j \in \cN(k)}T_{s+1}(k,j) = \sum_{j \in \cN(k)}T_{s}(k,j) + 1\: .
\]
Therefore, using Lemma~\ref{lem:simple_key_observation} for $\cA = [K] \setminus \{i^\star\}$ concludes the proof.
\end{proof}

\subsection{Upper Bound on the Expected Number of Samples Before Leader Stays Correct}
\label{app:ssec_leader_stays_correct}

Let us define the set of undersampled arms as
	\[
		U_{s}(t) = \left\{ i \in \cN(i^\star) \cup \{i^\star\} \mid N_{i}(s) \le \frac{16 f_{u}(t)}{\min_{i \in \cN(i)}(\mu_{i^\star} - \mu_{i})^2}\right\} \quad \text{where} \quad f_{u}(t) = 2\gamma_{1}(1+\gamma_{2}) \log t\: .
	\]
Lemma~\ref{lem:necessary_cdt_error} shows that a necessary condition for a unimodal instance $\lambda$ in our confidence region to have a best arm $i^\star(\boldsymbol{\lambda})$ which is different from $i^\star$ is that there are still undersampled arms in $\cN(i^\star) \cup \{i^\star\} $.
\begin{lemma} \label{lem:necessary_cdt_error}
	Let $(\cE_{1,t})_{t}$ as~\eqref{eq:event_concentration_per_arm}.
	Under $\cE_{1,t}$, there exists $\boldsymbol{\lambda} \in \widetilde \Theta_{t}$ such that $i^\star(\boldsymbol{\lambda}) \ne i^\star$ implies that $U_{t}(t) \ne \emptyset$.
\end{lemma}
\begin{proof}
	Under $\cE_{1,t}$, suppose that there exists $\boldsymbol{\lambda} \in \widetilde \Theta_{t}$ such that $i^\star(\boldsymbol{\lambda}) \ne i^\star$.
	This implies that
	\[
	\mu_{i^\star} - \sqrt{\frac{4f_{u}(t)}{N_{i^\star}(t)}} \le \hat{\mu}_{i^\star}(t) - \sqrt{\frac{f_{u}(t)}{N_{i^\star}(t)}} \le \max_{i \in \cN(i^\star)} \left\{ \hat{\mu}_{i}(t) + \sqrt{\frac{f_{u}(t)}{N_{i}(t)}} \right\}  \le \max_{i \in \cN(i^\star)} \left\{ \mu_{i} + \sqrt{\frac{4f_{u}(t)}{N_{i}(t)}} \right\} \: ,
	\]
	since $i^\star$ is the only possible choice otherwise. 
	For the two outer inequalities, we leverage the concentration event $\cE_{1,t}$.
	By rewriting this inequality, this implies that there exists $i \in \cN(i^\star)$ such that $\min\{N_{i}(t), N_{i^\star}(t)\} \le \frac{16 f_{u}(t)}{(\mu_{i^\star} - \mu_{i})^2}$, hence we have $U_{t}(t) \ne \emptyset$.
\end{proof}

Lemma~\ref{lem:undersampled_not_too_much} shows that the set of undersampled arms can be non-empty when the leader is the best arm for only a sublinear number of times under a certain concentration event.
\begin{lemma} \label{lem:undersampled_not_too_much}
	Let $d_{\gamma_{1},\gamma_{2}} = (2\sqrt{\gamma_{1}(1+\gamma_{2})}+\sqrt{2+\gamma_{2}})^2$.
	Let $(\cE_{1,t})_{t}$ and $(\cE_{2,t})_{t}$ as~\eqref{eq:event_concentration_per_arm} and~\eqref{eq:event_concentration_per_pair}.
	Let $H_{1}(\boldsymbol{\mu})$ as in Lemma~\ref{lem:ucb_leader_lower_bound_counts}.
	Under the event $\cE_{t} = \cE_{1,t} \cap \cE_{2,t}$,
	\begin{align} \label{eq:ucb_leader_optimal}
		&\sum_{s \in [t^{1/\gamma_{1}}, t]} \indi{B_s = i^\star, U_{s}(t) \ne \emptyset} \le \frac{32 d_{\gamma_{1},\gamma_{2}}}{\min\{\beta, 1-\beta\}\min_{i \in \cN(i^\star)}(\mu_{i^\star} - \mu_i)^2} \log t  + \frac{9}{\min\{\beta, 1-\beta\}} \: .
	\end{align}
\end{lemma}
\begin{proof}	
	Let $s \in [t^{1/\gamma_{1}}, t]$ such that $B_s = i^\star$ and $U_{s}(t) \ne \emptyset$.
	By definition in~\eqref{eq:localTCchallenger}, we know that $C_{t} \in \cN(i^\star)$.\\
	When $i^\star \in U_{s}(t)$, we satisfied the condition of Lemma~\ref{lem:simple_key_observation} since $ T_{s+1}(i^\star)= T_{s}(i^\star) + 1$ and
	\[
		 T_{s}(i^\star) \le \frac{1}{\min\{\beta, 1-\beta\}}(N_{i^\star}(s) + 3) \le \frac{16 f_{u}(t)}{\min\{\beta, 1-\beta\}\min_{i \in \cN(i^\star)}(\mu_{i^\star} - \mu_{i})^2} + \frac{3}{\min\{\beta, 1-\beta\}} \: ,
	\]
	where we used Lemma~\ref{lem:tracking_guaranties} for the first inequality.
	We assume that $i^\star \notin U_{s}(t)$. 
	When $C_{s} = i $ for $ i \in U_{s}(t)$, we satisfied the condition of Lemma~\ref{lem:simple_key_observation} since $ T_{s+1}(i)= T_{s}(i) + 1$ and
	\[
		T_{s}(i) \le \frac{1}{\min\{\beta, 1-\beta\}}(N_{i}(s) + 3) \le \frac{16 f_{u}(t)}{\min\{\beta, 1-\beta\}\min_{i \in \cN(i^\star)}(\mu_{i^\star} - \mu_{i})^2} + \frac{3}{\min\{\beta, 1-\beta\}}  \: ,
	\]
	where we used Lemma~\ref{lem:tracking_guaranties} for the first inequality.
	We assume that $C_{s} = i$ where $i \notin U_{s}(t)$. Then, under $\cE_{2,t}$, 
	\begin{align*}
		\frac{\hat{\mu}_{i^\star}(s) - \hat{\mu}_{i}(s)}{\sqrt{1/N_{i^\star}(s) + 1/N_{i}(s)}}  &\ge   \frac{\mu_{i^\star} - \mu_i}{\sqrt{1/N_{i^\star}(s)+ 1/N_{i}(s)}} - \sqrt{2(2+\gamma_{2})  \log t} \\
		&\ge   (\mu_{i^\star} - \mu_i)  \sqrt{\min\{N_{i^\star}(s) , N_{i}(s)\}} - \sqrt{2(2+\gamma_{2})  \log t}  \: .
	\end{align*}
	Let $i_0 \in U_{s}(t)$ for which we know that $i \ne i_0$. Then, under $\cE_{2,t}$, we have
	\begin{align*}
		\frac{\hat{\mu}_{i^\star}(s) - \hat{\mu}_{i_0}(s)}{\sqrt{1/N_{i^\star}(s) + 1/N_{i_0}(s)}}  \le   \frac{\mu_{i^\star} - \mu_{i_0}}{\sqrt{1/N_{i^\star}(s)+ 1/N_{i_0}(s)}} + \sqrt{2 (2+\gamma_{2})  \log t} &\le   (\mu_{i^\star} - \mu_{i_0}) \sqrt{N_{i_0}(s)}  + \sqrt{2(2+\gamma_{2})  \log t} \\
		&\le \sqrt{16 f_{u}(t)}  + \sqrt{2 (2+\gamma_{2})  \log t} \: .
	\end{align*}
	Leveraging the definition of $C_s = i$ in~\eqref{eq:localTCchallenger} where $W_t(i,j)$ as in~\eqref{eq:Gaussian_TC} and using Lemma~\ref{lem:tracking_guaranties}, we have 
	\begin{align*}
		     \min\{T_{s}(i^\star) , T_{s}(i) \} &\le  \frac{1}{\min\{\beta, 1-\beta\}}( \min\{N_{i^\star}(s) , N_{i}(s)\} +3)  \\
		     &\le \frac{\left( \sqrt{16 f_{u}(t)}  + \sqrt{8(2+\gamma_{2})  \log t} \right)^2}{\min\{\beta, 1-\beta\}\min_{i \in \cN(i^\star)}(\mu_{i^\star} - \mu_i)^2}  + \frac{3}{\min\{\beta, 1-\beta\}}  \: .
	\end{align*}
	Therefore, there exists $j \in \{i,i^\star\} = \{C_s,B_s\}$ such that $T_{s+1}(j) = T_{s}(j)+1$ and
	\[
	T_{s}(j) \le \frac{\left( \sqrt{16 f_{u}(t)}  + \sqrt{8(2+\gamma_{2})  \log t} \right)^2}{\min\{\beta, 1-\beta\}\min_{i \in \cN(i^\star)}(\mu_{i^\star} - \mu_i)^2}  + \frac{3}{\min\{\beta, 1-\beta\}} \: .
	\]
Therefore, using $f_{u}(t) = 2\gamma_{1} (1+\gamma_{2})\log t$ and Lemma~\ref{lem:simple_key_observation} for $\cA = \cN(i^\star) \cup \{i^\star\}$ concludes the proof.
\end{proof}

Lemma~\ref{lem:time_leader_correct} gives a deterministic time after which the leader will always be the best arm provided concentration holds, and an upper bound on the expected number of samples before the leader stays the best arm forever.
\begin{lemma} \label{lem:time_leader_correct}
	Let $(\cE_{1,t})_{t}$ and $(\cE_{2,t})_{t}$ as~\eqref{eq:event_concentration_per_arm} and~\eqref{eq:event_concentration_per_pair}.
	Let $H_{\gamma_{1},\gamma_{2}}(\boldsymbol{\mu}) \eqdef \gamma_{1} (1+\gamma_{2}) H_{1}(\boldsymbol{\mu}) + \frac{4 d_{\gamma_{1},\gamma_{2}}}{\min_{i \in \cN(i^\star)}(\mu_{i^\star} - \mu_i)^2}$.
	\[
		L_{\boldsymbol{\mu}} = \sup \left\{ t \mid t - 1 \le \lceil t^{1/\gamma_{1}} \rceil + \frac{8 H_{\gamma_{1},\gamma_{2}}(\boldsymbol{\mu})}{\min\{\beta, 1-\beta\}} \log t  + \frac{7 + 2K}{\min\{\beta, 1-\beta\}}   \right\}\: .
	\]
	Then, for all $t > L_{\boldsymbol{\mu}}$, under the event $\cE_{t} = \cE_{1,t} \cap \cE_{2,t}$, we have $B_{t} = i^\star$ and
	\[
	\beta(t - 1)  + 1 \ge N^{i^\star}_{i^\star}(t) \ge \beta(t - 1 - L_{\boldsymbol{\mu}}) - 2 \: .
	\]
	Moreover, we have $\bE_{\bm{\mu}}[\inf\{T \mid \forall t \ge T , \: B_{t} = i^\star\}] \le L_{\boldsymbol{\mu}} + 1 + (K+3)\zeta(s)$ where $\zeta$ is the Riemann $\zeta$ function.
\end{lemma}
\begin{proof}
	Building on Lemmas~\ref{lem:ucb_leader_lower_bound_counts} and~\ref{lem:undersampled_not_too_much}, we obtain that
	\begin{align*}
		&\sum_{s \in [t^{1/\gamma_{1}}, t]} \indi{U_{s}(t) \ne \emptyset} \le \frac{8 }{\min\{\beta, 1-\beta\}} \left( \gamma_{1} (1+\gamma_{2}) H_{1}(\boldsymbol{\mu}) + \frac{4 d_{\gamma_{1},\gamma_{2}}}{\min_{i \in \cN(i^\star)}(\mu_{i^\star} - \mu_i)^2} \right)\log t  + \frac{7 + 2K}{\min\{\beta, 1-\beta\}} \: .
	\end{align*}
	For all $i \in \cN(i^\star) \cup \{i^\star\}$, let us define $n_{i}(t) = \max \{s \in [t^{1/\gamma_{1}},t] \mid i \in U_{s}(t) \}$.
	By definition, we have $i \in U_{s}(t)$ for all $s \in [t^{1/\gamma_{1}}, n_{i}(t)] $ and  $i \notin U_{s}(t)$ for all $s \in (n_{i}(t),n] $.
	Therefore, we have $U_{s}(t) \ne \emptyset$ for all $s \in [t^{1/\gamma_{1}}, \max_{i \in \cN(i^\star) \cup \{i^\star\}} n_{i}(t)]$ and $U_{s}(t) = \emptyset$ for all $s > \max_{i \in \cN(i^\star) \cup \{i^\star\}} n_{i}(t)$, hence
	\begin{align*}
		 \max_{i \in \cN(i^\star) \cup \{i^\star\}} n_{i}(t) - \lceil t^{1/\gamma_{1}} \rceil - 1 = \sum_{s \in [t^{1/\gamma_{1}}, t]} \indi{U_{s}(t) \ne \emptyset} \: .
	\end{align*}
	Let $t > L_{\boldsymbol{\mu}}$. Then, we have $t > \max_{i \in \cN(i^\star) \cup \{i^\star\}} n_{i}(t)$ since
	\begin{align*}
	 t - \lceil t^{1/\gamma_{1}} \rceil - 1 &>  \frac{8 }{\min\{\beta, 1-\beta\}} \left( \gamma_{1} (1+\gamma_{2}) H_{1}(\boldsymbol{\mu}) + \frac{4 d_{\gamma_{1},\gamma_{2}}}{\min_{i \in \cN(i^\star)}(\mu_{i^\star} - \mu_i)^2} \right)\log t  + \frac{7 + 2K}{\min\{\beta, 1-\beta\}}  \\
	 & \ge \sum_{s \in [t^{1/\gamma_{1}}, t]} \indi{U_{s}(t) \ne \emptyset}  = \max_{i \in \cN(i^\star) \cup \{i^\star\}} n_{i}(t) - \lceil t^{1/\gamma_{1}} \rceil - 1 \: .
	\end{align*}
	Therefore, we obtain that $U_{t}(t) = \emptyset$ as $U_{t}(t) \subseteq U_{s}(t)$.
	Using Lemma~\ref{lem:necessary_cdt_error}, we obtain that there cannot exist $\boldsymbol{\lambda} \in \widetilde \Theta_{t}$ such that $i^\star(\boldsymbol{\lambda}) \ne i^\star$.
	This implies that $B_{t} = i^\star$.
	Therefore, we have shown that, for all $t > L_{\boldsymbol{\mu}}$, we have $B_{t} = i^\star$ under the event $\cE_{t}$.

	Using Lemma~\ref{lem:tracking_guaranties} with the above result, we can show directly that
	\begin{align*}
		\beta(t - 1)  + 1 \ge N^{i^\star}_{i^\star}(t) \ge \beta(t - 1 - L_{\boldsymbol{\mu}}) - 2 \: .
	\end{align*}

	Let $l_{\boldsymbol{\mu}} = \inf\{T \mid \forall t \ge T , \: B_{t} = i^\star\}$.
	Using a similar reasoning as for the proof of Lemma~\ref{lem:lemma_1_Degenne19BAI}, we have 
	\[
	\bE_{\bm{\mu}}[l_{\boldsymbol{\mu}}] = \sum_{t \ge 0} \bP_{\bm{\mu}}(l_{\boldsymbol{\mu}} > t) \le  L_{\boldsymbol{\mu}} + 1 + \sum_{t > L_{\boldsymbol{\mu}}} \bP_{\bm{\mu}}(\cE_{t}^{\complement}) \le L_{\boldsymbol{\mu}} + 1 + (K+3)\zeta(s) \: ,
	\]
	where we used the first part of the lemma.
\end{proof}

While Lemma~\ref{lem:time_leader_correct} can be used in the non-asymptotic analysis, it is possible to obtain tighter upper bounds for the case $\cN(i^\star) = \{j^\star\}$ by not using it directly.

\paragraph{Explicit Upper Bound}
For increased clarity, we provide a more explicit upper bound on $L_{\boldsymbol{\mu}}$ by splitting the time.
Let $\epsilon \in (0,1)$.
Then, it is direct to see that $L_{\boldsymbol{\mu}} \le \max\{\widetilde L_{\boldsymbol{\mu}} , L_{0}\}$ with
\begin{align*}
	\widetilde L_{\boldsymbol{\mu}} &= \sup \left\{ t \mid t  \le (1+\epsilon) \left( \frac{8 H_{\gamma_{1},\gamma_{2}}(\boldsymbol{\mu})}{\min\{\beta, 1-\beta\}} \log t  + \frac{9 + 2K}{\min\{\beta, 1-\beta\}}   \right)\right\}\\
	&\le h_{1}\left(\frac{8 (1+\epsilon)H_{\gamma_{1},\gamma_{2}}(\boldsymbol{\mu})}{\min\{\beta, 1-\beta\}} , \frac{(2K + 9 )(1+\epsilon)}{\min\{\beta, 1-\beta\}}  \right) = \cO(H_{1}(\boldsymbol{\mu}) \log H_{1}(\boldsymbol{\mu})) \: , \\
	L_{0} &= \sup \left\{ t \mid t^{(\gamma_{1}-1)/\gamma_{1}}  \le  \frac{1+\epsilon}{\epsilon} \right\} \le \left(\frac{2}{\epsilon} \right)^{\gamma_{1}/(1-\gamma_{1})} = \cO(1) \: ,
\end{align*}
where we used Lemma~\ref{lem:inversion_upper_bound} which defines $h_{1}(z,y) = z \overline{W}_{-1}(y/z + \log z) \approx_{z \to + \infty} z \log z$.

\subsection{Upper Bound on Overshooting Challenger When $|\cN(i^\star)| = 2$}
\label{app:ssec_overshooting_challenger}

Lemma~\ref{lem:time_challenger_not_oversampled} gives a deterministic time after which the challenger will never overshoot its optimal allocation provided concentration holds.
\begin{lemma} \label{lem:time_challenger_not_oversampled}
Let $\gamma \in (0,\min_{i \in \cN(i^\star)} \omega^{\star}_{\beta,i} )$.
	Let $(\cE_{1,t})_{t}$ and $(\cE_{2,t})_{t}$ as~\eqref{eq:event_concentration_per_arm} and~\eqref{eq:event_concentration_per_pair}, $L_{\boldsymbol{\mu}}$ as in Lemma~\ref{lem:time_leader_correct}, and $\cS_{t} = \left\{i \in \cN(i^\star) \mid \frac{N^{i^\star}_{i}(t) + 2}{\omega^{\star}_{\beta,i}+\gamma}  \ge t - 1 - L_{\boldsymbol{\mu}} \right\}$.
Let $h_{1}$ as in Lemma~\ref{lem:inversion_upper_bound}, $r(x) = 1+2(\sqrt{1+2x}-1)^{-1}$ and
\begin{align*}
	D_{\boldsymbol{\mu}} = h_{1} \left( \frac{4(2+\gamma_{2})(1-\beta)^2}{\gamma^2\beta^2}  T^{\star}_{\beta}(\bm{\mu}) , \frac{(L_{\boldsymbol{\mu}}+ 4)r \left( \gamma \beta /(1-\beta)\right)}{\min\{\beta, \min_{i \in \cN(i^\star)}\omega^{\star}_{\beta,i} -\gamma \}}  \right) \: .
\end{align*}
	Then, for all $t > D_{\boldsymbol{\mu}}$, under the event $\cE_{t} = \cE_{1,t} \cap \cE_{2,t}$, we have $C_{t} \notin \cS_{t}$.
\end{lemma}
\begin{proof}
	Let $t > L_{\boldsymbol{\mu}}$.
	Under the event $\cE_{t}$, Lemma~\ref{lem:time_leader_correct} yields that $B_{t} = i^\star$, hence $C_{t} \in \cN(i^\star)$.
	Lemma~\ref{lem:tracking_guaranties} yields $(N^{i^\star}_{i^\star}(t) + 2)/\beta \ge \sum_{i \in \cN(i^\star)} T_{t}(i^\star,i)  \ge t - 1 - L_{\boldsymbol{\mu}}$. When $\cS_{t} = \cN(i^\star) $, using that $\sum_{i \in \cN(i^\star)} \omega^{\star}_{\beta,i} = 1-\beta$, we obtain
\[
	\sum_{ i \in \cN(i^\star)} N^{i^\star}_{i}(t) \ge (1-\beta + 2 \gamma)(t - 1 - L_{\boldsymbol{\mu}}) - 4 \quad \text{and} \quad N^{i^\star}_{i^\star}(t) \ge \beta(t - 1 - L_{\boldsymbol{\mu}}) - 2 \: .
\]
Therefore, we have a contradiction for $t > D_{1,\boldsymbol{\mu}} =  (L_{\boldsymbol{\mu}} + 3)/\gamma  +1 \ge  L_{\boldsymbol{\mu}}\left(1+\frac{1}{2\gamma} \right) + 3/\gamma  +1$ since 
\[
	t - 1 \ge \sum_{ i \in \cN(i^\star) \cup \{i^\star\}} N^{i^\star}_{i}(t)  \ge (1+2\gamma)(t - 1 - L_{\boldsymbol{\mu}}) - 6 > t - 1 \: .
\] 
In the following, we consider $t > D_{1,\boldsymbol{\mu}} \ge L_{\boldsymbol{\mu}}$.
When $\cS_{t} = \emptyset$, then the result holds.
Suppose that $\cS_{t} \ne \emptyset$, hence $|\cS_{t} |=1$.
Let $i \in \cS_{t}$ and $j \in \cS_{t}  \setminus \{i\}$.
We aim at showing that $W_{t}(i^\star,i) > W_{t}(i^\star,j)$ where $ W_{t}(i,j)$ is defined in~\eqref{eq:Gaussian_TC}, hence $C_{t} = j \notin \cS_{t}$ by definition in~\eqref{eq:localTCchallenger}.
First, 
\[
	\max\left\{(t-1)\beta/N_{i^\star}^{i^\star}(t), (t-1)(\omega^{\star}_{\beta,i}+\gamma)/N_{i}^{i^\star}(t) \right\} \le \left( 1 - \frac{L_{\boldsymbol{\mu}} +2/\min\{\beta, \omega^{\star}_{\beta,i} +\gamma\}}{t - 1} \right)^{-1} \: .
\]
Therefore, we obtain
\begin{align*}
	&\frac{\hat{\mu}_{i^\star}(t) - \hat{\mu}_{i}(t)}{\sqrt{1/N_{i^\star}(t)+ 1/N_{i}(t)}} \ge \sqrt{t - 1}\frac{\mu_{i^\star} - \mu_{i}}{\sqrt{1/\beta + 1/( \omega^{\star}_{\beta,i} +\gamma )}} \frac{\sqrt{1/\beta + 1/( \omega^{\star}_{\beta,i} +\gamma )}}{\sqrt{\frac{\beta (t-1)}{\beta N_{i^\star}^{i^\star}(t)}+ \frac{(t-1)(\omega^{\star}_{\beta,i} +\gamma )}{(\omega^{\star}_{\beta,i} +\gamma )N_{i}^{i^\star}(t)}}} - \sqrt{2(2+\gamma_{2})  \log t} \\
	&\qquad \ge \frac{\mu_{i^\star} - \mu_{i}}{\sqrt{1/\beta + 1/( \omega^{\star}_{\beta,i} +\gamma )}} \sqrt{t - 1 - L_{\boldsymbol{\mu}} - 2/\min\{\beta, \omega^{\star}_{\beta,i} +\gamma \}} - \sqrt{2(2+\gamma_{2})  \log t}  \: .
\end{align*}
Then, using that $1 - \beta - \omega^{\star}_{\beta,i} -\gamma = \omega^{\star}_{\beta,j} - \gamma $, we obtain
\begin{align*}
	N_{j}(t) &\le t - 1 - N_{i^\star}^{i^\star}(t) - N_{i}^{i^\star}(t) \le  (t - 1)( \omega^{\star}_{\beta,j} - \gamma ) + (\beta + \omega^{\star}_{\beta,i} + \gamma) L_{ \mu}+  4 \: .
\end{align*}
Likewise, we have
\begin{align*}
	N_{i^\star}(t) &\le \beta \sum_{i \in \cN(i^\star)} T_{t}(i^\star, i) + (1-\beta)\sum_{i \in \cN(i^\star)} T_{t}(i,i^\star) + 3 \le \beta (t-1) + (1-\beta) L_{\boldsymbol{\mu}} + 3 \: .
\end{align*}
Then, we obtain that
\begin{align*}
	&\min\{(t-1)\beta/N_{i^\star}(t) , (t-1)(\omega^{\star}_{\beta,j} - \gamma )/N_{j}(t) \} \ge \left( 1 + \frac{ (L_{\boldsymbol{\mu}}+ 4)/\min\{\beta,\omega^{\star}_{\beta,j} - \gamma \}}{t - 1} \right)^{-1} \: .
\end{align*}
Therefore, we obtain
\begin{align*}
	&\frac{\hat{\mu}_{i^\star}(t) - \hat{\mu}_{j}(t)}{\sqrt{1/N_{i^\star}(t)+ 1/N_{j}(t)}} \le \sqrt{t - 1}\frac{\mu_{i^\star} - \mu_{j}}{\sqrt{1/\beta + 1/( \omega^{\star}_{\beta,j} - \gamma  )}} \frac{\sqrt{1/\beta + 1/( \omega^{\star}_{\beta,j} - \gamma  )}}{\sqrt{\frac{\beta (t-1)}{\beta N_{i^\star}(t)}+ \frac{(t-1)(\omega^{\star}_{\beta,j} - \gamma )}{(\omega^{\star}_{\beta,j} - \gamma ) N_{j}(t)}}} + \sqrt{2(2+\gamma_{2})  \log t} \\
	&\qquad \le \frac{(\mu_{i^\star} - \mu_{j})}{\sqrt{1/\beta + 1/( \omega^{\star}_{\beta,j} - \gamma )}}  \sqrt{t - 1 +(L_{\boldsymbol{\mu}}+ 4)/\min\{\beta,\omega^{\star}_{\beta,j} - \gamma \}}  + \sqrt{2(2+\gamma_{2})  \log t}  \: .
\end{align*}
Therefore, we have $W_{t}(i^\star,i) > W_{t}(i^\star,j)$ for $t > \max\{D_{2,\boldsymbol{\mu}}, D_{1,\boldsymbol{\mu}}, L_{\boldsymbol{\mu}}\}$ where 
\[
	D_{2,\boldsymbol{\mu}} = \sup\{t \mid b_{\boldsymbol{\mu}} \sqrt{t - 1 + c_{\boldsymbol{\mu}}} \ge a_{\boldsymbol{\mu}} \sqrt{t - 1 - c_{\boldsymbol{\mu}}} - \sqrt{8(2+\gamma_{2})  \log t}\} \: ,
\]
with $c_{\boldsymbol{\mu}} = (L_{\boldsymbol{\mu}}+ 4)/\min\{\beta, \omega^{\star}_{\beta,j} -\gamma, \omega^{\star}_{\beta,i} +\gamma \}$,
\begin{align*}
	a_{\boldsymbol{\mu}} &= \frac{\mu_{i^\star} - \mu_{i}}{\sqrt{1/\beta + 1/( \omega^{\star}_{\beta,i} +\gamma )}} \quad \text{and} \quad
	b_{\boldsymbol{\mu}} = \frac{(\mu_{i^\star} - \mu_{j})}{\sqrt{1/\beta + 1/( \omega^{\star}_{\beta,j} - \gamma )}} \: .
\end{align*}
Combining the equality at equilibrium~\eqref{eq:equality_equilibrium} and the fact that transportation costs are increasing function of their allocations, we know that $a_{\boldsymbol{\mu}} > b_{\boldsymbol{\mu}}$ hence $D_{\boldsymbol{\mu}} < + \infty$.
For increased clarity, we provide a more explicit upper bound.
For $x\ge 0$, direct manipulation yields
\begin{align*}
	\left( a_{\boldsymbol{\mu}} \sqrt{x - c_{\boldsymbol{\mu}}} - b_{\boldsymbol{\mu}} \sqrt{x + c_{\boldsymbol{\mu}}}  \right)^2 = (a_{\boldsymbol{\mu}} - b_{\boldsymbol{\mu}})^2 x - c_{\boldsymbol{\mu}}(a_{\boldsymbol{\mu}}^2 - b_{\boldsymbol{\mu}}^2) + 2a_{\boldsymbol{\mu}} b_{\boldsymbol{\mu}} \left(x - \sqrt{x^2-c_{\boldsymbol{\mu}}^2}\right) \: .
\end{align*}
Therefore, we have 
\begin{align*}
	D_{2,\boldsymbol{\mu}} &\le \sup\left\{t \mid t - 1 \le \frac{8(2+\gamma_{2})}{(a_{\boldsymbol{\mu}} - b_{\boldsymbol{\mu}})^2}  \log t + c_{\boldsymbol{\mu}}\left(1 + 2 \left(\frac{a_{\boldsymbol{\mu}}}{b_{\boldsymbol{\mu}}}-1\right)^{-1}\right)\right\}  \: ,
\end{align*}
where we used that $\frac{a+b}{a-b} = 1 + 2(a/b-1)^{-1}$
Note that $a_{\boldsymbol{\mu}} =  f(\gamma)$ and $b_{\boldsymbol{\mu}} = g(\gamma)$ with $f(\gamma) = \frac{\mu_{i^\star} - \mu_{i}}{\sqrt{1/\beta + 1/(\omega^{\star}_{\beta,i} +\gamma )}} $ and $g(\gamma) = \frac{\mu_{i^\star} - \mu_{j}}{\sqrt{1/\beta + 1/( \omega^{\star}_{\beta,j} - \gamma )}}  $. 
The equality at equilibrium~\eqref{eq:equality_equilibrium} yields that $f(0)^2 = g(0)^2 = 2 T^{\star}_{\beta}(\bm{\mu})^{-1}$.
Then, we have
\begin{align*}
	\frac{f(\gamma)^2}{f(0)^2} - 1 &= \gamma \frac{\beta}{\omega^{\star}_{\beta,i}} \left( \beta + \omega^{\star}_{\beta,i} + \gamma\right)^{-1}  \ge \gamma \frac{\beta}{\omega^{\star}_{\beta,i}} \ge \gamma \frac{\beta}{1-\beta} \: , \\
	\frac{g(\gamma)^2}{g(0)^2} - 1 &= -\gamma \frac{\beta}{\omega^{\star}_{\beta,j}}  \left( \beta + \omega^{\star}_{\beta,j} - \gamma\right)^{-1} \le -\gamma \frac{\beta}{\omega^{\star}_{\beta,j}} \le - \gamma \frac{\beta}{1-\beta} \: .
\end{align*}
Using that $(\sqrt{1+x} - \sqrt{1-x})^2 = 2(1-\sqrt{1-x^2})\ge x^2$ for $x \in [0,1]$, we have
\begin{align*}
	(a_{\boldsymbol{\mu}} - b_{\boldsymbol{\mu}})^{-2}  &\le \frac{(1-\beta)^2}{2\gamma^2\beta^2}  T^{\star}_{\beta}(\bm{\mu}) \quad \text{and} \quad 
	\left(\frac{a_{\boldsymbol{\mu}}}{b_{\boldsymbol{\mu}}}-1\right)^{-1} \le \left(\sqrt{1+\frac{2\gamma \beta}{1-\beta}}-1\right)^{-1}\: .
\end{align*}
Let $r(x) = 1+2(\sqrt{1+2x}-1)^{-1}$.
Therefore, we have 
\begin{align*}
	D_{2,\boldsymbol{\mu}} &\le \sup\left\{t \mid t - 1 \le  \frac{4(2+\gamma_{2})(1-\beta)^2}{\gamma^2\beta^2}  T^{\star}_{\beta}(\bm{\mu})\log t + \frac{(L_{\boldsymbol{\mu}}+ 4)r \left( \gamma \beta /(1-\beta)\right)}{\min\{\beta, \min_{i \in \cN(i^\star)}\omega^{\star}_{\beta,i} -\gamma \}}  \right\} \\
	&\le D_{\boldsymbol{\mu}} = h_{1} \left( \frac{4(2+\gamma_{2})(1-\beta)^2}{\gamma^2\beta^2}  T^{\star}_{\beta}(\bm{\mu}) ,  \frac{(L_{\boldsymbol{\mu}}+ 4)r \left( \gamma \beta /(1-\beta)\right)}{\min\{\beta, \min_{i \in \cN(i^\star)}\omega^{\star}_{\beta,i} -\gamma \}}   \right) \: ,
\end{align*}
where $h_{1}(z,y)$ as in Lemma~\ref{lem:inversion_upper_bound}.
It is direct to see that $D_{\boldsymbol{\mu}} \ge D_{1,\boldsymbol{\mu}} =  (L_{\boldsymbol{\mu}} + 3)/\gamma  +1$, e.g. notice that $r(x) \ge 1/x$.
\end{proof}

Lemma~\ref{lem:time_neighbors_controlled} shows that the empirical allocations of the neighbors $\cN(i^\star)$ are close to their optimal allocation provided concentration holds.
\begin{lemma} \label{lem:time_neighbors_controlled}
	Let $\gamma \in (0,\min_{i \in \cN(i^\star)} \omega^{\star}_{\beta,i} )$, $(\cE_{1,t})_{t}$ and $(\cE_{2,t})_{t}$ as~\eqref{eq:event_concentration_per_arm} and~\eqref{eq:event_concentration_per_pair}, $L_{\boldsymbol{\mu}}$ and $D_{\boldsymbol{\mu}}$ as in Lemmas~\ref{lem:time_leader_correct} and~\ref{lem:time_challenger_not_oversampled}.
	Then, for all $t > D_{\boldsymbol{\mu}} (\gamma)$, under the event $\cE_{t} = \cE_{1,t} \cap \cE_{2,t}$, for all $i \in \cN(i^\star)$,
	\[
	(\omega^{\star}_{\beta,i} - \gamma)(t - 1- L_{\boldsymbol{\mu}})  - 2D_{\boldsymbol{\mu}}  \le  N^{i^\star}_{i}(t) \le (\omega^{\star}_{\beta,i} + \gamma)(t - 1-L_{\boldsymbol{\mu}})+ D_{\boldsymbol{\mu}} \: .
	\]
\end{lemma}
\begin{proof}
	Let $t \ge n > D_{\boldsymbol{\mu}} \ge L_{\boldsymbol{\mu}}$.
	Using Lemma~\ref{lem:time_leader_correct}, we know that $\beta(t - 1)  + 1 \ge N^{i^\star}_{i^\star}(t) \ge \beta(t - 1 - L_{\boldsymbol{\mu}}) - 2$.
	For all $i \in \cN(i^\star)$, let $n_{t}(i) = \max\left\{n \ne t \mid \frac{N_{i}^{i^\star}(t)+2}{\omega^{\star}_{\beta,i} + \gamma} \le t - 1-L_{\boldsymbol{\mu}}\right\}$.
	Using Lemma~\ref{lem:time_challenger_not_oversampled}, we know that
	\[
	\frac{N_{i}^{i^\star}(t)+2}{\omega^{\star}_{\beta,i} + \gamma} > t - 1-L_{\boldsymbol{\mu}} \quad  \implies  \quad  i \in \cS_{n}  \quad  \implies  \quad  C_{n} \ne i \: .
	\]
	Then, it is direct to show that
	\begin{align*}
		N^{i^\star}_{i}(t) &\le D_{\boldsymbol{\mu}} + \sum_{n = D_{\boldsymbol{\mu}}+1}^{t} \indi{\frac{N_{i}^{i^\star}(t)+2}{\omega^{\star}_{\beta,i} + \gamma} \le t - 1-L_{\boldsymbol{\mu}}} \indi{B_n = i^\star, C_{n} = i = I_{n}}\\
		&\le N^{i^\star}_{i}(n_{t}(i)) + D_{\boldsymbol{\mu}} \le (\omega^{\star}_{\beta,i} + \gamma)(t - 1-L_{\boldsymbol{\mu}})+ D_{\boldsymbol{\mu}}  -2 \: .
	\end{align*}
	Let $i\in \cN(i^\star)$ and $j \in \cN(i^\star)\setminus\{i\}$.
	Then, we have
	\begin{align*}
		N^{i^\star}_{i}(t) &= \sum_{i \in \cN(i^\star)} T_{t}(i^\star,i) - N^{i^\star}_{i^\star}(t) - N_{j}^{i^\star}(t) \ge (\omega^{\star}_{\beta,i} - \gamma)(t - 1- L_{\boldsymbol{\mu}})  - D_{\boldsymbol{\mu}} - L_{\boldsymbol{\mu}} + 1 \: .
	\end{align*}
	This concludes the proof.
\end{proof}

\subsection{Non-Asymptotic Upper Bounds}
\label{app:ss_proof_finite_time_upper_bound}

Let $t > K$ such that $\cE_{t}$ holds and the algorithm has not stop yet, i.e. $\cE_{t} \cap \{t < \tau_{\delta}\}$.
Let $s \in [t^{1/\gamma_{1}}, t]$ such that $B_s = i^\star$.
Let $c(t,\delta)$ as in~\eqref{eq:stopping_threshold}, which satisfies that $t \mapsto c(t,\delta)$ is increasing.
Using the stopping rule~\eqref{eq:GLR_stopping_rule} where $ W_{t}(i,j)$ is defined in~\eqref{eq:Gaussian_TC} and $s \le t < \tau_{\delta}$, we obtain
\begin{align*}
	\sqrt{2c(t-1,\delta)} \ge \sqrt{2c(s-1,\delta)} \ge \min_{i \in \cN(\hat \imath_s)} \frac{(\hat{\mu}_{\hat \imath_s}(s) - \hat{\mu}_{i}(s))_{+}}{\sqrt{1/N_{\hat \imath_s}(s)+ 1/N_{i}(s)}} &\ge \min_{i \in \cN(B_s)} \frac{(\hat{\mu}_{B_s}(s) - \hat{\mu}_{i}(s))_{+}}{\sqrt{1/N_{B_s}(s)+ 1/N_{i}(s)}} \\
	&= \frac{(\hat{\mu}_{B_s}(s) - \hat{\mu}_{C_s}(s))_{+}}{\sqrt{1/N_{B_s}(s)+ 1/N_{C_s}(s)}} \: ,
\end{align*}
where the equality is obtained by definition of $C_s$ as in~\eqref{eq:localTCchallenger}.
The last inequality is true by definition of the candidate answer $\hat \imath_s$ as in~\eqref{eq:IFanswer} since $\hat \imath_s \in \argmax_{i \in [K]} \min_{j \in \cN(i)} W_{s}(i,j)$ implies that $\min_{j \in \cN(\hat \imath_s)} W_{s}(\hat \imath_s,j) \ge \min_{j \in \cN(B_s)} W_{s}(B_s,j)$.
Under $\cE_{2,t}$, we obtain
\begin{align*}
	\frac{\hat{\mu}_{B_s}(s) - \hat{\mu}_{C_s}(s)}{\sqrt{1/N_{B_s}(s)+ 1/N_{C_s}(s)}} &\ge\frac{\mu_{i^\star} - \mu_{C_s}}{\sqrt{1/N_{i^\star}(s)+ 1/N_{C_s}(s)}} - \sqrt{2(2+\gamma_{2})  \log t} \\
	&\ge\frac{\mu_{i^\star} - \mu_{C_s}}{\sqrt{1/N_{i^\star}^{i^\star}(s)+ 1/N_{C_s}^{i^\star}(s)}} - \sqrt{2(2+\gamma_{2})  \log t} \\
	&\ge  \sqrt{\frac{1/\beta + 1/\omega^{\star}_{\beta,C_s}}{1/N_{i^\star}^{i^\star}(s)+ 1/N_{C_s}^{i^\star}(s)}}  \min_{i \in \cN(i^\star)} \frac{\mu_{i^\star} - \mu_{i}}{\sqrt{1/\beta + 1/\omega^{\star}_{\beta,i}}} - \sqrt{2(2+\gamma_{2})  \log t}  \\
	&= \sqrt{\frac{1/\beta + 1/\omega^{\star}_{\beta,C_s}}{1/N_{i^\star}^{i^\star}(s)+ 1/N_{C_s}^{i^\star}(s)}}  \sqrt{2T^{\star}_{\beta}(\bm{\mu})^{-1}} - \sqrt{2(2+\gamma_{2})  \log t}  \: ,
\end{align*}
where the second inequality is obtained by artificially making appear $1/\beta + 1/\omega^{\star}_{\beta,C_s}$ and taking the minimum of $i \in \cN(i^\star)$.
The equality simply uses the definition of $\bm \omega^{\star}_{\beta}$ and $T^{\star}_{\beta}(\bm{\mu})$ in~\eqref{eq:beta_characteristic_time}.

\paragraph{Outline} 
We conclude the proof of the non-asymptotic upper bound when $|\cN(i^\star)|=1$ in Appendix~\ref{app:sssec_2arms}, which recovers asymptotic $\beta$-optimality.
When $|\cN(i^\star)|=2$, we provide two non-asymptotic upper bounds.
First, a clipped analysis is leveraged in Appendix~\ref{app:sssec_3arms_clipped} to obtain better non-asymptotic terms at the cost of a factor $2$ in the asymptotic term.
Second, we derive a tighter analysis in Appendix~\ref{app:sssec_3arms_tight} which recovers $\beta$-optimality at the cost of a worse non-asymptotic term.

\subsubsection{Tight Analysis When $\cN(i^\star)=\{j^{\star}\}$}
\label{app:sssec_2arms}

We know that $C_s = j^\star$, $\omega^{\star}_{\beta,C_s} = 1-\beta$ and $N^{i^\star}_{i^\star}(s) + N^{i^\star}_{j^\star}(s) = T_{s}(i^\star,j^\star) = \sum_{s \in [t-1]}\indi{B_s = i^\star}$.
Using Lemmas~\ref{lem:tracking_guaranties} and~\ref{lem:ucb_leader_lower_bound_counts}, upper bounding the second order terms in $s$ by their equivalent in $t$ yields
\begin{align*}
	\min\left\{ \frac{1}{\beta}(N^{i^\star}_{i^\star}(s) + 1), \frac{1}{1-\beta}(N^{i^\star}_{j^\star}(s) + 1/2) \right\} \ge  T_{s}(i^\star,j^\star)  &\ge s - 1 -\sum_{s \in [t^{1/\gamma_{1}}, t]} \indi{B_s \ne i^\star} - t^{1/\gamma_{1}} \\
	&\ge s - 1 -\frac{4 f_{u}(t)}{\beta} H_{1}(\boldsymbol{\mu}) - \frac{2(K-1)}{\beta}  - t^{1/\gamma_{1}} \: .
\end{align*}
Let $n_t = \sup\{s \in [t^{1/\gamma_{1}}, t] \mid B_s = i^\star\}$. 
Then, we know that $n_t \ge t - \frac{4 f_{u}(t)}{\beta} H_{1}(\boldsymbol{\mu}) - \frac{2(K-1)}{\beta}$.
Therefore, we have 
\[
	\max\left\{(t-1)\beta/N_{i^\star}^{i^\star}(n_t), (t-1)(1-\beta)/N_{j^\star}^{i^\star}(n_t) \right\} \le f_{2}(t) =  \left( 1 - \frac{f_{1}(t) +1/\min\{\beta, 1-\beta\}}{t - 1} \right)^{-1} \: ,
\]
where $f_{1}(t) =  t^{1/\gamma_{1}} +\frac{8 f_{u}(t)}{\beta} H_{1}(\boldsymbol{\mu}) + \frac{4(K-1)}{\beta} $.
Plugging in this inequality in the ones derived above and re-ordering the different terms in the inequalities, we obtain 
\begin{align*}
	t - 1 &\le T^{\star}_{\beta}(\bm{\mu}) \left( \sqrt{c(t-1,\delta)} + \sqrt{(2+\gamma_{2})  \log t} \right)^2  \frac{(t-1)/N_{i^\star}^{i^\star}(n_t)+ (t-1)/N_{j^\star}^{i^\star}(n_t)}{1/\beta + 1/(1-\beta)} \\
	&\le T^{\star}_{\beta}(\bm{\mu}) \left( \sqrt{c(t-1,\delta)} + \sqrt{(2+\gamma_{2})  \log t} \right)^2  f_{2}(t) \: ,
\end{align*}
where we removed the $(\cdot)_{+}$ since this will be trivially true for $t > T_{\boldsymbol{\mu}}(\delta)$ where $T_{\boldsymbol{\mu}}(\delta)$ is defined below.
Using that $(t-1)/f_{2}(t) = t - 1 -f_{1}(t) -1/\min\{\beta, 1-\beta\}$, we define
\begin{align*}
	T_{\boldsymbol{\mu}}(\delta) = \sup &\left\{ t\mid t  \le T^{\star}_{\beta}(\bm{\mu}) \left( \sqrt{c(t-1,\delta)} + \sqrt{(2+\gamma_{2})  \log t} \right)^2  + t^{1/\gamma_{1}} + \frac{16\gamma_{1}(1+\gamma_{2})}{\beta} H_{1}(\boldsymbol{\mu}) + \frac{2(2K+1)}{\min\{\beta, 1-\beta\}}\right\} \: .
\end{align*}
Therefore, we have $\cE_{t} \cap \{t < \tau_{\delta}\} = \emptyset$ (i.e. $\cE_{t} \subset \{\tau_{\delta} \le t\}$) for all $t \ge T_{\boldsymbol{\mu}}(\delta) + 1$.
Using Lemma~\ref{lem:lemma_1_Degenne19BAI}, we have shown that $\bE_{\bm{\mu}}[\tau_{\delta}] \le T_{\boldsymbol{\mu}}(\delta) + 1 + (K+3)\zeta(s)$.

\paragraph{Asymptotic $\beta$-Optimality}
It is direct to show that $\limsup_{\delta \to 0} \frac{\bE_{\bm{\mu}}[\tau_{\delta}]}{\log(1/\delta)} \le  \limsup_{\delta \to 0} \frac{T_{\boldsymbol{\mu}}(\delta)}{\log(1/\delta)} \le T^{\star}_{\beta}(\bm{\mu})$, hence UniTT is asymptotically $\beta$-optimality.
For $\beta=1/2$, this shows asymptotic optimality since $T^{\star}_{1/2}(\bm{\mu}) = T^{\star}(\bm{\mu})$ when $|\cN(i^\star)| = 1$.

\paragraph{Explicit Upper Bound}
For increased clarity, we provide a more explicit upper bound by splitting the time.
Let $\epsilon \in (0,1)$.
Then, it is direct to see that $T_{\boldsymbol{\mu}}(\delta) \le \max\{T_{0}(\delta), C_{\boldsymbol{\mu}}, C_{0}\}$ with
\begin{align*}
	T_{0}(\delta) &= \sup \left\{ t > K \mid t \le (1+\epsilon)T^{\star}_{\beta}(\bm{\mu}) \left( \sqrt{c(t-1,\delta)} + \sqrt{(2+\gamma_{2})  \log t} \right)^2 \right\} \: , \\
	C_{\boldsymbol{\mu}} &= \sup \left\{ t \mid t  \le \frac{(1+\epsilon)}{\epsilon} \frac{32\gamma_{1}(1+\gamma_{2}) }{\beta} H_{1}(\boldsymbol{\mu}) \log t + \frac{4(1+\epsilon)(2K+1)}{\epsilon\min\{\beta, 1-\beta\}}\right\}\\
	&\le h_{1}\left(\frac{64\gamma_{1}(1+\gamma_{2}) }{\beta\epsilon} H_{1}(\boldsymbol{\mu}), \frac{8(2K+1)}{\epsilon\min\{\beta, 1-\beta\}} \right) \: , \\
	C_{0} &= \sup \left\{ t \mid t^{(\gamma_{1}-1)/\gamma_{1}}  \le  \frac{2(1+\epsilon)}{\epsilon} \right\} \le \left(\frac{4}{\epsilon} \right)^{\gamma_{1}/(\gamma_{1}-1)}  \: ,
\end{align*}
where we used Lemma~\ref{lem:inversion_upper_bound} which defines  $h_{1}(z,y) = z \overline{W}_{-1}(y/z + \log z) \approx_{z \to + \infty} z \log z$.

\subsubsection{Clipped Analysis When $|\cN(i^\star)| = 2$}
\label{app:sssec_3arms_clipped}

Let $\omega_{0} \in (0,1/2]$ be an allocation threshold.
Since $\sum_{i\in \cN(i^\star)} \omega^{\star}_{\beta,i} = 1-\beta$, there exists at most one arm $i \in \cN(i^\star)$ such that $\omega^{\star}_{\beta,i}<(1-\beta)\omega_{0}$. Otherwise, we would have a contradiction as it would imply that $1-\beta <2\omega_0(1-\beta )$.

We know that $\sum_{i \in \cN(i^\star)\cup\{i^\star\}} N^{i^\star}_{i}(t) = \sum_{i \in \cN(i^\star)} T_{t}(i^\star,i) $.
By the pigeonhole principle, at time $t$, there is an index $i_{1} \in \cN(i^\star)$ such that
\[
	N^{i^\star}_{i_1}(t) \ge \frac{\omega^{\star}_{\beta, i_{1}}}{1-\beta} \left( \sum_{i \in \cN(i^\star)} T_{t}(i^\star,i)  - N^{i^\star}_{i^\star}(t)\right) \: .
\]
Let $i_1$ be such an arm and denote by $i_2$ the unique arm in $\cN(i^\star) \setminus\{i_1\}$. 
We distinguish between three cases.
\begin{itemize}
	\item \textbf{Case 1:} $\omega^{\star}_{\beta, i_{1}} \ge (1-\beta)\omega_{0}$.
	\item \textbf{Case 2:} $\omega^{\star}_{\beta, i_{1}} < (1-\beta)\omega_{0}$ and $N^{i^\star}_{i_1}(t) \ge \omega_0 \left( \sum_{i \in \cN(i^\star)} T_{t}(i^\star,i)  - N^{i^\star}_{i^\star}(t)\right)$.
	\item \textbf{Case 3:} $\omega^{\star}_{\beta, i_{1}} < (1-\beta)\omega_{0}$ and $N^{i^\star}_{i_1}(t) < \omega_0 \left( \sum_{i \in \cN(i^\star)} T_{t}(i^\star,i)  - N^{i^\star}_{i^\star}(t)\right)$.
	Therefore, we have 
	\[
	\omega^{\star}_{\beta, i_{2}} > (1-\beta)(1-\omega_{0}) \quad \text{and} \quad N^{i^\star}_{i_2}(t) \ge (1-\omega_0)\left(\sum_{i \in \cN(i^\star)} T_{t}(i^\star,i)  - N^{i^\star}_{i^\star}(t) \right) \: .
	\]
\end{itemize}
Let us define $\omega_{-} = \max\{(1-\beta)\omega_0, \min_{i \in \cN(i^\star)} \omega^{\star}_{\beta,i}\}$.
In case 1, we define $\omega_{1} = \omega^{\star}_{\beta, i_{1}}$ and we have either $\omega_{-} = (1-\beta)\omega_0 \le \omega_{1}$ or $\omega_{-} = \min_{i \in \cN(i^\star)} \omega^{\star}_{\beta,i} \le \omega_{1}$.
In case 2, we define $\omega_{2} = (1-\beta)\omega_{0}$ and we have $\omega_{-} = (1-\beta)\omega_0 \le \omega_{2}$.
In case 3, we define $\omega_{3} = (1-\beta)(1-\omega_0)$ and we have $\omega_{-} = (1-\beta)\omega_0 \le \omega_{3}$.
Let $k_0$ denote the case in which we are in.
Combining the three above cases, we have shown that
\begin{align*}
	\exists i_0 \in \cN(i^\star), \quad \omega_{k_0} \ge \omega^{\star}_{\beta,i_0} \quad \text{and} \quad N^{i^\star}_{i_0}(t) \ge \frac{\omega_{k_0}}{1-\beta} \left(\sum_{i \in \cN(i^\star)} T_{t}(i^\star,i)  - N^{i^\star}_{i^\star}(t) \right) \: .
\end{align*}
We take such an arm $i_0$.
Using Lemmas~\ref{lem:tracking_guaranties} and~\ref{lem:ucb_leader_lower_bound_counts}, we have 
\[
	N^{i^\star}_{i_0}(t) \ge  \omega_{k_0} \sum_{i \in \cN(i^\star)} T_{t}(i^\star,i) - \frac{\omega_{k_0}}{1-\beta} \ge \omega_{k_0}(t-1) - \omega_{k_0}\left( t^{1/\gamma_{1}} + \frac{4 f_{u}(t)}{\beta} H_{1}(\boldsymbol{\mu})+ \frac{2(K-1)}{\beta} + \frac{1}{1-\beta} \right) \: .
\]
Let $n_t = \sup\{n \le t \mid (B_n,C_n) = (i^\star,i_0)\}$. 
We distinguish between two cases in the following.

\noindent\textbf{Case A:} $n_t \ge t^{1/\gamma_{1}}$.
Then, we have $N^{i^\star}_{i_0}(n_t) \ge N^{i^\star}_{i_0}(t) - 1$.
Using that $\omega_{k_0} \ge \omega_{-}$, hence
\[
	\frac{(t-1)\omega_{k_0}}{N^{i^\star}_{i_0}(n_t)} \le \frac{(t-1)\omega_{k_0}}{N^{i^\star}_{i_0}(n_t)-1} \le \left( 1 - \frac{f_{1}(t)}{t - 1} \right)^{-1} \: \text{with} \: f_{1}(t) =  t^{1/\gamma_{1}} + \frac{4 f_{u}(t)}{\beta} H_{1}(\boldsymbol{\mu})+ \frac{2(K-1)}{\beta} + \frac{1}{1-\beta} + \frac{2}{\omega_{-}} \: .
\]
Using that $N^{i^\star}_{i^\star}(n_t) \ge T_{n_t}(i^\star,i_0) - N^{i^\star}_{i_0}(n_t)$ and $1/\beta \ge 0$, we obtain
\begin{align*}
	1/N^{i^\star}_{i^\star}(n_t) + 1/N^{i^\star}_{i_0}(n_t) \le \frac{1}{t - 1} \left( 1/\beta + \frac{1}{\omega_{k_0}}\frac{(t-1)\omega_{k_0}}{N^{i^\star}_{i_0}(n_t)} \right) \left( \frac{N^{i^\star}_{i_0}(n_t)}{T_{n_t}(i^\star,i_0) - N^{i^\star}_{i_0}(n_t)} + 1\right) \: .
\end{align*}
Since $T_{n_t}(i^\star,i_0) \ge \frac{N^{\star}_{i_0}(n_t)-1}{1-\beta}$, we have
\begin{align*}
	\left(\frac{N^{i^\star}_{i_0}(n_t)}{T_{n_t}(i^\star,i_0) - N^{i^\star}_{i_0}(n_t)} + 1 \right)^{-1}\ge \left(\frac{1-\beta}{\beta -1/N^{i^\star}_{i_0}(n_t) } + 1 \right)^{-1}= \beta - \frac{1-\beta}{N^{i^\star}_{i_0}(n_t)-1} \ge \beta - \frac{1-\beta}{(t - 1 - f_{1}(t))\omega_{k_0}} \: .
\end{align*}
Using that $\left( 1 - \frac{f_{1}(t)}{t - 1} \right)^{-1} \ge 1$ and the above results, we obtain
\begin{align*}
	\left( 1/N^{i^\star}_{i^\star}(n_t) + 1/N^{i^\star}_{i_0}(n_t) \right)^{-1/2} \ge \frac{\sqrt{t - 1 - f_{1}(t)}}{\sqrt{1/\beta + 1/\omega_{k_0}}} \sqrt{\beta - \frac{1-\beta}{(t - 1 - f_{1}(t))\omega_{-}} } \: .
\end{align*}
Putting everything together and using that $\omega^{\star}_{\beta,i_0} \le \omega_{k_{0}}$, we have
\begin{align*}
	t - 1 \le \frac{T^{\star}_{\beta}(\bm{\mu})}{\beta} \left(\sqrt{c(t-1,\delta)} + \sqrt{(2+\gamma_{2})  \log t} \right)^2 + f_{1}(t) + \frac{1-\beta}{\beta \omega_{-}} \: .
\end{align*}

\noindent\textbf{Case B:} $n_t < t^{1/\gamma_{1}}$.
Let $j_0 \in \cN(i^\star) \setminus \{i_0\}$.
Then, we have $C_s = j_0$ for all $s \in [t^{1/\gamma_{1}},t]$ such that $B_s = i^\star$, i.e. $ \sum_{s \in [t^{1/\gamma_{1}},t]} \indi{B_s = i^\star} =  \sum_{s \in [t^{1/\gamma_{1}},t]} \indi{B_s = i^\star, C_s = j_0}$.
Since $N^{i^\star}_{i^\star}(s) \ge T_{s}(i^\star,j_0)  - N^{i^\star}_{j_0}(s)$, using Lemmas~\ref{lem:tracking_guaranties} and~\ref{lem:ucb_leader_lower_bound_counts} yields
\begin{align*}
	\min\left\{ \frac{1}{\beta}(N^{i^\star}_{i^\star}(s) + 1), \frac{1}{1-\beta}(N^{i^\star}_{j_0}(s) + 1/2) \right\} \ge  T_{s}(i^\star,j_0) &\ge s -1 -\sum_{s \in [t^{1/\gamma_{1}},t]} \indi{B_s \ne i^\star} - t^{1/\gamma_{1}} \\
	& \ge s-1 - \left( t^{1/\gamma_{1}} + \frac{4 f_{u}(t)}{\beta} H_{1}(\boldsymbol{\mu})+ \frac{2(K-1)}{\beta} \right)  \: .
\end{align*}
Let $n_t = \sup\{s \in [t^{1/\gamma_{1}}, t] \mid B_s = i^\star\}$. 
Then, we know that $C_{n_t} = j_0$ and $n_t \ge t - \frac{4 f_{u}(t)}{\beta} H_{1}(\boldsymbol{\mu}) - \frac{2(K-1)}{\beta}$.
Let $f_{2}(t) =  t^{1/\gamma_{1}} +\frac{8 f_{u}(t)}{\beta} H_{1}(\boldsymbol{\mu}) + \frac{4(K-1)}{\beta} $. Therefore, we have shown that
\[
	\max\left\{(t-1)\beta/N_{i^\star}^{i^\star}(n_t), (t-1)(1-\beta)/N_{j_0}^{i^\star}(n_t) \right\} \le \left( 1 - \frac{f_{2}(t) +1/\min\{\beta, 1-\beta\}}{t - 1} \right)^{-1} \: .
\]
Using that $\omega^{\star}_{\beta,j_0} \le 1 - \beta$, we conclude as in Appendix~\ref{app:sssec_2arms} that
\begin{align*}
	t - 1 &\le T^{\star}_{\beta}(\bm{\mu}) \left( \sqrt{c(t-1,\delta)} + \sqrt{(2+\gamma_{2})  \log t} \right)^2 + f_{2}(t) + 1/\min\{\beta, 1-\beta\} \: .
\end{align*}

\noindent\textbf{Summary.}
Let us define
\begin{align*}
	\widetilde T_{\boldsymbol{\mu}}(\delta,\omega_0) &= \sup \left\{ t > K \mid t \le \frac{T^{\star}_{\beta}(\bm{\mu})}{\beta} \left( \sqrt{c(t-1,\delta)} + \sqrt{(2+\gamma_{2})  \log t} \right)^2 + t^{1/\gamma_{1}} \right. \\
	&\qquad \left. + \frac{8 f_{u}(t)}{\beta} H_{1}(\boldsymbol{\mu}) + \frac{2(2K-1)}{\min\{\beta, 1-\beta\}}  + \frac{1+\beta}{\beta \max\{(1-\beta)\omega_0, \min_{i \in \cN(i^\star)} \omega^{\star}_{\beta,i}\}} \right\} \: .
\end{align*}
Combining both cases, for all $\omega_0 \in (0,1/2]$, we have $\cE_{t} \cap \{t < \tau_{\delta}\} = \emptyset$ (i.e. $\cE_{t} \subset \{\tau_{\delta} \le t\}$) for all $t \ge \widetilde T_{\boldsymbol{\mu}}(\delta,\omega_0) + 1$.
The infimum of $ \widetilde T_{\boldsymbol{\mu}}(\delta,\omega_0)$ is achieved at $\omega_{0}=1/2$ since there is no trade-off.
Using Lemma~\ref{lem:lemma_1_Degenne19BAI}, we have shown that $\bE_{\bm{\mu}}[\tau_{\delta}] \le \widetilde T_{\boldsymbol{\mu}}(\delta,1/2) + 1 + (K+3)\zeta(s)$.

\paragraph{Explicit Upper Bound}
For increased clarity, we provide a more explicit upper bound by splitting the time.
Let $\epsilon \in (0,1)$.
Then, it is direct to see that $\widetilde T_{\boldsymbol{\mu}}(\delta,1/4) \le \max\{\widetilde T_{0}(\delta), \widetilde C_{\boldsymbol{\mu}}, C_{0}\}$ with
\begin{align*}
	\widetilde T_{0}(\delta) &= \sup \left\{ t > K \mid t  \le \frac{(1+\epsilon)T^{\star}_{\beta}(\bm{\mu})}{\beta} \left( \sqrt{c(t-1,\delta)} + \sqrt{(2+\gamma_{2})  \log t} \right)^2 \right\} \: ,\\
	\widetilde C_{\boldsymbol{\mu}} &= \sup \left\{ t \mid t  \le \frac{(1+\epsilon)}{\epsilon} \frac{32\gamma_{1}(1+\gamma_{2}) }{\beta} H_{1}(\boldsymbol{\mu}) \log t + \frac{2(1+\epsilon)}{\epsilon}  \left(\frac{2(2K-1)}{\min\{\beta, 1-\beta\}}  + \frac{2(1+\beta)}{\beta (1-\beta)}  \right)\right\}\\
	&\le h_{1}\left(\frac{64\gamma_{1}(1+\gamma_{2}) }{\beta\epsilon} H_{1}(\boldsymbol{\mu}),  \frac{4}{\epsilon}  \left(\frac{2(2K-1)}{\min\{\beta, 1-\beta\}}  +  \frac{2(1+\beta)}{\beta (1-\beta)}  \right) \right)  \: , \\
	C_{0} &= \sup \left\{ t \mid t^{(\gamma_{1}-1)/\gamma_{1}}  \le  \frac{2(1+\epsilon)}{\epsilon} \right\} \le \left(\frac{4}{\epsilon} \right)^{\gamma_{1}/(\gamma_{1}-1)} \: ,
\end{align*}
where we used Lemma~\ref{lem:inversion_upper_bound} which defines  $h_{1}(z,y) = z \overline{W}_{-1}(y/z + \log z) \approx_{z \to + \infty} z \log z$.

\subsubsection{Tight Analysis When $|\cN(i^\star)|=2$}
\label{app:sssec_3arms_tight}

Let $L_{\boldsymbol{\mu}}$ and $D_{\boldsymbol{\mu}}$ as in Lemmas~\ref{lem:time_leader_correct} and~\ref{lem:time_challenger_not_oversampled}.
Let $t \ge D_{\boldsymbol{\mu}} \ge L_{\boldsymbol{\mu}}$ such that $\cE_{t} \cap \{t < \tau_{\delta}\}$ holds.
Using Lemmas~\ref{lem:time_leader_correct} and~\ref{lem:time_neighbors_controlled}, we have $B_{t} = i^\star$, $C_{t} = i \in \cN(i^\star)$, and
\begin{align*}
	\max\left\{\frac{\beta(t-1)}{N^{i^\star}_{i^\star}(t)} , \frac{(\omega^{\star}_{\beta,i} - \gamma)(t-1)}{N^{i^\star}_{i}(t)} \right\}&\le \left( 1 - \frac{L_{\boldsymbol{\mu}}+2D_{\boldsymbol{\mu}}/\min\{\beta, \min_{i \in \cN(i^\star)}\omega^{\star}_{\beta,i} - \gamma\}}{t - 1}\right)^{-1}\: .
\end{align*}
Using the stopping rule~\eqref{eq:GLR_stopping_rule} and similar arguments as above, we obtain
\begin{align*}
	&\sqrt{2c(t-1,\delta)} \ge \frac{(\hat{\mu}_{i^\star}(t) - \hat{\mu}_{C_t}(t))_{+}}{\sqrt{1/N_{i^\star}(t)+ 1/N_{C_{t}}(t)}} \ge \frac{\mu_{i^\star} - \mu_{C_{t}}}{\sqrt{1/N_{i^\star}^{i^\star}(t)+ 1/N_{C_t}^{i^\star}(t)}} -\sqrt{2(2+\gamma_{2})\log t} \\
	&\ge \sqrt{\frac{1/\beta + 1/\omega^{\star}_{\beta,C_{t}}}{1/\beta+ 1/(\omega^{\star}_{\beta,C_{t}} - \gamma)}}  \sqrt{2T^{\star}_{\beta}(\bm{\mu})^{-1}} \sqrt{t - 1 - \frac{3D_{\boldsymbol{\mu}}}{\min\{\beta, \min_{i \in \cN(i^\star)}\omega^{\star}_{\beta,i} - \gamma\}}}  -\sqrt{2(2+\gamma_{2})\log t}
\end{align*}
Let $j \in \cN(i^\star)$.
Using the same notation as in the proof of Lemma~\ref{lem:time_challenger_not_oversampled}, one can show that
\begin{align*}
	\frac{g(\gamma)^2}{g(0)^2} - 1 &= - \frac{\gamma\beta}{\omega^{\star}_{\beta,j}( \beta + \omega^{\star}_{\beta,j} - \gamma)} \ge - \frac{\gamma}{\min_{i \in \cN(i^\star)}\omega^{\star}_{\beta,i}} \: .
\end{align*}
Using this result and re-ordering the terms, we obtain
\begin{align*}
t - 1 \le \frac{T^{\star}_{\beta}(\bm{\mu})}{1-\frac{\gamma}{\min_{i \in \cN(i^\star)}\omega^{\star}_{\beta,i}}} \left(\sqrt{c(t-1,\delta)} + \sqrt{(2+\gamma_{2})\log t} \right)^2 + \frac{3D_{\boldsymbol{\mu}}}{\min\{\beta, \min_{i \in \cN(i^\star)}\omega^{\star}_{\beta,i} - \gamma\}} \: .
\end{align*}
Let us define
\begin{align*}
	T_{\boldsymbol{\mu}}(\delta) = \sup &\left\{ t > K \mid t - 1 \le \frac{T^{\star}_{\beta}(\bm{\mu})\left(\sqrt{c(t-1,\delta)} + \sqrt{(2+\gamma_{2})\log t} \right)^2 }{1-\gamma/\min_{i \in \cN(i^\star)}\omega^{\star}_{\beta,i}} + \frac{3D_{\boldsymbol{\mu}}}{\min\{\beta, \min_{i \in \cN(i^\star)}\omega^{\star}_{\beta,i} - \gamma\}}\right\} \: .
\end{align*}
Therefore, we have $\cE_{t} \cap \{t < \tau_{\delta}\} = \emptyset$ (i.e. $\cE_{t} \subset \{\tau_{\delta} \le t\}$) for all $t \ge T_{\boldsymbol{\mu}}(\delta) + 1$.
Using Lemma~\ref{lem:lemma_1_Degenne19BAI}, we have shown that $\bE_{\bm{\mu}}[\tau_{\delta}] \le T_{\boldsymbol{\mu}}(\delta) + 1 + (K+3)\zeta(s)$.

\paragraph{Asymptotic $\beta$-Optimality}
It is direct to show that $\limsup_{\delta \to 0} \frac{\bE_{\bm{\mu}}[\tau_{\delta}]}{\log(1/\delta)} \le  \limsup_{\delta \to 0} \frac{T_{\boldsymbol{\mu}}(\delta)}{\log(1/\delta)} \le \frac{T^{\star}_{\beta}(\bm{\mu})}{1-\gamma/\min_{i \in \cN(i^\star)}\omega^{\star}_{\beta,i}}$.
Taking $\gamma \to 0$, we obtain asymptotic $\beta$-optimality.

\paragraph{Explicit Upper Bound}
For increased clarity, we provide a more explicit upper bound by splitting the time.
Let $\epsilon \in (0,1)$.
Then, it is direct to see that $T_{\boldsymbol{\mu}}(\delta) \le \max\{T_{0}(\delta), C_{\boldsymbol{\mu}}\}$ with
\begin{align*}
	T_{0}(\delta) &= \sup \left\{ t > K \mid t \le \frac{(1+\epsilon)T^{\star}_{\beta}(\bm{\mu})}{1-\gamma/\min_{i \in \cN(i^\star)}\omega^{\star}_{\beta,i}} \left( \sqrt{c(t-1,\delta)} + \sqrt{(2+\gamma_{2})  \log t} \right)^2 \right\} \\
	\text{where} &\quad \limsup_{\delta \to 0}  \frac{T_{0}(\delta)}{\log(1/\delta)} \le \frac{(1+\epsilon)T^{\star}_{\beta}(\bm{\mu})}{1-\gamma/\min_{i \in \cN(i^\star)}\omega^{\star}_{\beta,i}}  \: , \\
	C_{\boldsymbol{\mu}} &=  \frac{6D_{\boldsymbol{\mu}}}{\epsilon\min\{\beta, \min_{i \in \cN(i^\star)}\omega^{\star}_{\beta,i} - \gamma\}} + \frac{2}{\epsilon} \: .
\end{align*}

\subsection{Efficient Implementation}\label{subsec-app:unitt-eff-impl}
We now discuss in detail how to efficiently implement UniTT. In this sense, we recall that the most crucial step is the computation of the leader $B_t$, that is:
\begin{equation*}
B_{t} \in \argmax_{i \in [K]} \max_{\boldsymbol{\lambda} \in \widetilde{\Theta}_t}  \lambda_i  \: ,
\end{equation*}
Similar to what we have presented for O-TaS, we first show how to compute $B_t$ with a $\mathcal{O}(K^2)$ operations (i.e., Algorithm \ref{alg:uni-tt-squared}), and we then discuss how to leverage dynamic programming in order to obtain a fast $\mathcal{O}(K)$ procedure. 

\begin{algorithm}[t]
	\caption{A $\mathcal{O}(K^2)$ procedure for UniTT.}
    \label{alg:uni-tt-squared}
	\begin{algorithmic}
		\REQUIRE{$\Theta_t = \bigotimes_{i \in [K]} [\alpha_i(t), \beta_i(t)]$}
		\STATE{ {\color{blue} // Step 1: Compute $\bm{\mu}^{(i)}(t)$ for all $i \in [K]$ }}		
		\FOR{$i \in [K]$}
	 	\STATE{Initialize $\bm{\mu}^{(i)}(t) = \{ 0 \}_{i=1}^K$ }
		\STATE{Set ${\mu}^{(i)}_i(t) = \beta_i(t)$, $\mu^{(i)}_1(t) = \gamma_{1}(t)$, and $\mu^{(i)}_K(t) = \alpha_K(t)$ }	 	
	 	\FOR{$j \in \{1, \dots, i-1 \}$}
	 	\STATE{Set $\mu^{(i)}_j(t) = \max \{ \mu^{(i)}_{j-1}(t), \alpha_j(t) \}$}
	 	\ENDFOR
	 	\FOR{$j \in \{K-1, \dots, i + 1 \}$}
	 	\STATE{Set $\mu^{(i)}_j(t) = \max \{ \mu^{(i)}_{j+1}(t), \alpha_j(t) \}$}
	 	\ENDFOR
	 	\ENDFOR
	 	\STATE{ {\color{blue} // Step 2: Compute the set $\mathcal{Z}$ of valid unimodal bandits }}
	 	\STATE{ $\mathcal{Z} \coloneqq \{ i \in [K]: {\mu}^{(i)}_j(t) \ge {\mu}^{(i)}_{j-1}(t) ~\forall j \le i \textup{ and } {\mu}^{(i)}_j(t) \ge {\mu}^{(i)}_{j+1}(t) ~ \forall j \ge i \} $}
	 	\STATE{ {\color{blue} // Step 3: Return the optimistic arm within $\mathcal{Z}$}}
	 	\RETURN{ $i \in \argmax_{j \in \mathcal{Z}} \mu^{(j)}_j$}
	\end{algorithmic}
\end{algorithm}

As one can note, Algorithm \ref{alg:uni-tt-squared} is almost identical to Algorithm \ref{alg:O-TaS-k-squared}; the only difference is in the return step, as UniTT needs to compute the optimistic leader. Thus, for the same reason that we discussed in Section \ref{subsec-app:O-TaS-impl}, the computational complexity of Algorithm \ref{alg:uni-tt-squared} is $\mathcal{O}(K^2)$. Furthermore, proving that the algorithm is correct, namely:
\begin{align*}
\argmax_{j \in \mathcal{Z}} \mu^{(j)}_j = \argmax_{i \in [K]} \max_{\boldsymbol{\lambda} \in \widetilde{\Theta}_t}  \lambda_i,
\end{align*}
is direct from the proof that we presented for Algorithm \ref{alg:O-TaS-k-squared}. Indeed, Lemma \ref{lemma:z-empty-set} holds unchanged, and the equivalent of Lemma \ref{lemma:correct-eff} is even simpler, as one can verify from the following lemma.
\begin{lemma}\label{lemma:correct-eff-unitt}
It holds that:
\begin{align*}
\argmax_{j \in \mathcal{Z}} \mu^{(j)}_j = \argmax_{i \in [K]} \max_{\boldsymbol{\lambda} \in \widetilde{\Theta}_t}  \lambda_i.
\end{align*}
\end{lemma}
\begin{proof}
If $\widetilde{\Theta}_t = \emptyset$, then, the claim is direct from Lemma \ref{lemma:z-empty-set}. We now consider the case in which $\widetilde{\Theta}_t \ne \emptyset$. Let Let us denote by $\widetilde{\Theta}_{t,i}$ the subset of $\widetilde{\Theta}_t$ where $i$ is the optimal arm, that is $\widetilde{\Theta}_{t,i} \coloneqq \left\{ \bm{\mu} \in \widetilde{\Theta}_t: i^{\star}(\bm{\mu}) = i \right\}.$. Then, the result follows by noticing that, for all $i \in \mathcal{Z}$, and all $\bm{\mu} \in \widetilde{\Theta}_{t,i}$ it holds that $\mu_{i}^{(i)} \ge \mu_i$. Indeed, by definition, $\mu_i^{(i)} = \beta_i$. Moreover, for all $i \notin \mathcal{Z}$, $\widetilde{\Theta}_{t,i} = \emptyset$ (i.e., Lemma \ref{lemma:z-empty-set}).
\end{proof}

At this point, one can apply the same identical tricks presented for O-TaS to reduce the computational complexity of Algorithm \ref{alg:uni-tt-squared} from $\mathcal{O}(K^2)$ to $\mathcal{O}(K)$. We refer the reader to Section \ref{subsec-app:O-TaS-impl} (in particular, Lemma \ref{lemma:dynamic-programming} and Lemma \ref{lemma:dyn-prog-2}).

% !TeX root = ../paper.tex

\section{TECHNICALITIES}
\label{app:technicalities}

Appendix~\ref{app:technicalities} gathers existing and new technical results which are used for our proofs.

\paragraph{Methodology}
Lemma~\ref{lem:lemma_1_Degenne19BAI} is a standard result to upper bound the expected sample complexity of an algorithm, e.g. see Lemma 1 in \citet{degenne2019non} or~\citet{garivier2016optimal}.
This is a key method extensively used in the literature.
\begin{lemma} \label{lem:lemma_1_Degenne19BAI}
	Let $(\cE_{t})_{t > K}$ be a sequence of events and $T(\delta) > K $ be such that for $t \ge T(\delta)$, $\cE_{t} \subseteq \{\tau_{\delta} \le t\}$. Then, $\bE_{\bm{\mu}}[\tau_{\delta}] \le T(\delta)  + \sum_{t > K} \bP_{\bm{\mu}}(\cE_{t}^{\complement})$.
\end{lemma}
\begin{proof}
	Since the random variable $\tau_{\delta}$ is positive and $\{\tau_{\delta} > t\} \subseteq \cE_{t}^{\complement}$ for all $t \ge T(\delta)$, we have
	\[
	\bE_{\bm{\mu}}[\tau_{\delta}] = \sum_{t \ge 0} \bP_{\bm{\mu}} (\tau_{\delta} > t) \le T(\delta)  + \sum_{t \ge T(\delta)} \bP_{\bm{\mu}}(\cE_{t}^{\complement}) \: .
	\]
\end{proof}

\paragraph{Tracking}
Lemma~\ref{lemma:tracking} gathers propertis of the C-Tracking procedure as in~\ref{eq:C_tracking} used by TaS and O-TaS.
\begin{lemma}\label{lemma:tracking}
Let $t \in \mathbb{N}$ and $i \in [K]$. Then, we have
\begin{align}
-\ln(K) \le N_i(t) - \sum_{s \in [t]} \omega_i(s) \le 1\: .\label{eq:tracking-1}
\end{align}
Furthermore, C-tracking satisfies:
\begin{align}
& \sum_{s \in [t]} \sum_{i \in [K]} \frac{\omega_i(s)}{\sqrt{N_i(s)}} \le K\ln(K) + 2 \sqrt{2Kt} \: , \label{eq:tracking-2} \\
& \sum_{s \in [t]} \sum_{i \in [K]} \frac{\omega_i(s)}{N_i(s)} \le 2K \ln(t) \: . \label{eq:tracking-3}
\end{align}
Finally, consider any subset $\mathcal{A} \subseteq [K]$. Then, it holds that:
\begin{align}\label{eq:tracking-4}
\sum_{s \in [t]} \sum_{i \in \mathcal{A}} \frac{\omega_i(s)}{\sqrt{N_i(s)}} \le |\mathcal{A}| \ln |\mathcal{A}| + 2 \sqrt{2 |\mathcal{A}| t}.
\end{align}
\end{lemma}
\begin{proof}
The proof of Equation~\eqref{eq:tracking-1} is from Theorem 6 in \cite{degenne2020structure}. The proof of Equations \eqref{eq:tracking-2}-\eqref{eq:tracking-3} are from Lemma 6 in \cite{degenne2020structure}. Similarly, the proof of Equation \eqref{eq:tracking-4} is a straightforward generalization of the proof of Equation \eqref{eq:tracking-2}. 
\end{proof}

Lemma~\ref{lem:tracking_guaranties} provide general results for the $2(K-1)$ tracking procedures with fixed design $\beta \in (0,1)$ used by \hyperlink{UniTT}{UniTT}.
\begin{lemma} \label{lem:tracking_guaranties}
	For all $t > K$, $i \in [K]$, $j \in \cN(i)$, we have
	\[
		-1/2 \le N_{j}^{i}(t) - (1 - \beta) T_{t}(i, j)  \le 1 \quad \text{i.e.} \quad - 1 \le T_{t}(i,j) - N^{i}_{j}(t) - \beta T_{t}(i,j) \le 1/2
	\]
	and $(N^{i}_{i}(t) - 1)/ \beta \le \sum_{j \in \cN(i)} T_{t}(i,j) = \sum_{j \ne i} T_{t}(i,j) \le (N^{i}_{i}(t) + 2)/ \beta$.
	Let $T_{t}(i) \eqdef \sum_{j \in \cN(i)}(T_{t}(i,j)+ T_{t}(j,i))$. Then, we have $N_{i}(t) \ge \min\{\beta, 1-\beta\}\left( T_{t}(i)   - 3 \right)$.
\end{lemma}
\begin{proof}
	We have $2(K-1)$ independent two-arms C-Tracking between a leader $i \in [K]$ and a challenger $j \in \cN(i)$.
	Theorem 6 in \cite{degenne2020structure} yields the result.
	The second inequality is a simple re-ordering.

	To obtain the second part, we have by definition that the challenger is a neighbor of the leader, i.e. $\sum_{j \in \cN(i)} T_{t}(i,j) = \sum_{j \ne i} T_{t}(i,j)$.
	Then, direct manipulations yield that
	\begin{align*}
	 \sum_{j \in \cN(i)} T_{t}(i,j) \le \sum_{j \in \cN(i)} \frac{T_{t}(i,j) - N^{i}_{j}(t) + 1}{\beta} \le  \frac{\sum_{j \in \cN(i)}(T_{t}(i,j) - N^{i}_{j}(t)) + 2}{ \beta}  \: ,
	\end{align*}
	which allows to conclude since $N^{i}_{i}(t) = \sum_{j \in \cN(i)}(T_{t}(i,j) - N^{i}_{j}(t))$. The lower bound is obtained similarly.
	Using that $j \in \cN(i)$ if and only if $i \in \cN(j)$, the third is a direct consequence of the first part since
	\begin{align*}
		N_{i}(t) =  \sum_{j \in \cN(i)}(T_{t}(i,j) - N^{i}_{j}(t)) + \sum_{j \in \cN(i)} N^{j}_{i}(t) &\le \beta \sum_{j \in \cN(i)}T_{t}(i,j) + (1-\beta)\sum_{j \in \cN(i)} T_{t}(j,i) + 3 \\
		&\ge \beta\sum_{j \in \cN(i)}T_{t}(i,j) + (1-\beta)\sum_{j \in \cN(i)} T_{t}(j,i) - 3  \: .
	\end{align*}
\end{proof}

\paragraph{Simple but Key Observation}
Lemma~\ref{lem:simple_key_observation} is based on the key but simple observation that the number of times one can increase a bounded positive quantity by one is also upper bounded.
\begin{lemma}[Lemma 4 in~\citet{jourdan2023epsilonbestarm}] \label{lem:simple_key_observation}
	Let $\cA \subseteq [K]$ and $t > K$.
	Let $T_{t}(i) \eqdef \sum_{j \ne i}(T_{t}(i,j) + T_{t}(j,i))$ be the number of times arm $i$ was selected in the leader/challenger pair.
	Assume there exists a sequence of events $(A_{s}(t))_{t \ge s \ge t^{1/\gamma_{1}}}$ and positive reals $(D_{i}(t))_{i \in \cA}$ such that, for all $s \in \{t^{1/\gamma_{1}},\ldots, t\}$, under the event $A_{s}(t)$,
	\begin{equation} \label{eq:bad_event_implies_bounded_quantity_increases}
		\exists I_{s} \in \cA, \quad T_{s}(I_{s}) \le D_{I_{s}}(t) \quad \text{and} \quad T_{s+1}(I_{s}) = T_{s}(I_{s}) + 1 \: .
	\end{equation}
	Then, we have $\sum_{s \in [t^{1/\gamma_{1}},t]} \indi{A_{s}(t)} \le \sum_{i \in \cA} D_{i}(t)$.
\end{lemma}

\paragraph{Inversion Result}
Lemma~\ref{lem:property_w_lambert} gathers properties on the function $\overline{W}_{-1}$, which is used in the literature to obtain concentration results.
\begin{lemma}[Appendix A in~\citet{jourdan2024SolvingPureExploration}] \label{lem:property_w_lambert}
	Let $\overline{W}_{-1}(x)  = - W_{-1}(-e^{-x})$ for all $x \ge 1$, where $W_{-1}$ is the negative branch of the Lambert $W$ function.
	The function $\overline{W}_{-1}$ is increasing on $(1, +\infty)$ and strictly concave on $(1, + \infty)$.
	In particular, $\overline{W}_{-1}'(x) = \left(1-\frac{1}{\overline{W}_{-1}(x)} \right)^{-1}$ for all $x > 1$.
	Then, for all $y \ge 1$ and $x \ge 1$,
	\[
	 	\overline{W}_{-1}(y) \le x \quad \iff \quad y \le x - \ln(x) \: .
	\]
	Moreover, for all $x > 1$,
	\[
	 x + \log(x) \le \overline{W}_{-1}(x) \le x + \log(x) + \min \left\{ \frac{1}{2}, \frac{1}{\sqrt{x}} \right\} \: .
	\]
\end{lemma}

Lemma~\ref{lem:inversion_upper_bound} is an inversion result to upper bound a time which is implicitly defined.
\begin{lemma}[Lemma 51 in~\citet{jourdan2023epsilonbestarm}] \label{lem:inversion_upper_bound}
	Let $\overline{W}_{-1}$ defined in Lemma~\ref{lem:property_w_lambert}.
	Let $A > 0$, $B > 0$ such that $B/A + \log A >  1$ and $C(A, B) = \sup \left\{ x \mid \: x < A \log x + B \right\} $.
	Then, $C(A,B) < h_{1}(A,B)$ with $h_{1}(z,y) \eqdef z \overline{W}_{-1} \left(y/z  + \log z\right)$.
\end{lemma}
% \begin{proof}
% 	Since $B/A + \log A >  1$, we have $C(A, B) \ge A$, hence
% 	\[
% 	C(A, B) = \sup \left\{ x \mid \: x < A \log (x) + B \right\} = \sup \left\{ x \ge A \mid \: x < A \log (x) + B \right\} \: .
% 	\]
% 	Using Lemma~\ref{lem:property_w_lambert} yields that
% 	\begin{align*}
% 		x \ge A \log x + B  \: \iff \: \frac{x}{A} - \log \left( \frac{x}{A} \right) \ge \frac{B}{A} + \log A \: \iff \: x \ge A \overline{W}_{-1} \left( \frac{B}{A} + \log A \right) \: .
% 	\end{align*}
% \end{proof}

% !TeX root = ../paper.tex

\section{EXPERIEMENT DETAILS AND ADDITIONAL RESULTS} 
\label{app:add_expe}

This last section is structured in the following way:
\begin{itemize}
\item First, in Section~\ref{subsec-app:exp-details}, we provide additional details on the experiments (computing infrastructure, process to generate random unimodal bandits, and implementation details).
\item Then, in Section~\ref{subsec-app:k-rnd}, we ablate the performance of UniTT and O-TaS against TTUCB in random unimodal bandits with different values of $K$. As $K$ increases, the performance gap between UniTT/O-TaS and TTUCB increases as well.
\item In Section~\ref{subsec-app:uni-tt-abl}, we present an ablation on UniTT, to understand which components leads to the performance improvement over TTUCB.
\item Finally, in Section~\ref{subsec-app:theta}, we present an ablation whose goal is understanding the behavior of O-TaS and UniTT. Specifically, we focus on presenting some visualization of the structured confidence intervals and how they evolve over time.
\end{itemize}

\subsection{Experiment Details}\label{subsec-app:exp-details}
We now provide additional details on the experiments. 

\paragraph{Computing Infrastructure} All results are obtained using the following computer infrastructure: 100 Intel(R) Xeon(R) Gold 6238R CPU @ 2.20GHz cpus and 256GB of RAM. 

\paragraph{Codebase} The codebase can be found at \url{https://github.com/riccardopoiani/unimodal-pure-exp}.

\paragraph{Randomly Generated Instances} We now detail the process that we used to generate the random unimodal bandits that we used in our experiments. Given the number of arms $K$, we first sample (with a uniform distribution over $\{1, \dots, K\}$) the position of the optimal arm. We then fix the value of mean of the optimal arm to $\mu_{\star} = 1$. Once this is done, we specify three real positive parameters $k_{\textup{min}}, k_{\textup{max}}$ and $k$ that are used as follows. For all arms $i \in \mathcal{N}(\star)$, we sample the value of $\mu_i$ uniformly in the range $[\mu_\star - k_{\textup{max}}, \mu_\star - k_{\textup{min}}]$. Then, for all the remaining arms $j < \star - 1$, we sample $\mu_j$ uniformly in $[\mu_{j+1} - k,\mu_{j+1}]$, and for all $j > \star + 1$, we sample $\mu_j$ uniformly in $[\mu_{j-1} - k,\mu_{j-1}]$. In all cases, we set $k_{\textup{min}} = 0.2$ and $k_{\textup{max}} = 0.3$. These values ensure that (i) there is a minimum positive gap with the optimal arm, making the best arm identifiable, and (ii) the gap is not too large to avoid making the BAI problem too easy. Concerning $k$, instead, we set $k = 0.1$. 
Concerning the scenario with $K=10$, we obtained the following $\bm{\mu}$ vector (values are rounded to three decimal places for the ease of read):
\begin{align*}
\bm{\mu} = (0.568, 0.573, 0.598, 0.645, 0.724, 1.0, 0.770, 0.712, 0.648, 0.622).
\end{align*}
The bandit model that we generated for $K=100$ can be found in the codebase.

\paragraph{Threshold} For all algorithms and in all domains, we calibrated the threshold $c_K(t, \delta)$ as $c_K(t, \delta) = \log\left( \frac{K}{\delta} \right) + \log( \log(t) + 1)$. This value has satisfied the $\delta$-correctness requirement for all algorithms in all experiments.

\subsection{Ablation on $K$ in Random Instances}\label{subsec-app:k-rnd}
In this section, we study the performance of the algorithms in the random instances when varying the number of arms. 
As algorithms, we consider UniTT, O-TaS, and TTUCB \citep{jourdan2023NonAsymptoticAnalysis}. We excluded U-TaS since, as shown in Section \ref{sec:expe}, it underperforms due to the use of forced exploration (when we further increase $K$, the performance of U-TaS on these random instances even amplifies). 

\begin{figure*}[t!]
\centering
\begin{minipage}{.45\textwidth}
  \centering
  \includegraphics[width=\textwidth]{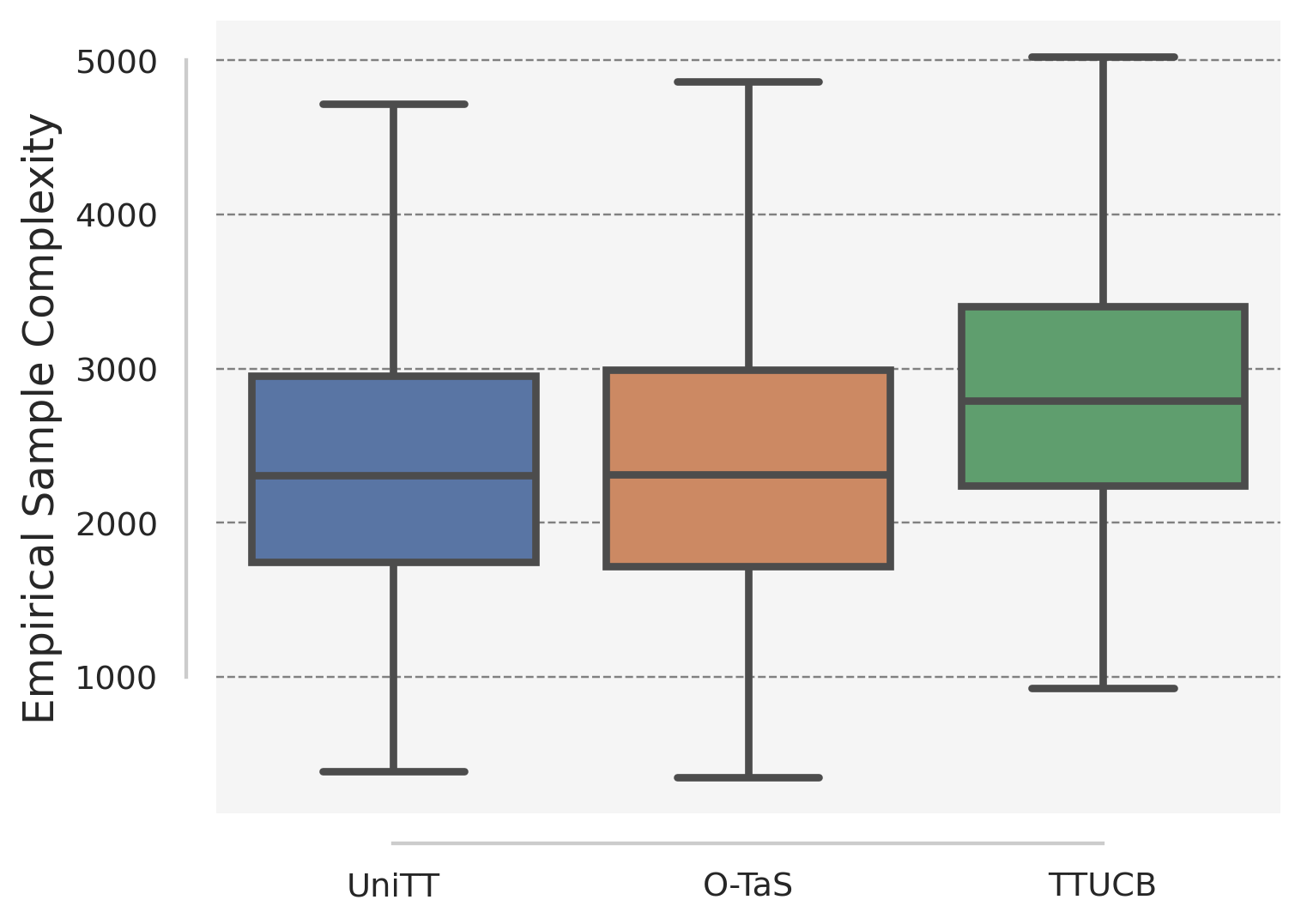}
  \caption{Ablation on $K$ in random instances. Empirical sample complexity on $\bm{\mu}_{\textup{R}, 10}$.}
  \label{fig:k10}
\end{minipage}%
\hspace{.55cm} % Adjust the horizontal space between figures here
\begin{minipage}{.45\textwidth}
  \centering
  \includegraphics[width=\textwidth]{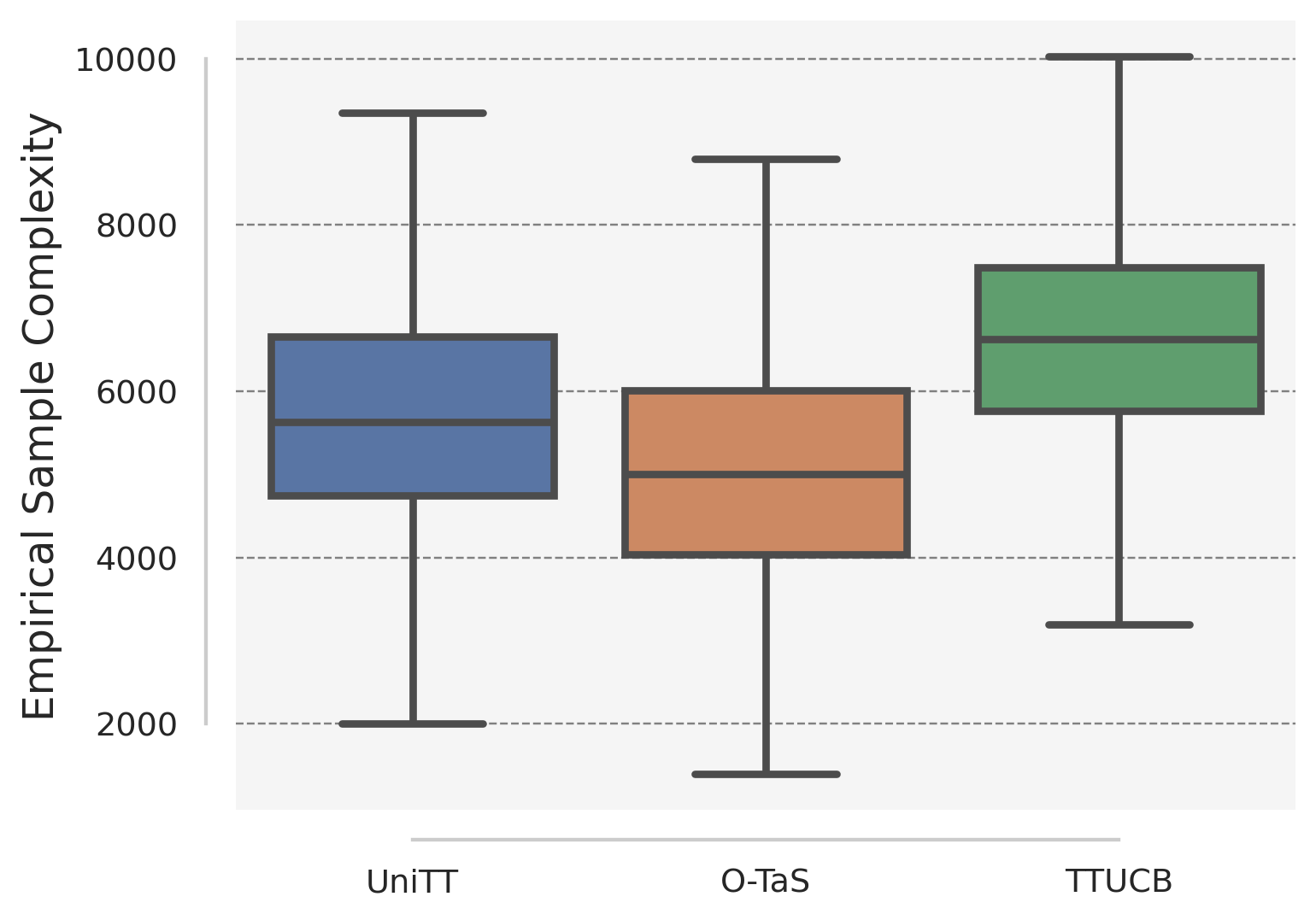}
  \caption{Ablation on $K$ in random instances. Empirical sample complexity on $\bm{\mu}_{\textup{R}, 100}$.}
  \label{fig:k100}
\end{minipage}
%\hfill
\end{figure*}

\begin{figure*}[t!]
\centering
\begin{minipage}{.45\textwidth}
  \centering
  \includegraphics[width=\textwidth]{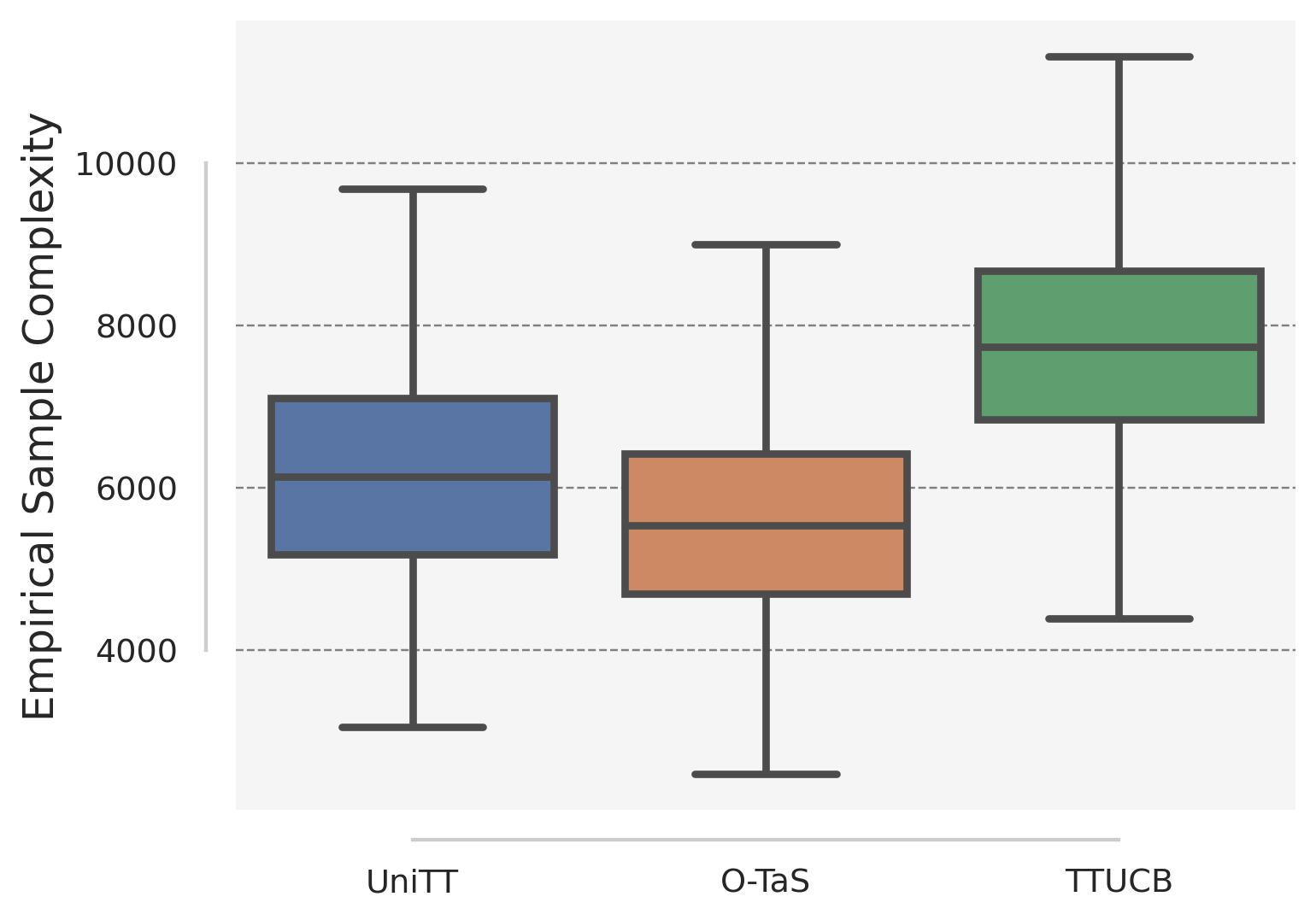}
  \caption{Ablation on $K$ in random instances. Empirical sample complexity on $\bm{\mu}_{\textup{R}, 500}$.}
  \label{fig:k500}
\end{minipage}%
\hspace{.55cm} % Adjust the horizontal space between figures here
\begin{minipage}{.45\textwidth}
  \centering
  \includegraphics[width=\textwidth]{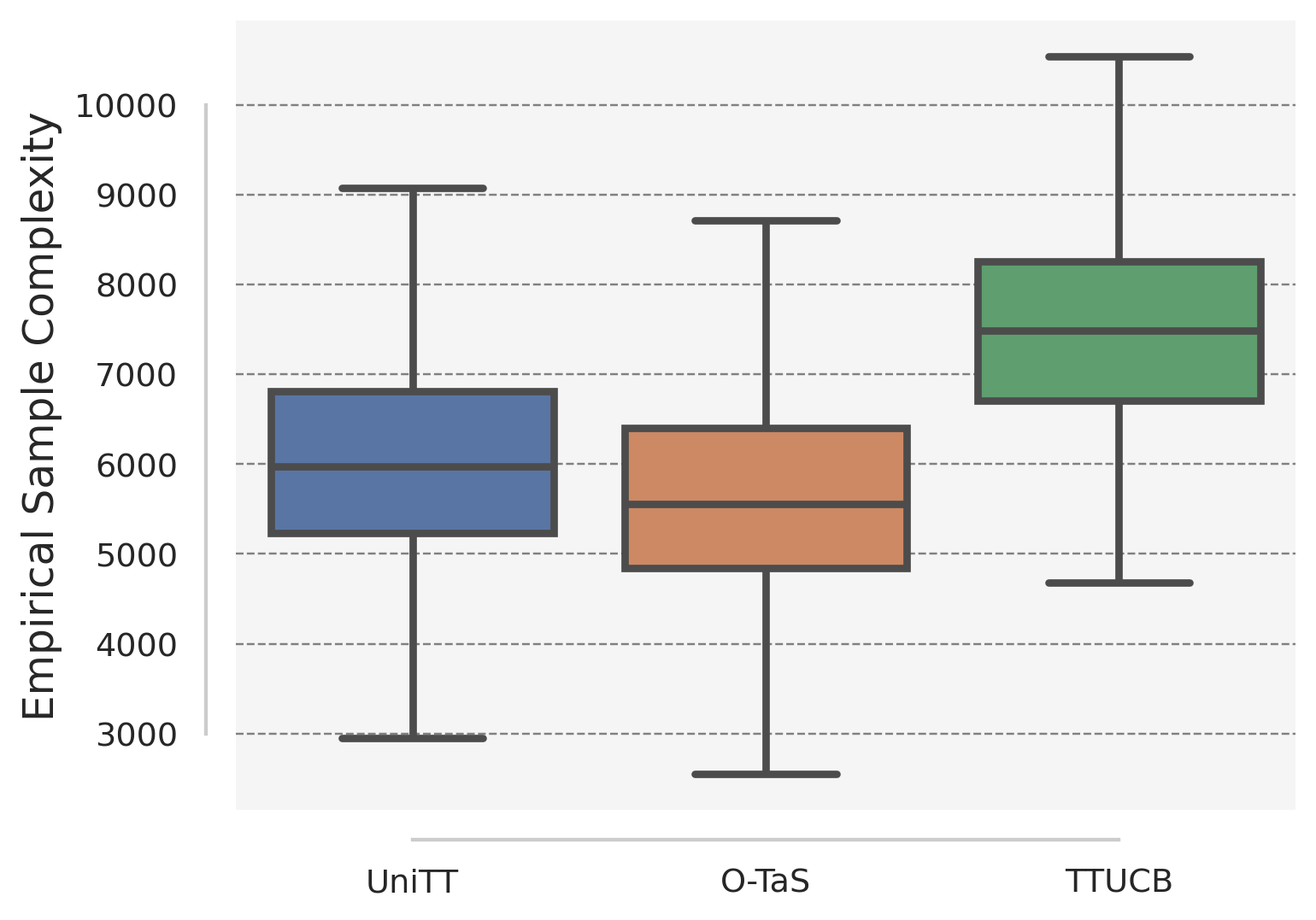}
  \caption{Ablation on $K$ in random instances. Empirical sample complexity on $\bm{\mu}_{\textup{R}, 1000}$.}
  \label{fig:k1000}
\end{minipage}
%\hfill
\end{figure*}

That being said, our goal is studying the benefits of exploiting the unimodal structure against relying on algorithms for unstructured BAI.
We consider $K = 10$, $K=100$, $K=500$ and $K=1000$.\footnote{The exact bandit models that have been generated can be found in the codebase.} We denote these bandit models as $\bm{\mu}_{\textup{R}, K}$ (note that $\bm{\mu}_{\textup{R}, 10}$ and $\bm{\mu}_{\textup{R}, 100}$ are the same bandit models that we presented in Section~\ref{sec:expe}; here, we report the same results restricted to the aforementioned algorithms for the ease of read). Figures~\ref{fig:k10}-\ref{fig:k1000} reports the results of $1000$ independent runs. As one can see, for small values of $K$, TTUCB is competitive (although slightly worse) with UniTT and O-TaS. As soon as $K$ increases, the difference between the algorithms amplifies. 

As a final remark, we note that the results for $K=500$ and $K=1000$ are similar. The main reason is that, given the random generation process that we adopted (see Section~\ref{subsec-app:exp-details}, adding additional arms do not affect much the performance of the algorithms. Indeed, by adding new arms, we obtain mean values which are far from being optimal (and, consequently, they can be discarded with very few data).

\subsection{Ablation on UniTT}\label{subsec-app:uni-tt-abl}

\begin{figure*}[t!]
\centering
\begin{minipage}{.45\textwidth}
  \centering
  \includegraphics[width=\textwidth]{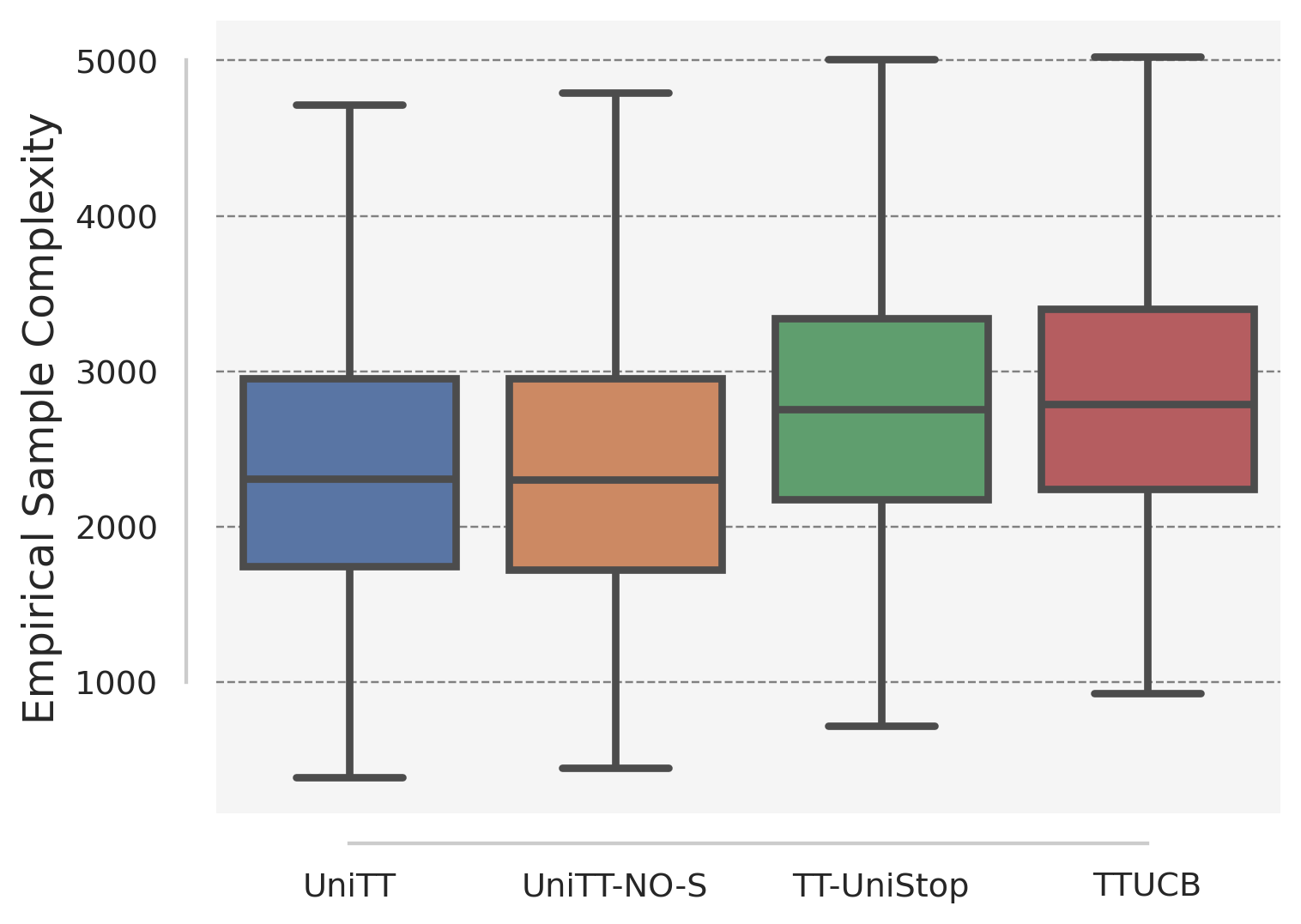}
  \caption{UniTT ablation. Empirical sample complexity on $\bm{\mu}_{\textup{R}, 10}$.}
  \label{fig:uni-tt-abl-r}
\end{minipage}%
\hspace{.55cm} % Adjust the horizontal space between figures here
\begin{minipage}{.45\textwidth}
  \centering
  \includegraphics[width=\textwidth]{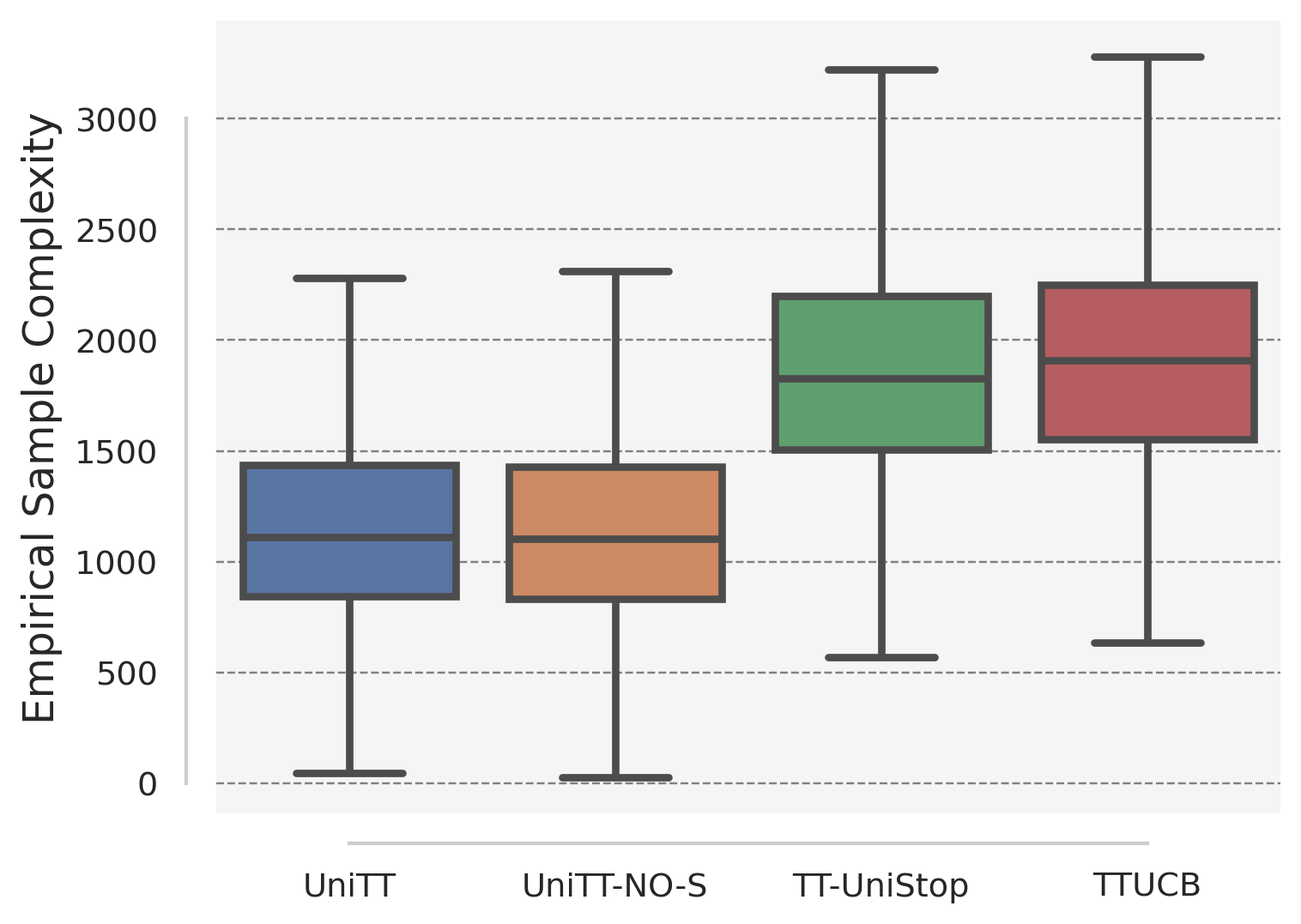}
  \caption{UniTT ablation. Empirical sample complexity on $\bm{\mu}_{\textup{F}, 11}$.}
  \label{fig:uni-tt-abl-f}
\end{minipage}
%\hfill
\end{figure*}

\begin{figure*}[t!]
\centering
\begin{minipage}{.45\textwidth}
  \centering
  \includegraphics[width=\textwidth]{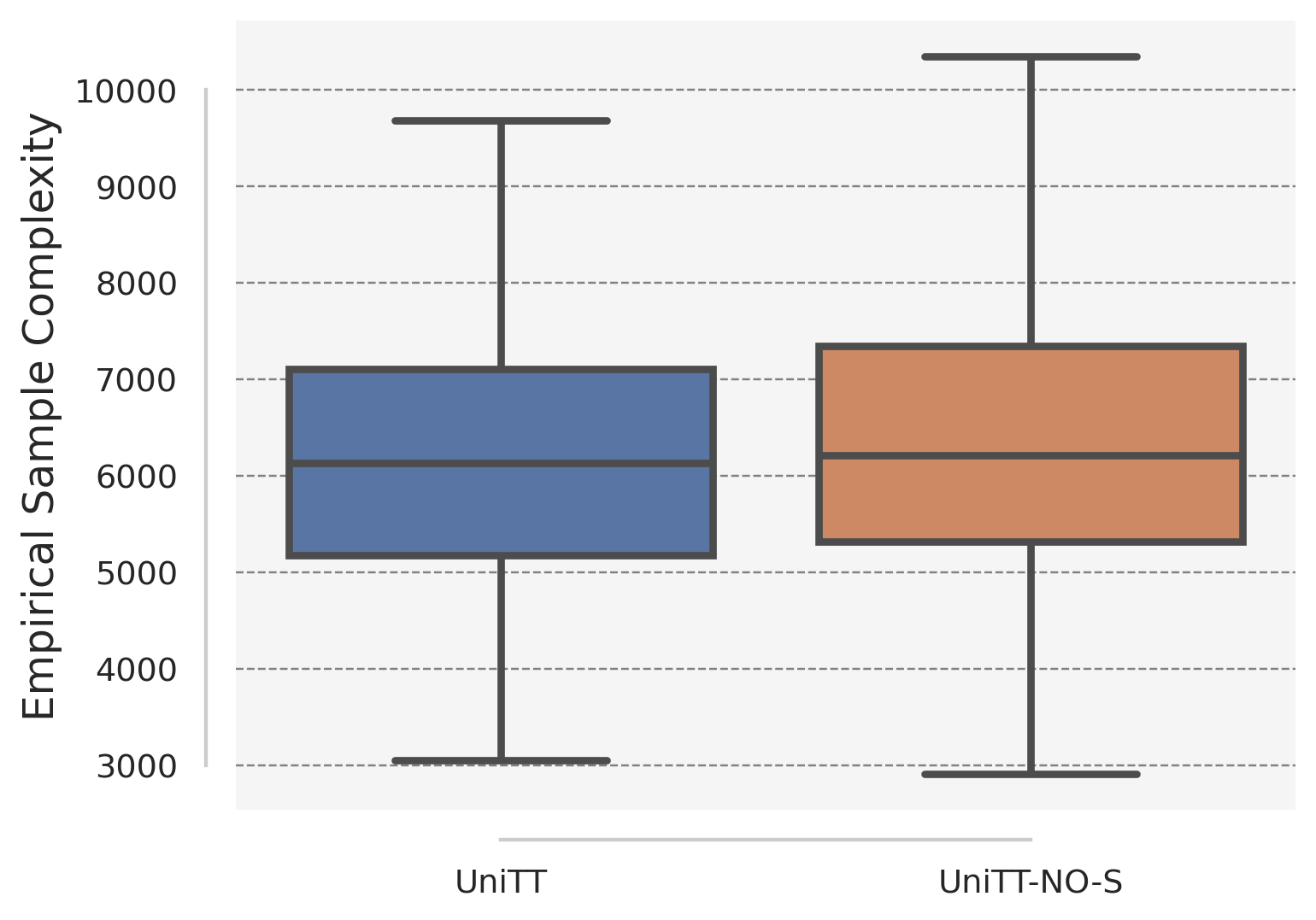}
  \caption{UniTT ablation. Empirical sample complexity on $\bm{\mu}_{\textup{R}, 500}$.}
  \label{fig:uni-tt-abl-r500}
\end{minipage}%
\hspace{.55cm} % Adjust the horizontal space between figures here
\begin{minipage}{.45\textwidth}
  \centering
  \includegraphics[width=\textwidth]{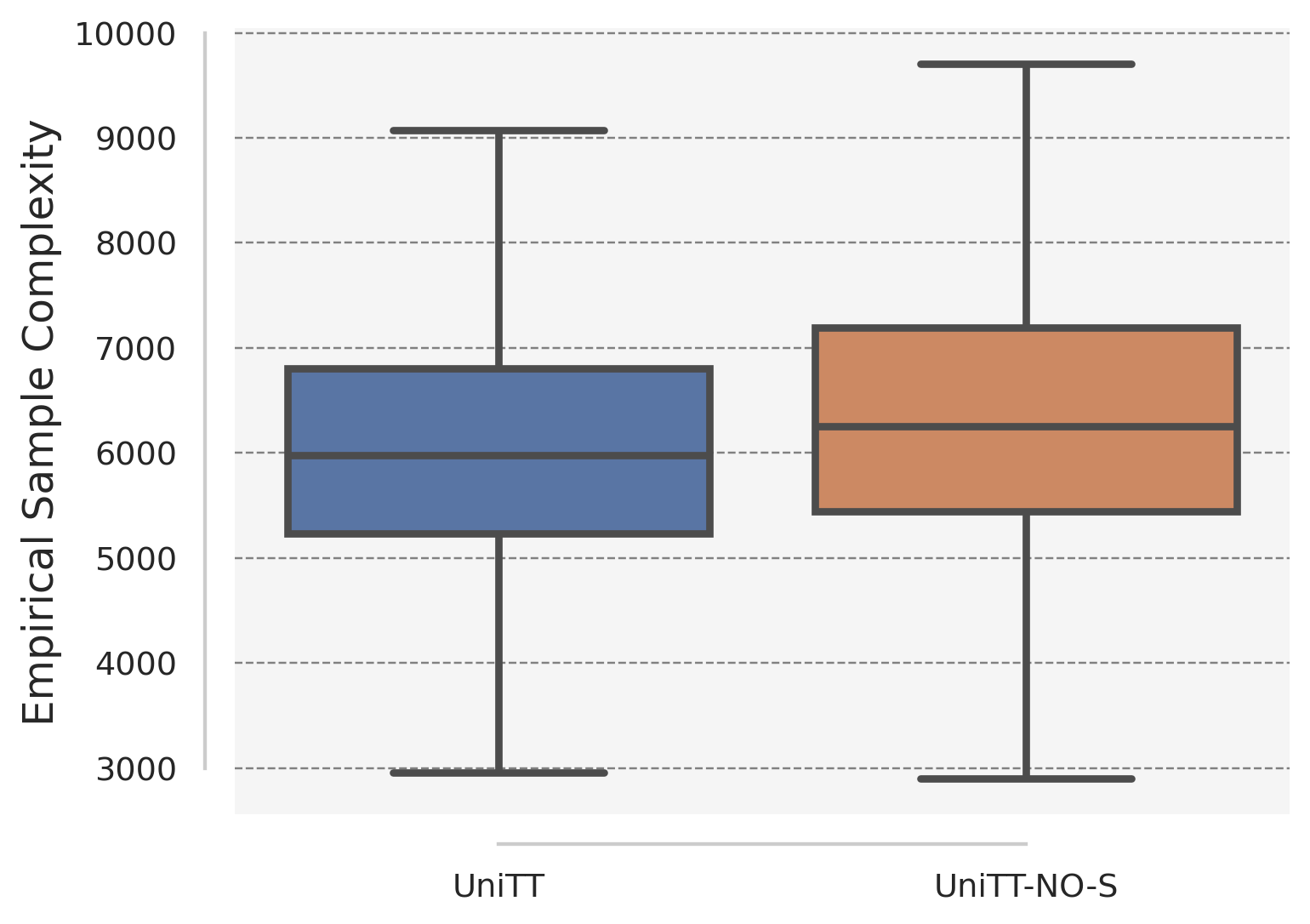}
  \caption{UniTT ablation. Empirical sample complexity on $\bm{\mu}_{\textup{R}, 1000}$.}
  \label{fig:uni-tt-abl-r1000}
\end{minipage}
%\hfill
\end{figure*}

In this section, we present an ablation on UniTT, to understand which components leads to the performance improvement over TT. Specifically, we consider the following algorithms.
\begin{itemize}
\item UniTT, as presented in Section~\ref{sec:algos}.
\item UniTT, as presented in Section~\ref{sec:algos}, but without the structured confidence intervals. In other words, the sampling rule is replaced with $B_{t} \in \argmax_{i \in [K]} \max_{\boldsymbol{\lambda} \in \Theta_t}  \lambda_i$. We refer to this algorithm as UniTT-NO-S. 
\item UniTT, as presenterd in Section~\ref{sec:algos}, but where the stopping rule is replaced with the one of TTUCB \citep{jourdan2023NonAsymptoticAnalysis} (i.e., the usual GLR tests for unstructured BAI problems). We refer to this algorithm as UniTT-BAIStop.
\item TTUCB from \cite{jourdan2023NonAsymptoticAnalysis}, but we replace the stopping rule with the one that we presented in Section~\ref{sec:algos}. We refer to this algorithm as TT-UniStop.
\item TTUCB \citep{jourdan2023NonAsymptoticAnalysis}.
\end{itemize}
We first benchmark the performance of these algorithms in $\bm{\mu}_{R,10}$ and $\bm{\mu}_{F,11}$ that we presented in Section~\ref{sec:algos}. Figures~\ref{fig:uni-tt-abl-r} and \ref{fig:uni-tt-abl-f} report the results. 

First of all, we can observe that the gap between TTUCB and TT-UniStop only plays a minor role in the performance. Indeed, TT-UniStop is only slightly better than TTUCB for unstructured BAI. Secondly, we actually excluded UniTT-BAIStop from the plots. Indeed, we have verified that it might requires a huge number of samples in order to stop. The reason is a combination of the following two factors: (i) the goal of the sampling rule is to target arms within $\mathcal{N}(\star)$, (ii) the stopping rule from unstructured BAI requires to distinguish $\mu_{\star}$ from every arm $a \ne \star$. From the results of TT-UniStop and UniTT-BAIStop, we can infer that, in order to gain benefits from the the unimodal structure, it is important to consider the combination of both the sampling and the stopping rule.

Secondly, we observe that the performance of UniTT and UniTT-No-S are identical in these two bandits. In other words, in these domains, the structured confidence intervals do not play a role in the performance, and it is sufficient to consider the optimism from $\Theta_t$ in order to obtain a good performance level.\footnote{In this sense, the reason for performance improvement, conditioned on the fact that we are using a stopping rule tailored for unimodal bandits, come from how the challenged is selected, i.e., it is searched among the arms in $\mathcal{N}(B_t)$, while in unstructed BAI the list of possible challengers is $[K]$.} Nevertheless, as we analyzed in our theory, we know that, structured confidence intervals can help in mitigating the number of pulls to arms for which $\omega^*_i = 0$. For this reason, we ablate UniTT against UniTT-No-S by increasing the number of arms $K$ in random unimodal bandits. Specifically, we compare the performance of the two algorithms on two random unimodal bandits with $K=500$ and $K=1000$. Figures~\ref{fig:uni-tt-abl-r500} and \ref{fig:uni-tt-abl-r1000} reports the results. As we can see, as soon as we increase the number of arms, the performance gap between the two methods increases.

\subsection{Ablation on $\widetilde{\Theta}_t$}\label{subsec-app:theta}
Finally, we now present a visualization of the set $\widetilde{\Theta}_t$, and how it evolves during training (both for O-TaS and UniTT). Specifically, we visualize the number of arms for which there exists a unimodal bandits within $\widetilde{\Theta}_t$ whose optimal arm $i$. We refer to this number as the number of "active" arms.

\begin{figure*}[t]
\centering
  \includegraphics[width=0.4\textwidth]{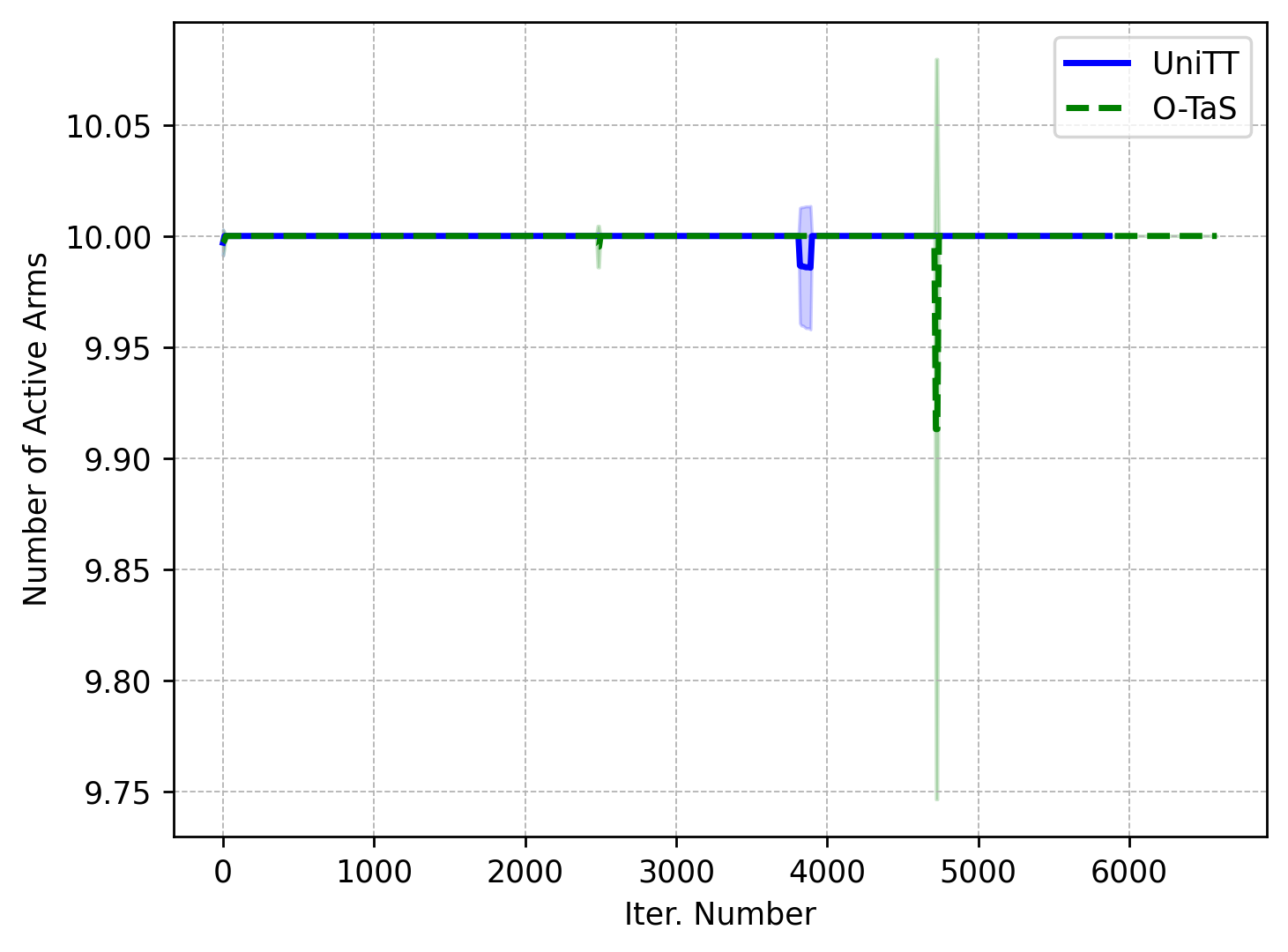}
  \hspace{0.2cm}
  \includegraphics[width=0.4\textwidth]{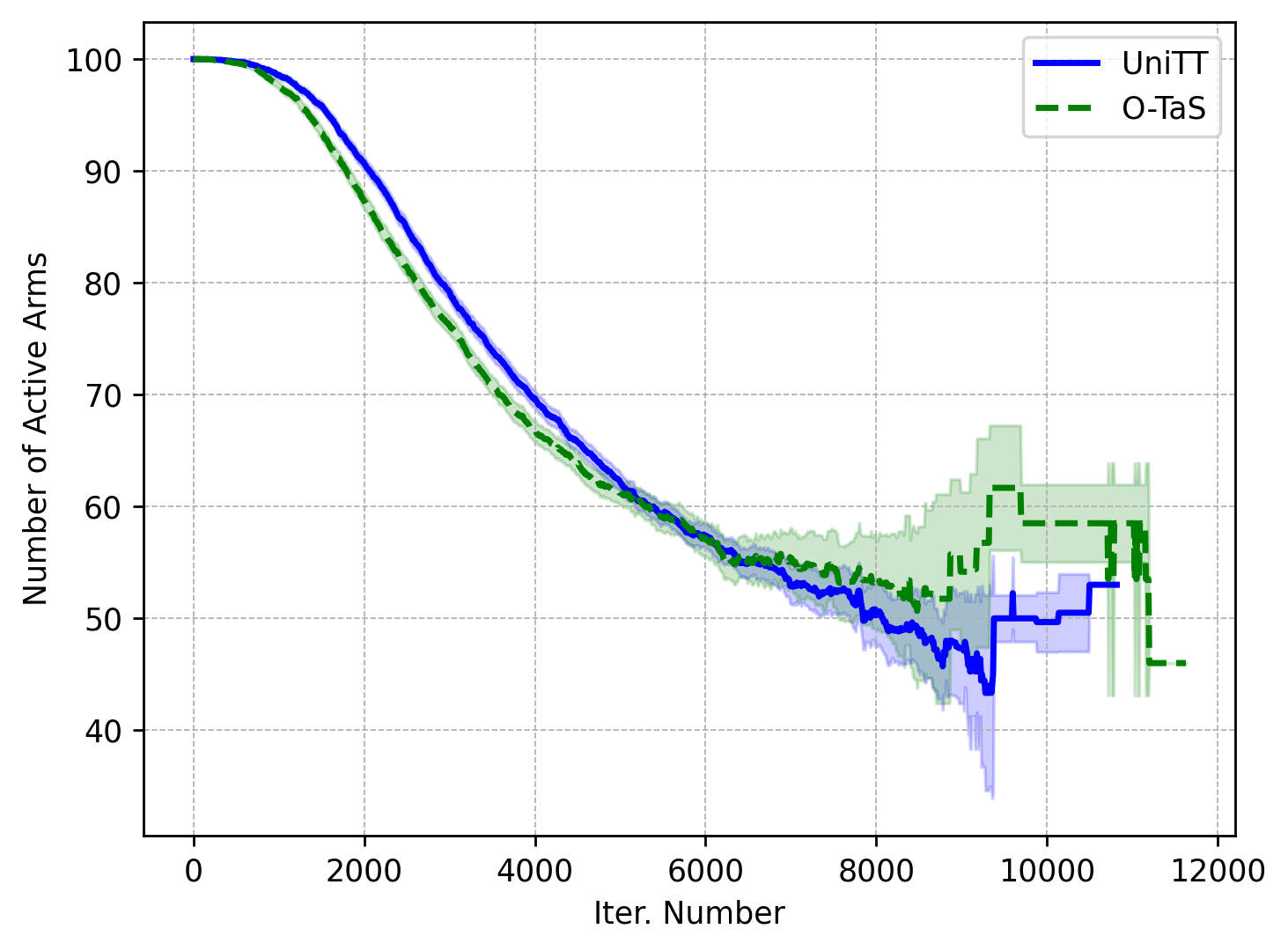}\\
  \includegraphics[width=0.4\textwidth]{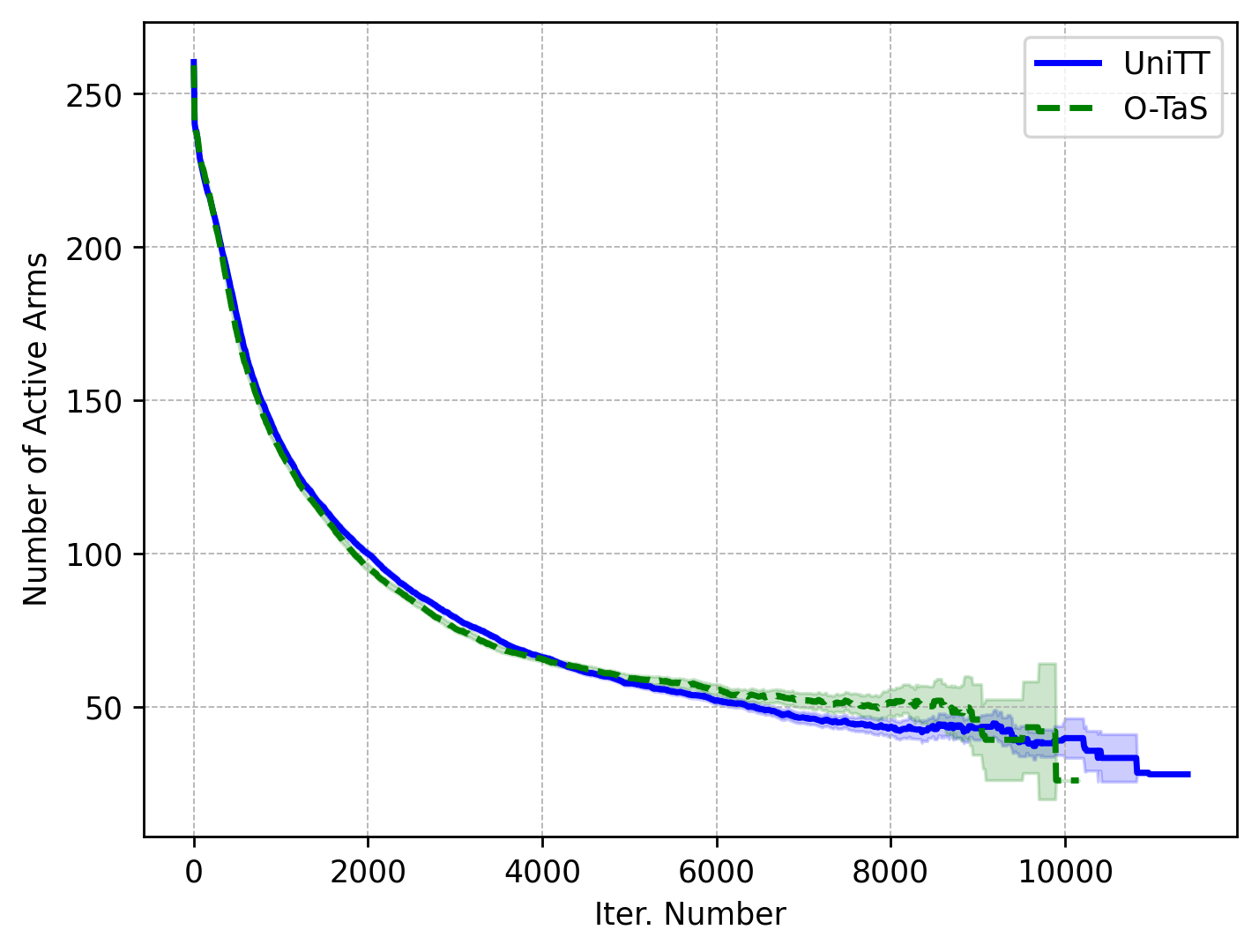}
  \hspace{0.2cm}
  \includegraphics[width=0.4\textwidth]{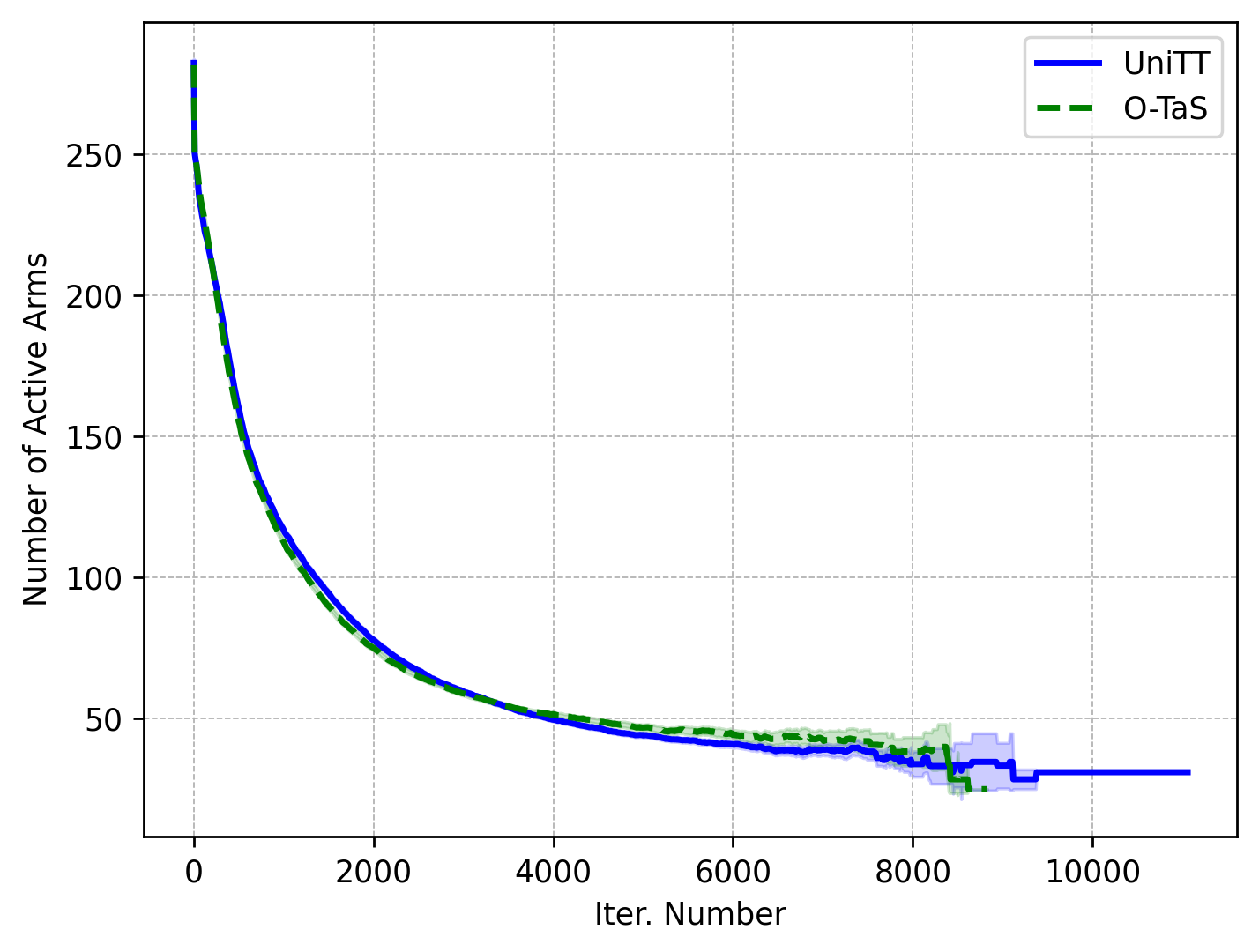}
  \caption{Empirical number of active arms when running UniTT and O-TaS with $\delta = 0.01$. The considered instances are $\bm{\mu}_{R,10}$ (top, left), $\bm{\mu}_{R,100}$ (top, right), $\bm{\mu}_{R,500}$ (bottom, left) and $\bm{\mu}_{R,1000}$ (bottom right). Results are average results of $1000$ independent runs with $95\%$ confidence intervals.}
  \label{fig:theta-all}
\end{figure*}

As bandit instances, we consider $\bm{\mu}_{R,K}$ with $K \in \{ 10, 100, 500, 1000 \}$. Figure~\ref{fig:theta-all} reports the average number of active arms for $1000$ independent runs of O-TaS and UniTT. We observe that for a small value of $K$ (i.e., $K=10$) the structured confidence intervals do not play a significant role as the number of active arms remain constant before that the algorithm can stop. This, intuitively, is justified by two facts: (i) for a small number of arms, both algorithms stop sooner, and (ii) given the random generation process that we described in Section~\ref{subsec-app:exp-details}, the values of the means are all close to the one of the optimal arm. As soon as we increase the number of arms, both these effects vanishes, and the number of active arms significantly decrease while learning progress. 

Finally, as we highlighted in Section~\ref{subsec-app:k-rnd}, the results for $K=500$ and $K=1000$ are similar (the main difference is that the starting point for $K=1000$ is slightly higher). The reason is the same that we presented in Section~\ref{subsec-app:k-rnd}, i.e., given the random generation process that we adopted (see Section~\ref{subsec-app:exp-details}, adding additional arms do not affect much the behavior of the algorithms. Indeed, by adding new arms, we obtain mean values which are far from being optimal. As a consequence, these new arms can easily discarded from the set of active arms.

\end{appendix}

\end{document}



% Supplementary material: To improve readability, you must use a single-column format for the supplementary material.
\onecolumn
\aistatstitle{Supplementary Materials}

\section{FORMATTING INSTRUCTIONS}

To prepare a supplementary pdf file, we ask the authors to use \texttt{aistats2025.sty} as a style file and to follow the same formatting instructions as in the main paper.
The only difference is that the supplementary material must be in a \emph{single-column} format.
You can use \texttt{supplement.tex} in our starter pack as a starting point, or append the supplementary content to the main paper and split the final PDF into two separate files.

Note that reviewers are under no obligation to examine your supplementary material.

\section{MISSING PROOFS}

The supplementary materials may contain detailed proofs of the results that are missing in the main paper.

\subsection{Proof of Lemma 3}

\textit{In this section, we present the detailed proof of Lemma 3 and then [ ... ]}

\section{ADDITIONAL EXPERIMENTS}

If you have additional experimental results, you may include them in the supplementary materials.

\subsection{The Effect of Regularization Parameter}

\textit{Our algorithm depends on the regularization parameter $\lambda$. Figure 1 below illustrates the effect of this parameter on the performance of our algorithm. As we can see, [ ... ]}

\vfill